\documentclass{article}

\PassOptionsToPackage{numbers,sort&compress}{natbib}

\usepackage[preprint,nonanonymous]{neurips_2026}
\usepackage[T1]{fontenc}
\usepackage[utf8]{inputenc}
\usepackage{microtype}
\usepackage{amsmath,amssymb,mathtools}
\usepackage{booktabs}
\usepackage{array}
\usepackage{tabularx}
\usepackage{graphicx}
\usepackage{float}
\usepackage{xcolor}
\usepackage{hyperref}
\usepackage{url}
\usepackage{tcolorbox}
\usepackage{fvextra}
\usepackage{subcaption}
\usepackage{enumitem}
\usepackage{multirow}
\usepackage{amsthm}

\DefineVerbatimEnvironment{breakverb}{Verbatim}{%
  fontsize=\footnotesize,
  breaklines=true,
  breakanywhere=true,
  breakautoindent=true,
  breaksymbolleft=\hbox{}}

\hypersetup{
  colorlinks=true,
  linkcolor=blue!50!black,
  citecolor=blue!50!black,
  urlcolor=blue!50!black
}

\newcommand{\talan}{\textsc{TALAN}}

\newcommand{\bench}[1]{\textsc{#1}}
\newcolumntype{Y}{>{\raggedright\arraybackslash}X}
\newcolumntype{P}[1]{>{\raggedright\arraybackslash}p{#1}}

\definecolor{talanblue}{RGB}{33,94,173}
\definecolor{talanlight}{RGB}{236,243,252}
\definecolor{talanorange}{RGB}{203,109,44}
\definecolor{talangray}{RGB}{88,99,117}
\definecolor{talangreen}{RGB}{46,139,87}
\definecolor{talanred}{RGB}{180,60,60}

\newtcolorbox{killerbadge}{
  colback=talanlight, colframe=talanblue, boxrule=0.6pt,
  arc=2mm, left=6pt, right=6pt, top=4pt, bottom=4pt,
  fontupper=\small
}

\title{\talan{}: Task-Aligned Latent Adaptation Networks\\
for Targeted Post-Training of Large Language Models}

\author{
Chengkai Zhang$^{1}$\thanks{Sole first author. Correspondence to \href{mailto:chengkai@meta.com}{chengkai@meta.com}.}\quad
Ziteng Liu$^{1}$\thanks{Ziteng Liu, Junpu Wang, and Zeyi Tao contributed equally as co-second authors.}\quad
Junpu Wang$^{1}$\footnotemark[2]\quad
\textbf{Zeyi Tao}$^{1}$\footnotemark[2]\\
Yang Wang$^{1}$\quad
Sagar Chordia$^{1}$\quad
Qin Huang$^{1}$\\
$^{1}$Meta AI\\
\texttt{\{chengkai,zitengliu,junpuwang,taozeyi1990,yvv,sagarc,huginhuang\}@meta.com}
}
\date{}

\begin{document}
\maketitle

\begin{abstract}
\noindent
Targeted post-training aims to improve capabilities such as scientific
reasoning, mathematics, and code generation without regressing
already-strong behavior. Standard low-rank adapters are efficient but
task-global, while activation-intervention methods are more input-aware
but often require separate probe fitting, vector extraction, or
inference-time steering. We introduce \talan{} (\emph{Task-Aligned
Latent Adaptation Networks}), a sequence-conditioned latent side path
inserted into a transformer's residual stream and co-trained with a
low-rank adapter in a single SFT loop. \talan{} summarizes the active
sequence into a compact latent memory, remixes that memory into
token-level perturbations, and writes the perturbations back through a
controlled residual update. The block is fixed; adapting it to a host
selects a six-axis configuration over insertion location, memory
size, mixer, writeback rule, trainability scope, and gradient scale.

Across four Qwen3-family backbones and four STEM/code benchmarks,
selected \talan{} operating points improve matched adapter baselines
under both LoRA and DoRA.
Under LoRA, \talan{} reaches a cross-model mean gain of $+1.41$
percentage points, is positive on all four backbones, and is
non-negative on all 16 model--benchmark cells. Under DoRA, it reaches a
cross-model mean gain of $+1.85$ percentage points and is positive on
all four backbones, with positive deltas on 13 of 16 cells. Paired
seed checks support positive average effects while showing nontrivial
variance, so we use them as sensitivity checks rather than as the main
claim. The added computational cost is small:
less than $1\%$ trainable parameters relative to the backbone and
$1.01$--$1.02\times$ inference overhead relative to matched LoRA.
A selected non-Qwen transfer check on Llama-3.2-1B is also positive
under both LoRA and rsLoRA across seven paired seeds, supporting
a limited transfer probe beyond the primary Qwen-family suite while
not claiming exhaustive non-Qwen coverage.

Internal-state analyses support the view that \talan{} acts as a small,
complementary activation intervention: the matched adapter update is
$80$--$1{,}700\times$ larger in norm than the \talan{} perturbation,
yet the two directions have near-zero cosine, and per-layer activation
measurements show that the small orthogonal perturbation propagates and
amplifies through depth. These results establish \talan{} as a
practical platform for studying steerable activation-level adaptation
within standard adapter-based post-training.
\end{abstract}


\section{Introduction}

As language models strengthen, post-training becomes increasingly
targeted: the goal is often to improve a capability region such as
scientific reasoning, mathematics, or code generation without trading
off already-strong behavior. This raises a narrower question than
generic parameter-efficient fine-tuning: can adaptation be made
sequence-aware and capability-directed while retaining the simplicity
of a standard supervised fine-tuning pipeline?

Parameter-efficient methods, including adapters, prompt tuning,
IA$^3$, LoRA, and DoRA, remain attractive because they are efficient,
stable, and easy to deploy
~\citep{houlsby2019parameter,pfeiffer2021adapterfusion,li2021prefix,lester2021power,liu2022ia3,hu2022lora,liu2024dora}.
Their limitation in this setting is granularity. A low-rank adapter
acts as a task-global parameter patch: the same update is available
regardless of which sequence is being processed. Intervention-oriented
methods such as activation steering, representation editing, ReFT,
CogSteer, and reasoning-time interventions move closer to input-aware
control
~\citep{turner2024activation,zou2023representation,wu2024reft,cogsteer2025,thinkingintervention2025},
but often require probe fitting, vector extraction, inference-time
steering, or multi-stage procedures. We seek a single end-to-end SFT
module whose forward path is conditioned on the active sequence while
still composing cleanly with low-rank adapters.

We introduce \talan{} (\emph{Task-Aligned Latent Adaptation
Networks}), a sequence-conditioned latent side path inserted into the
residual stream and co-trained with a low-rank adapter. \talan{}
summarizes the active sequence into a compact latent memory, remixes it
into token-level perturbations, and writes those perturbations back
through a controlled residual update. The block is fixed; adapting it
to a new backbone means choosing a configuration
in a six-axis configuration space. This construction gives two
testable properties: \textbf{P1}, \talan{} introduces a small bounded
activation perturbation; and \textbf{P2}, the perturbation is
architecturally separate from the adapter update.

On four Qwen3-family backbones and four STEM/code benchmarks, selected
\talan{} operating points improve matched adapter baselines under both
LoRA and DoRA. Under the LoRA pipeline, \talan{} reaches a cross-model
mean gain of $+1.41$ percentage points (pp), is positive on all four
backbones, and is non-negative on all 16 model--benchmark cells. Under
the DoRA pipeline, \talan{} reaches a cross-model mean gain of
$+1.85$ pp and is positive on all four backbones, with positive deltas
on 13 of 16 model--benchmark cells. A matched-initialization multi-seed
audit supports positive family-level effects under both adapter
pipelines while also quantifying seed-level variance
(Appendix~\ref{app:matched}). The intervention is small in cost,
adding less than $1\%$ trainable parameters relative to
the backbone and keeping inference overhead within
$1.01$--$1.02\times$ relative to matched LoRA.
As a targeted transfer check outside the primary Qwen-family suite,
selected Llama-3.2-1B paired runs are also positive across seven seeds
under both LoRA and rsLoRA (Section~\ref{sec:nonqwen-transfer}).

\paragraph{Toward steerable adaptation.}
Beyond the practical gains, the orthogonality finding points toward a
broader goal: steerable adaptation, where the right activation
intervention is designed from a model's internal geometry rather than
found by search. \talan{} is a useful step in this direction because it
turns activation modification into a structured, sequence-conditioned
operation: the latent memory summarizes the current sequence, the remix
stage converts that summary into token-level perturbations, and the
writeback stage controls how the perturbation enters the residual
stream. The six-axis configuration space exposes the key intervention
choices, including where to insert the perturbation and how strongly to
write it back. In Section~\ref{sec:mechanism}, we analyze whether this
small complementary perturbation behaves as intended, focusing on
adapter-update orthogonality (Section~\ref{sec:insight-orthogonality})
and amplification through later layers
(Section~\ref{sec:insight-design-language}).

\paragraph{Contributions.}
\begin{enumerate}[nosep,leftmargin=*]
  \item We introduce \talan{}, a sequence-conditioned
    summarize--remix--writeback architecture with a six-axis
    configuration space and two testable properties: small bounded
    activation perturbations (P1) and orthogonality to the adapter
    update (P2).
  \item We validate selected \talan{} operating points on four
    Qwen3-family backbones and four STEM/code benchmarks under LoRA and
    DoRA, improving matched adapter baselines with cross-model mean
    gains of $+1.41$ pp and $+1.85$ pp, respectively; under LoRA, the
    audited configurations are non-negative on all 16 model--benchmark
    cells. Paired seed checks quantify variance while remaining
    positive on average at the family level. A selected
    non-Qwen Llama-3.2-1B transfer check is additionally positive under
    both LoRA and rsLoRA over seven paired seeds.
  \item We probe \talan{}'s internal states and show that the small
    orthogonal perturbation propagates and amplifies through depth,
    establishing \talan{} as a viable platform toward steerable
    activation-level adaptation.
\end{enumerate}


\section{Related Work}
\label{sec:related}

\paragraph{Parameter-efficient fine-tuning (PEFT).}
LoRA~\citep{hu2022lora} and its variants have
become the standard for adapting large LMs at low parameter cost,
typically through low-rank weight updates on attention projections.
\textbf{DoRA}~\citep{liu2024dora} extends LoRA by decomposing each
pre-trained weight matrix into a magnitude vector and a directional
matrix, then applying the LoRA-style low-rank update to the
direction component only; this magnitude/direction split has been
shown to close part of the gap between LoRA and full fine-tuning
while preserving LoRA's parameter-efficiency. Other low-rank
extensions include VeRA~\citep{kopiczko2024vera}, which freezes
shared random low-rank matrices and learns only a small number of
scaling vectors. Adapters~\citep{houlsby2019parameter,he2022unified}
insert MLP bottlenecks between transformer sub-layers;
AdapterFusion~\citep{pfeiffer2021adapterfusion} trains per-domain
adapters and learns a fusion stage on top. Prompt tuning, prefix
tuning, and IA$^3$~\citep{lester2021power,li2021prefix,liu2022ia3}
introduce parameters at the input/key/value level. These low-rank
adapter methods all share two structural properties relevant to our
setting: (i) they apply a \emph{task-global} parameter patch to the
base model, and (ii) they do not differentiate their effect across
the active sequence. \talan{} adds a sequence-conditioned latent
side path that is co-trained with a low-rank adapter (LoRA or
DoRA in our experiments) in a single SFT loop, addressing both
limitations without changing the adapter recipe itself; we evaluate
\talan{} under both adapter pipelines and find that the
architectural intervention is positive on every backbone under each
(\S\ref{sec:headline}). Expert-style methods such as
Branch-Train-MiX~\citep{branchtrainmix2024} route adaptation to
chosen capability clusters through multi-expert composition;
\talan{} achieves sequence-level conditioning within a single SFT
pipeline rather than through routing among separate experts.

\paragraph{Intervention-oriented adaptation.}
A separate line of work modifies internal activations rather than
parameters: representation engineering~\citep{zou2023representation},
activation steering~\citep{turner2024activation}, ReFT~\citep{wu2024reft},
CogSteer~\citep{cogsteer2025}, and reasoning-oriented
intervention~\citep{thinkingintervention2025} construct steering
vectors, direction-aligned probes, or reasoning-time perturbations.
These methods are typically inference-time interventions or require a
probe-fitting / vector-extraction stage; they do not co-train with a
low-rank adapter in a single end-to-end SFT loop. Related recent work
also targets reasoning adaptation from adjacent angles: reasoning-critical
neuron transfer uses activation steering to improve inference
reliability~\citep{adaras2026}, while DELIFT studies data-efficient
instruction fine-tuning~\citep{delift2025}. \talan{} differs from these
methods in (a) integrating the side path with a matched low-rank adapter
(LoRA or DoRA in our experiments) in a single SFT loop, (b) inducing
latent slots \emph{operationally} from the sequence rather than from
external task IDs or probe fitting, and (c) exposing a constrained
six-axis configuration space whose host-specific settings are tied to
the resulting non-regression rate.

\paragraph{Cross-attention and latent-array mechanisms.}
Cross-attention with learned latent queries is now a standard
architectural pattern, used for multimodal fusion in
Flamingo~\citep{alayrac2022flamingo} and for general-purpose
perception in the Perceiver~\citep{jaegle2021perceiver}, among
others. \talan{} adopts the same mechanism---learned slot queries
attending to the input sequence---as an efficient way to summarize
the active sequence into a compact latent memory within
adapter-based SFT.


\section{TALAN: Task-Aligned Latent Adaptation Networks}
\label{sec:method}

\noindent
We present \talan{} at three levels.
\textbf{As a module} (\S\ref{sec:method-block}), it is a fixed
\emph{summarize--remix--writeback} block inserted into a
transformer's residual stream and co-trained with a low-rank adapter
in a single SFT loop.
\textbf{As a configurable architecture} (\S\ref{sec:method-axes}),
adapting the module to a new backbone amounts to selecting one
configuration
$\pi = (\ell^{\star}, T, m, w, \tau, \gamma)$
inside a six-axis configuration space; the block, training objective,
data, and adapter recipe stay fixed.
We call a base model together with its matched adapter a
\emph{host}.
\textbf{As an activation intervention} (\S\ref{sec:theory}),
\talan{} is designed toward steerable activation intervention, where
the right perturbation---its location, geometry, and magnitude---is
determined from a model's internal structure rather than found by
undirected search.
Here we study a controlled setting in which small,
sequence-conditioned perturbations are inserted at selected
residual-stream locations and evaluated empirically.

\begin{figure*}[t]
\centering
\includegraphics[width=0.96\textwidth]{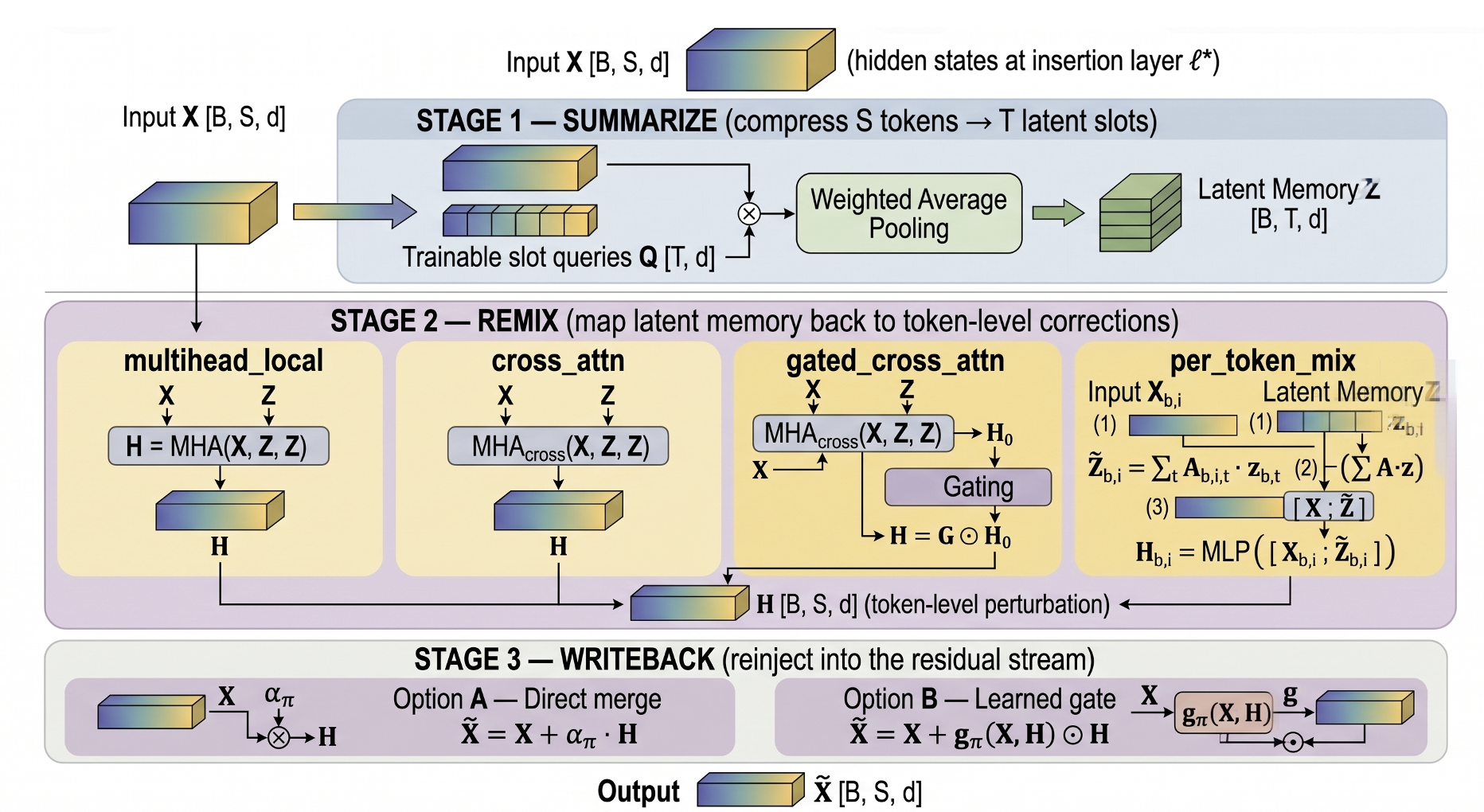}
\caption{\textbf{\talan{} architecture overview.} The block
summarizes the active sequence into a compact latent memory $Z$
(Stage~1, \emph{summarize}), remixes that memory into token-level
perturbation $H$ via one of four mixers (Stage~2, \emph{remix}), and
writes perturbation back through either direct merge or a learned gate
(Stage~3, \emph{writeback}). \talan{} is trained jointly with a
low-rank adapter under the standard autoregressive objective.
Configurations differ by choosing the six architectural axes
$\pi = (\ell^{\star}, T, m, w, \tau, \gamma)$, not by changing the
training data, objective, or adapter recipe.}
\label{fig:overview}
\end{figure*}

\subsection{The module: a canonical summarize--remix--writeback block}
\label{sec:method-block}

Consider a pretrained autoregressive transformer with hidden states
$X^{(\ell)} \in \mathbb{R}^{B \times S \times d}$ at layer $\ell$,
where $B$ is the batch size, $S$ the sequence length, and $d$ the
hidden width. \talan{} inserts a trainable intervention
$f_\theta(\cdot;\pi)$ at exactly one layer $\ell^\star$ and replaces
the residual stream there with
\begin{equation}
\widetilde{X}^{(\ell^\star)} =
f_\theta\!\bigl(X^{(\ell^\star)};\,\pi\bigr)
\;=\;
\mathrm{writeback}_w \circ \mathrm{remix}_m \circ
\mathrm{summarize}_T
\bigl(X^{(\ell^\star)}\bigr).
\label{eq:dataflow}
\end{equation}
The three stages produce, in order, a latent memory $Z$, a token-level
perturbation $H$, and the modified residual stream $\widetilde{X}$.
All other layers are unchanged. The intervention is trained jointly
with a low-rank adapter (LoRA or DoRA in our experiments) under the
same autoregressive language-modeling objective.

\paragraph{Summarize ($X \mapsto Z$).}
\talan{} compresses the active sequence into $T$ latent slots. Let
$Q \in \mathbb{R}^{T \times d}$ be trainable slot queries and
$X_{b,i} \in \mathbb{R}^d$ the residual-stream vector for batch
element $b$ at token position $i$. Each token computes a soft
assignment over the $T$ slots:
\begin{equation}
A_{b,i,t}
\;=\;
\mathrm{softmax}_t\!\left(
\frac{\langle X_{b,i}, Q_t\rangle}{\sqrt{d}}
\right),
\qquad
\sum_{t=1}^{T} A_{b,i,t} = 1.
\label{eq:summarize}
\end{equation}
Each latent slot then collects an assignment-weighted token average:
\begin{equation}
z_{b,t}
\;=\;
\frac{\sum_{i=1}^{S} A_{b,i,t} X_{b,i}}
{\sum_{i=1}^{S} A_{b,i,t} + \varepsilon}
\;\in\; \mathbb{R}^d,
\qquad
\varepsilon = 10^{-6},
\label{eq:pool}
\end{equation}
yielding $Z_b = [z_{b,1};\dots;z_{b,T}]^\top
\in \mathbb{R}^{T \times d}$ and
$Z \in \mathbb{R}^{B \times T \times d}$. The slot queries are
operational rather than semantic: they summarize the current sequence
without task IDs, domain labels, or human-interpretable assignments.
This input-dependent compression is what differentiates \talan{} from
a task-global parameter patch.

\paragraph{Remix ($Z \mapsto H$).}
Given $Z$, \talan{} maps the latent memory back to token-level
perturbation $H \in \mathbb{R}^{B \times S \times d}$ through one of
four mixer families. We write
$\mathrm{MHA}(\mathrm{queries},\mathrm{keys},\mathrm{values})$ for
multi-head attention and $\sigma(\cdot)$ for the elementwise sigmoid.
\begin{itemize}[nosep,leftmargin=*]
  \item \textbf{Latent-memory attention} (\texttt{multihead\_local}):
    $H = \mathrm{MHA}(X, Z, Z)$, so tokens read from the latent
    memory through multi-head attention.
  \item \textbf{Cross-attention} (\texttt{cross\_attn}):
    $H = \mathrm{MHA}_{\mathrm{cross}}(X, Z, Z)$, using separate
    learned projections for token queries and memory keys/values.
  \item \textbf{Gated cross-attention}
    (\texttt{gated\_cross\_attn}):
    $H_0 = \mathrm{MHA}_{\mathrm{cross}}(X, Z, Z)$,
    $G = \sigma(W_g [X;\,H_0])$, and $H = G \odot H_0$, allowing
    input-conditional bypass when the latent memory is unhelpful.
  \item \textbf{Per-token MLP} (\texttt{per\_token\_mix}):
    $\widetilde{Z}_{b,i} = \sum_t A_{b,i,t} z_{b,t}$ and
    $H_{b,i} = \mathrm{MLP}([X_{b,i};\,\widetilde{Z}_{b,i}])$, a
    non-attention alternative.
\end{itemize}

\paragraph{Writeback ($H \mapsto \widetilde{X}$).}
The perturbation $H$ is reinjected into the residual stream through one
of two writeback rules:
\begin{equation}
\widetilde{X}
\;=\;
X + \alpha_\pi H
\quad (\text{direct merge})
\qquad\text{or}\qquad
\widetilde{X}
\;=\;
X + g_\pi(X,H) \odot H
\quad (\text{learned gate}),
\label{eq:writeback}
\end{equation}
where $\alpha_\pi > 0$ is a trainable scalar writeback coefficient
and $g_\pi(X,H) \in \mathbb{R}^{d}$ is a trainable
input-conditional vector gate. The writeback rule expresses the
intervention budget: larger coefficients or wider gates allow a
stronger activation perturbation, while selective gates preserve
already-strong behavior.

\subsection{The configurable architecture: a six-axis space}
\label{sec:method-axes}

\begin{figure}[t]
\centering
\includegraphics[width=0.92\columnwidth]{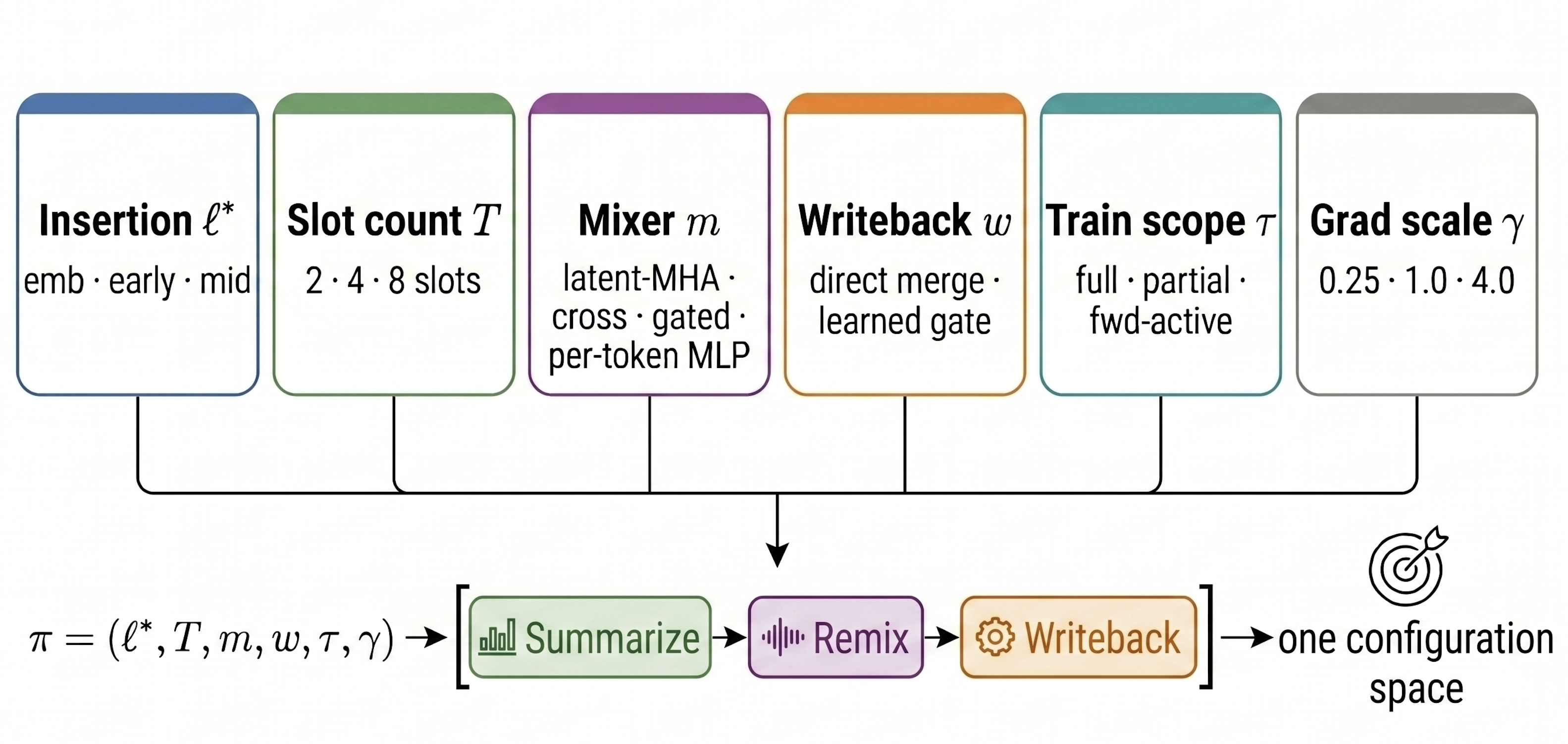}
\caption{\textbf{\talan{} configuration axes.} Six constrained axes
compose into one configuration
$\pi = (\ell^\star, T, m, w, \tau, \gamma)$. Per-backbone selection
chooses $\pi$ inside this family; it does not change the training
objective, data, or low-rank adapter recipe.}
\label{fig:axes}
\end{figure}

The block in Equation~\eqref{eq:dataflow} stays fixed across all
experiments. What changes per backbone is the configuration
\begin{equation}
\pi = (\ell^\star,\, T,\, m,\, w,\, \tau,\, \gamma),
\label{eq:pi}
\end{equation}
whose six axes are listed below. We refer to a selected value of
$\pi$ as a \emph{configuration}.
\begin{enumerate}[nosep,leftmargin=*]
  \item \textbf{Insertion layer $\ell^\star$:} where the block is
    inserted in the residual stream (embedding, early layer, or
    mid-layer).
  \item \textbf{Slot count $T$:} how many latent slots the summarize
    stage uses to compress the active sequence.
  \item \textbf{Mixer $m$:} which remix family maps the latent memory
    back to token-level perturbation.
  \item \textbf{Writeback rule $w$:} direct merge or learned gate.
  \item \textbf{Trainability scope $\tau$:} which \talan{}
    submodules receive gradients, ranging from fully trainable to
    forward-active-only.
  \item \textbf{Gradient scale $\gamma$:} multiplier applied to
    \talan{} parameter gradients.
\end{enumerate}

Only $\pi$ changes across backbones. Within each matched comparison,
the training data, autoregressive objective, adapter rank, adapter
target modules, optimizer recipe, and evaluation protocol are held
fixed. This makes per-backbone selection constrained navigation inside
one configuration space rather than a new method for each host.

\subsection{Architectural properties: small perturbations in a complementary space}
\label{sec:theory}

The construction above has two testable consequences for how
\talan{} interacts with a matched low-rank adapter. Let
\begin{equation}
\Delta_{\mathrm{TALAN}}(X) = f_\theta(X;\pi) - X
\end{equation}
be \talan{}'s additive activation perturbation at the insertion layer
and let
\begin{equation}
\Delta_{\mathrm{adapter}}(X)
= X \cdot s_{\mathrm{adapter}}(BA)^\top
\end{equation}
denote the matched adapter contribution at the same residual-stream
location, where $A \in \mathbb{R}^{r \times d}$ and
$B \in \mathbb{R}^{d \times r}$ are the low-rank factors and
$s_{\mathrm{adapter}}$ is the adapter's effective scale.

\begin{itemize}[nosep,leftmargin=*]
  \item \textbf{P1 (small bounded activation perturbation).}
    The long-term aim of steerable activation intervention is to
    influence model behavior with a targeted perturbation at the right
    internal location. \talan{} uses this aim as a design guide, but
    makes a narrower, testable claim: its residual-stream perturbation
    is explicitly bounded and small relative to the host
    representation. For a direct writeback configuration,
    \begin{equation}
    \|\Delta_{\mathrm{TALAN}}(X)\|_F
    \;\leq\;
    \alpha_\pi \,
    \|W_{\mathrm{mix},\pi}\|_{\mathrm{op}}
    \,\|X\|_F,
    \label{eq:talan-bound}
    \end{equation}
    with analogous gating-dependent bounds for learned writeback.
    Thus, a well-chosen insertion layer $\ell^\star$ and correction
    direction can, in principle, influence downstream computation
    without overwhelming the host representation. We test this
    empirically through layer-wise activation divergence and
    configuration-specific measurements.

  \item \textbf{P2 (orthogonality).}
    \talan{}'s perturbation is produced by a sequence-conditioned
    side path in the residual stream, while LoRA and DoRA update the
    model through low-rank adapter factors. The two paths have no
    shared weights and no adapter-specific parameterization. Thus,
    \talan{} is not another low-rank update inside the adapter
    subspace; it contributes a complementary residual-stream
    perturbation whose direction is not directly parameterized by the
    adapter. This architectural separation predicts near-zero
    adapter-vs-\talan{} alignment and motivates why the same
    \talan{} configuration space can compose with both LoRA and DoRA.
\end{itemize}

Together, P1 and P2 characterize \talan{} as a small,
sequence-conditioned activation intervention placed in a complementary
space to the low-rank adapter. The empirical sections test both
claims rather than assume them: whether small perturbations at the
selected insertion layer propagate through the host computation, and
whether the measured adapter-vs-\talan{} update directions remain
nearly orthogonal in the insertion-layer measurements of
Section~\ref{sec:insight-orthogonality}. The realized magnitude
asymmetry between the adapter update and the \talan{} perturbation is
therefore treated as an empirical measurement in
Section~\ref{sec:insight-orthogonality}, not as an architectural
premise of the method.

\section{Experiments}
\label{sec:experiments}

\noindent
We evaluate \talan{} against adapter-only, adapter-composition, and
intervention-style baselines on four Qwen3-family backbones and four
STEM/code benchmarks. Within each backbone, all methods share the same
training data and evaluation protocol; the adapter-based comparisons
also share the same adapter recipe. The main empirical question is
whether the small residual-stream intervention defined in
Section~\ref{sec:method} can improve a targeted capability suite
without broad regression relative to a matched low-rank adapter. Full
training and evaluation details, including extractor logic, seed
audits, statistical tests, and matched-initialization analyses, are
reported in the appendix.

\subsection{Setup and protocol}
\label{sec:setup}

\paragraph{Models.}
Four Qwen3-family backbones spanning parameter scale, distillation
provenance, and dense-vs-MoE architecture: Qwen3-32B (Q32B, large
dense), DeepSeek-R1-Distill-Qwen-32B (DS32B, reasoning-distilled
dense), Qwen3-8B (Q8B, small dense), and Qwen3-30B-A3B
(MoE)~\citep{qwen3,deepseekr1}.

\paragraph{Training and adapter recipe.}
A fixed 5{,}000-example bench-aligned mixture (math 1{,}729, code
1{,}592, STEM 974, general 705; no test-set leakage) is used
identically for every backbone and method. Two adapter
pipelines---LoRA and DoRA---use rank $r=32$, $\alpha=64$, target
modules $\{q,k,v,o\}$, BF16, DeepSpeed ZeRO-2 on $2\times$H100
80GB, 78 maximum optimization steps. Within each (backbone, adapter)
pair, the matched baseline shares every training-side hyperparameter;
the only treatment difference is the \talan{} module. Full data
composition and training schedule are in
Appendix~\ref{app:protocol}.

\paragraph{Configurations.}
Per-backbone \talan{} configurations are selected inside the six-axis
space of Equation~\eqref{eq:pi}; the four audited configurations are
listed in Table~\ref{tab:operating-points}. Exact configuration
fields are in Appendix~\ref{app:operating-points}.

\begin{table}[!ht]
\centering
\caption{\textbf{Audited \talan{} configurations.} LR/WD are shared
with the matched adapter-only baseline within each backbone.}
\label{tab:operating-points}
\small
\setlength{\tabcolsep}{4pt}
\resizebox{\textwidth}{!}{%
\begin{tabular}{lcccccccc}
\toprule
Backbone & Mixer $m$ & Layer $\ell^{\star}$ & $T$ & $\alpha$ & Writeback $w$ & Trainability $\tau$ & $\gamma$ & LR / WD \\
\midrule
Q32B   & multihead\_local  & layer 12  & 2 & 0.008  & vec gate    & all params                    & 500   & 2e-4 / 0     \\
DS32B  & cross\_attn       & embedding & 2 & 0.001  & vec gate    & all params                    & 500   & 2e-4 / 0     \\
Q8B    & per\_token\_mix   & layer 4   & 2 & 0.0015 & vec gate    & all params, frozen at step 20 & 1000  & 2e-4 / 0     \\
MoE    & multihead\_local  & embedding & 6 & 0.008  & scalar gate & frozen (forward only)         & ---   & 2.5e-5 / 0.02 \\
\bottomrule
\end{tabular}%
}
\\[0.4em]
\footnotesize
For the MoE configuration, \talan{} participates in the forward pass
but receives no gradient updates; the adapter learns around the fixed
side-path perturbation.
\end{table}

\paragraph{Evaluation.}
Four STEM/code benchmarks: \bench{GPQA Diamond} (198 items, 0-shot
MCQ)~\citep{rein2023gpqa}, \bench{GSM8K} (1{,}319 items, 8-shot
CoT)~\citep{cobbe2021training}, \bench{MATH-500} (500 items, 4-shot
CoT)~\citep{hendrycks2021math}, and \bench{MBPP} (500 items, 3-shot
code generation)~\citep{austin2021mbpp}. Inference uses greedy
decoding, maximum 8{,}192 new tokens, deterministic SGLang backend on
$2\times$H100 80GB. Scoring uses lm-evaluation-harness 0.4.8 with
\texttt{math\_verify}~0.9.0; as a robustness check the four audited
configurations are additionally re-scored under four alternative
methodologies with consistent results. Full extractor logic, scoring
details, and statistical tests are in Appendices~\ref{app:protocol}
and~\ref{app:stat-derivation}.

\paragraph{Seed handling.}
Headline numbers use the seed at which each configuration was first
trained (\texttt{seed=42} for Q32B, DS32B, Q8B; \texttt{seed=46} for
MoE), following standard PEFT-paper practice. Full seed-handling
details and a matched-initialization multi-seed audit are in
Appendices~\ref{app:protocol} and~\ref{app:matched}.

\paragraph{Notation.}
\emph{Full non-regression} denotes non-negative deltas on all four
cells of a backbone (per-backbone) or all 16 cells of the suite
(per-cell).

\subsection{Main results}
\label{sec:headline}

\paragraph{Overall improvements.}
Tables~\ref{tab:main} and~\ref{tab:main-dora} summarize the
backbone-level effects under LoRA and DoRA. In the LoRA pipeline,
\talan{} is the only compared method positive on every backbone, with
a cross-model mean of $+1.41$ percentage points (pp) over the matched
LoRA baseline. Detailed statistical analyses are reported in
Appendix~\ref{app:stat-derivation}.

\begin{table*}[t]
\centering
\caption{\textbf{Head-to-head under LoRA.}
$\Delta$ is measured against each backbone's matched LoRA baseline.
\talan{} is the only compared method positive on every backbone.}
\label{tab:main}
\small
\begin{tabular*}{\textwidth}{@{\extracolsep{\fill}}lcccccc@{}}
\toprule
Method & Q32B $\Delta$ & DS32B $\Delta$ & Q8B $\Delta$ & MoE $\Delta$ & Mean $\Delta$ & Positive on all? \\
\midrule
LoRA baseline (matched)       &  0.00 &  0.00 &  0.00 &  0.00 &  0.00 & --- \\
RepE                          & $-$0.03 & $-$0.45 & $+$1.76 & $-$24.15 & $-$5.72 & No \\
CogSteer                      & $-$0.07 & $+$0.05 & $+$0.71 & $-$0.70 & $0.00$ & No \\
DELIFT                        & $-$0.11 & $+$1.07 & $+$1.79 & $+$0.44 & $+$0.80 & No \\
DoRA-only                     & $-$0.03 & $+$1.27 & $+$0.83 & $+$0.02 & $+$0.52 & No \\
AdapterFusion                 & $-$1.52 & $+$2.71 & $+$2.90 & $-$0.56 & $+$0.88 & No \\
\textbf{\talan{}+LoRA (ours)} & \textbf{$+$0.43} & \textbf{$+$1.49} & \textbf{$+$2.67} & \textbf{$+$1.04} & \textbf{$+$1.41} & \textbf{Yes} \\
\bottomrule
\end{tabular*}
\end{table*}

The backbone-level summary in Table~\ref{tab:main} understates the
strongest non-regression property of the LoRA pipeline: the same
audited configurations are non-negative on all 16 model--benchmark
cells (Table~\ref{tab:benchdeltas}). Other compared methods sometimes
produce competitive average gains, but each regresses on at least one
backbone or benchmark cell; \talan{}+LoRA is the only method in this
comparison that is positive on every backbone and preserves every
cell. The corresponding sign-test, McNemar, and bootstrap analyses are
reported in Appendix~\ref{app:stat-derivation}. Because the audited
configurations are selected from a configuration-space audit, we treat
the 16-of-16 sign pattern as a non-regression profile of the selected
operating points rather than as a post-selection-free guarantee.

The same intervention family remains positive under DoRA
(Table~\ref{tab:main-dora}), with a cross-model mean of $+1.85$ pp
and positive mean effects on all four backbones. This supports the
orthogonality claim from Section~\ref{sec:theory}: the \talan{} side
path is not tied to LoRA-specific parameters. The detailed cell-level
profile is not identical to LoRA: DoRA has positive deltas on 13 of
16 cells, as shown in Table~\ref{tab:benchdeltas}; seed-level and
matched-init details are reported in Appendix~\ref{app:matched}.

\begin{table*}[t]
\centering
\caption{\textbf{\talan{} under DoRA.}
$\Delta$ is measured against each backbone's matched DoRA baseline.
\talan{}+DoRA is positive on every backbone.}
\label{tab:main-dora}
\small
\begin{tabular*}{\textwidth}{@{\extracolsep{\fill}}lccccc@{}}
\toprule
Method & Q32B $\Delta$ & DS32B $\Delta$ & Q8B $\Delta$ & MoE $\Delta$ & Mean $\Delta$ \\
\midrule
DoRA baseline (matched)       &  0.00 &  0.00 &  0.00 &  0.00 &  0.00 \\
\textbf{\talan{}+DoRA (ours)} & \textbf{$+$1.96} & \textbf{$+$1.33} & \textbf{$+$1.07} & \textbf{$+$3.05} & \textbf{$+$1.85} \\
\bottomrule
\end{tabular*}
\end{table*}

\paragraph{Seed sensitivity checks.}
We use matched-initialization multi-seed runs as sensitivity checks
rather than as additional headline tables. Aggregated paired deltas
remain positive under both adapter families (LoRA: $+0.50$ pp over
55 paired deltas; DoRA: $+0.47$ pp over 41 paired deltas), while the
seed-level spread confirms that individual cells can vary. We therefore
emphasize the selected operating-point profile and family-level trends
rather than claiming that every individual run is seed-invariant. Full
protocol details are in Appendix~\ref{app:matched}.

\paragraph{Detailed per-benchmark results.}
Table~\ref{tab:benchdeltas} gives the benchmark-level view behind the
overall deltas. \bench{GPQA Diamond} is the most consistent source of
improvement under both adapters. LoRA preserves \bench{MBPP} better
across backbones, while DoRA gives larger \bench{MATH-500} gains on
Q32B and MoE. Thus, DoRA supports cross-adapter composition at the
backbone-average level, but it does not match LoRA's full per-cell
non-regression profile; the full cross-method per-cell audit is in
Appendix~\ref{app:cross-method}.

\begin{table}[!ht]
\centering
\caption{\textbf{Per-benchmark deltas} (pp) under both adapter
variants.}
\label{tab:benchdeltas}
\small
\begin{tabular*}{\columnwidth}{@{\extracolsep{\fill}}lcccc@{}}
\toprule
Model & GPQA $\Delta$ & GSM8K $\Delta$ & MATH $\Delta$ & MBPP $\Delta$ \\
\midrule
\multicolumn{5}{l}{\emph{With LoRA}} \\
Qwen3-32B       & $+$1.01 & $+$0.30 & $+$0.00 & $+$0.40 \\
DeepSeek-32B    & $+$4.04 & $+$0.30 & $+$0.00 & $+$1.60 \\
Qwen3-8B        & $+$8.08 & $+$0.00 & $+$1.60 & $+$1.00 \\
Qwen3-30B-A3B   & $+$1.01 & $+$0.15 & $+$3.00 & $+$0.00 \\
\textit{Cross-model mean} & \textit{$+$3.54} & \textit{$+$0.19} & \textit{$+$1.15} & \textit{$+$0.75} \\
\midrule
\multicolumn{5}{l}{\emph{With DoRA}} \\
Qwen3-32B       & $+$2.53 & $+$0.91 & $+$5.20 & $-$0.80 \\
DeepSeek-32B    & $+$1.01 & $+$1.52 & $+$2.00 & $+$0.80 \\
Qwen3-8B        & $+$3.03 & $+$0.45 & $+$1.20 & $-$0.40 \\
Qwen3-30B-A3B   & $+$6.06 & $-$0.08 & $+$6.00 & $+$0.20 \\
\textit{Cross-model mean} & \textit{$+$3.16} & \textit{$+$0.70} & \textit{$+$3.60} & \textit{$-$0.05} \\
\bottomrule
\end{tabular*}
\end{table}

\paragraph{Selected non-Qwen transfer check.}
\label{sec:nonqwen-transfer}
The main suite intentionally fixes the backbone family so that
configuration changes can be audited tightly. To check whether the
effect is not confined to Qwen-family hosts, we additionally evaluate
selected authoritative paired \talan{}+baseline runs on
Llama-3.2-1B using the same four benchmark suite. Table~\ref{tab:nonqwen}
reports only the paired settings that are positive on average across
seven seeds; noisy or unpaired exploratory runs are not used as claims.
The standard-deviation column is included to make the seed variance
visible; the result should be read as a targeted transfer probe, not as
an exhaustive Gemma/Mistral/Llama-family sweep.

\begin{table}[!ht]
\centering
\caption{\textbf{Selected non-Qwen paired transfer check on
Llama-3.2-1B.} $\Delta$ is measured against the matched PEFT baseline
for the same seed and adapter variant.}
\label{tab:nonqwen}
\small
\setlength{\tabcolsep}{4pt}
\resizebox{\columnwidth}{!}{%
\begin{tabular}{lccccc}
\toprule
Backbone / adapter & Seeds & Positive seeds & Mean $\Delta$ & Std. $\Delta$ & GPQA / GSM8K / MATH / MBPP mean $\Delta$ \\
\midrule
Llama-3.2-1B + LoRA   & 7 & 5/7 & $+$1.34 & 3.04 & $+$1.37 / $+$0.08 / $+$1.26 / $+$2.66 \\
Llama-3.2-1B + rsLoRA & 7 & 5/7 & $+$0.60 & 0.58 & $+$1.95 / $+$0.27 / $+$0.20 / $+$0.00 \\
\bottomrule
\end{tabular}%
}
\end{table}

\subsection{Practical cost}
\label{sec:cost}

\talan{} adds $<1\%$ trainable parameters relative to the backbone and
keeps training and inference overhead small relative to matched LoRA
(Table~\ref{tab:cost}). Unlike LoRA weights, the \talan{} side path
cannot be merged away; it remains an active forward-pass component at
inference time.

\begin{table}[!ht]
\centering
\caption{\textbf{Practical cost} relative to matched LoRA.}
\label{tab:cost}
\small
\begin{tabular}{lcccc}
\toprule
Model & Trainable params (LoRA / TALAN) & Train time & Inference & Memory peak \\
\midrule
Q32B   & 52M / 17M (0.32\%) & $1.04\times$ & $1.02\times$ & $+1.4$ GB \\
DS32B  & 52M /  6M (0.11\%) & $1.02\times$ & $1.01\times$ & $+0.6$ GB \\
Q8B    & 30M /  9M (0.30\%) & $1.03\times$ & $1.02\times$ & $+0.8$ GB \\
MoE    & 49M / 12M (0.20\%) & $1.01\times$ & $1.01\times$ & $+0.7$ GB \\
\bottomrule
\end{tabular}
\end{table}

For comparison, AdapterFusion requires multiple adapter trainings plus
a fusion stage, and incurs larger inference overhead because multiple
adapter paths are active. \talan{} is not zero-overhead, but it keeps
the additional cost small relative to the matched adapter baseline.

\section{Analysis and Discussion}
\label{sec:mechanism}

\noindent
Section~\ref{sec:experiments} establishes the outcome-level pattern:
\talan{} improves matched adapter baselines on the tested backbones,
with the strongest non-regression profile under LoRA and positive mean
effects under DoRA. Here we
focus on the mechanism-level evidence tied directly to the two
architectural properties in Section~\ref{sec:theory}. First, we
measure whether the \talan{} perturbation is both small (P1) and
complementary to the adapter update (P2). Second, we ask whether such
a small perturbation can still propagate through the host when inserted
at an effective residual-stream location. The goal is not to provide a
complete mechanism or a predictive configuration rule, but to check
whether the observed behavior is consistent with the proposed small
activation-intervention view.

\subsection{Orthogonality: empirical confirmation}
\label{sec:insight-orthogonality}

P1 and P2 make two related predictions: \talan{} should introduce a
small residual-stream perturbation, and that perturbation should not
simply become a richer LoRA update. We test both on a synthetic Gaussian probe by measuring the norm
ratio and per-token cosine between $\Delta_{\mathrm{TALAN}}$ and
$\Delta_{\mathrm{LoRA}}$ at the insertion layer.
Table~\ref{tab:lora-subspace} shows that the LoRA perturbation is
$80$--$1{,}700\times$ larger in norm, consistent with P1's view of
\talan{} as a small activation intervention. At the same time, the
cosine is near zero on both winning and losing configurations,
supporting P2: \talan{} contributes a complementary residual-stream
direction rather than competing with the adapter in the same low-rank
subspace.

\begin{table}[!ht]
\centering
\caption{\textbf{Perturbation scale and adapter-subspace orthogonality.}
\talan{} is much smaller than the matched LoRA update while maintaining
near-zero cosine on both winners and losers, suggesting that
orthogonality is architectural rather than outcome-driven.}
\label{tab:lora-subspace}
\small
\begin{tabular}{lccc}
\toprule
Backbone / variant & $\|\Delta_{\mathrm{LoRA}}\|/\|\Delta_{\mathrm{TALAN}}\|$ & cosine & Std \\
\midrule
Q8B headline (winner)   & $81.2$       & $-0.0000$ & $0.017$ \\
Q8B alternate (loser)   & $82.8$       & $+0.0024$ & $0.017$ \\
DS32B headline (winner) & $816.4$      & $-0.0008$ & $0.014$ \\
DS32B alternate (loser) & $1{,}776.4$  & $-0.0009$ & $0.014$ \\
\bottomrule
\end{tabular}
\end{table}

\paragraph{Real-token corroboration.}
The synthetic probe is corroborated on natural-language tokens from
four domains (math, code, STEM, general). On three of four backbones
(Q8B, Q32B, MoE), the mean \talan{} perturbation directions across
domains have off-diagonal cosine $\geq 0.62$, suggesting a
predominantly \emph{domain-invariant} bias direction rather than a
per-domain router. DS32B is the exception ($+0.02$ to $+0.08$),
suggesting its winning geometry routes more selectively. Both
patterns are consistent with the orthogonality of
Table~\ref{tab:lora-subspace}. The full breakdown is in
Appendix~\ref{app:weights}.

\subsection{Activation amplification through depth}
\label{sec:insight-design-language}

Table~\ref{tab:lora-subspace} shows that \talan{}'s direct
perturbation is small relative to the adapter update. The next question
is whether such a small perturbation can still become consequential.
If the insertion layer is well chosen, a local residual-stream
perturbation may be amplified by subsequent transformer layers. This
analysis therefore probes the main intuition behind P1: small
perturbations can be sufficient for steering downstream activations when
they are introduced at an effective location.

\begin{figure}[!ht]
\centering
\begin{tabular}{cc}
\includegraphics[width=0.46\textwidth]{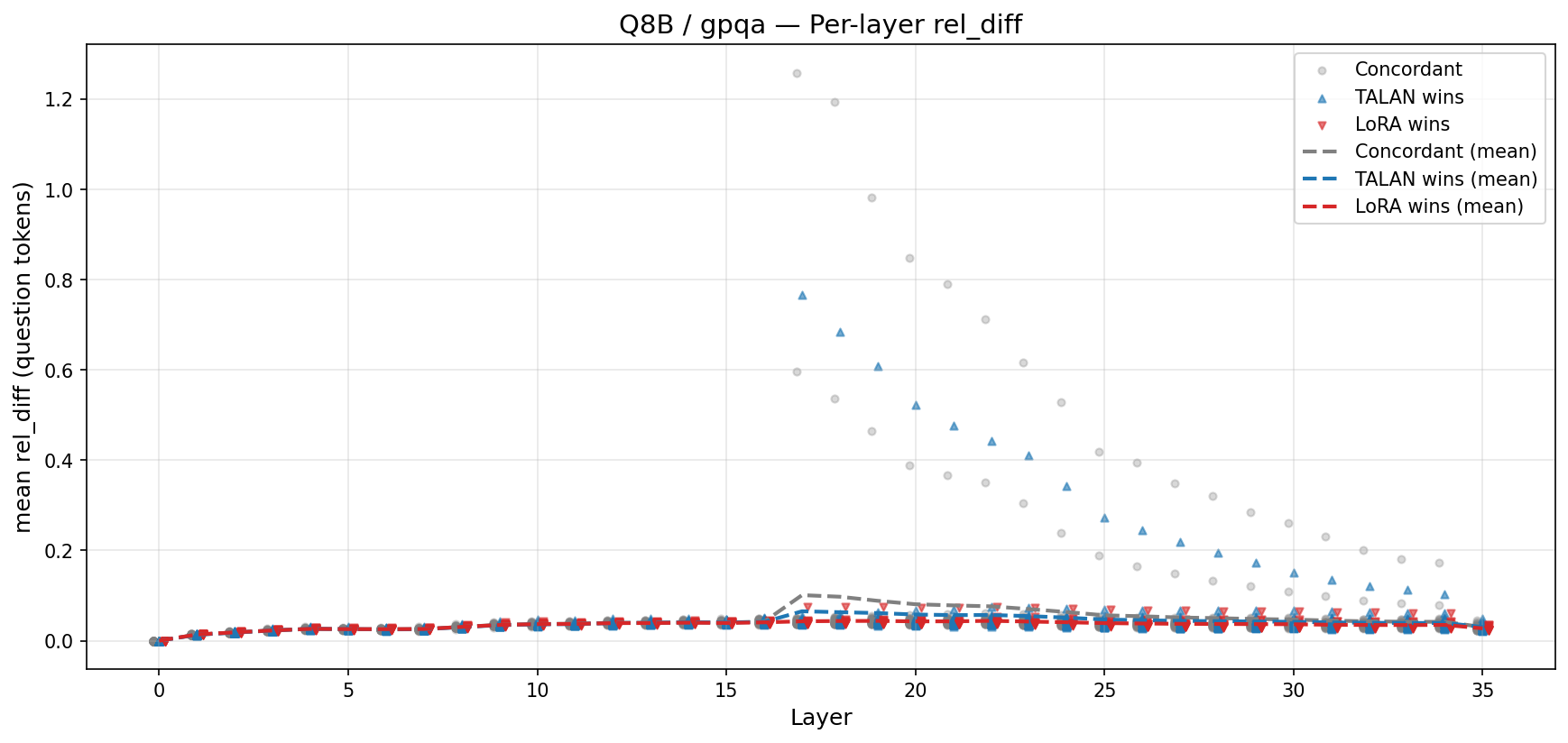} &
\includegraphics[width=0.46\textwidth]{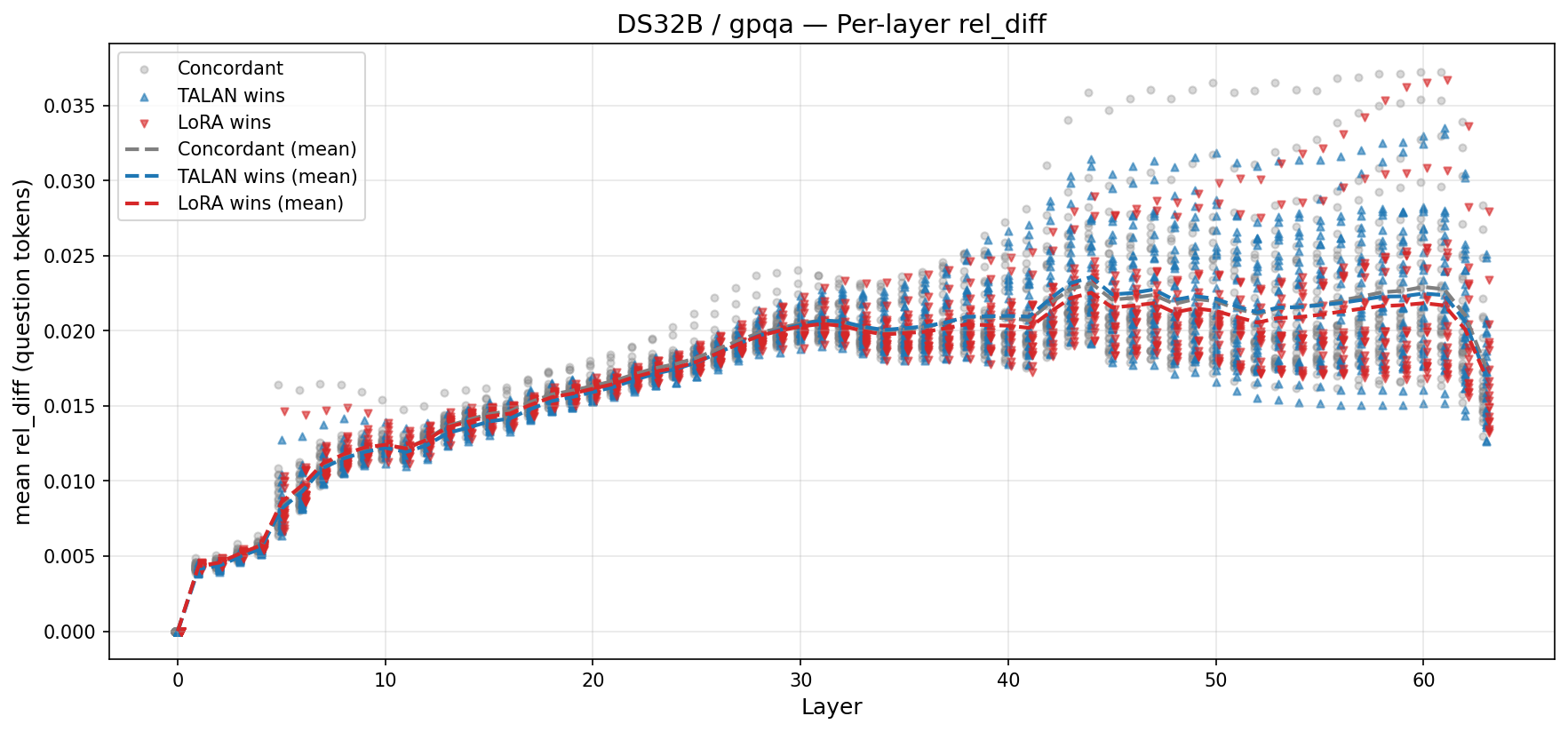} \\
\includegraphics[width=0.46\textwidth]{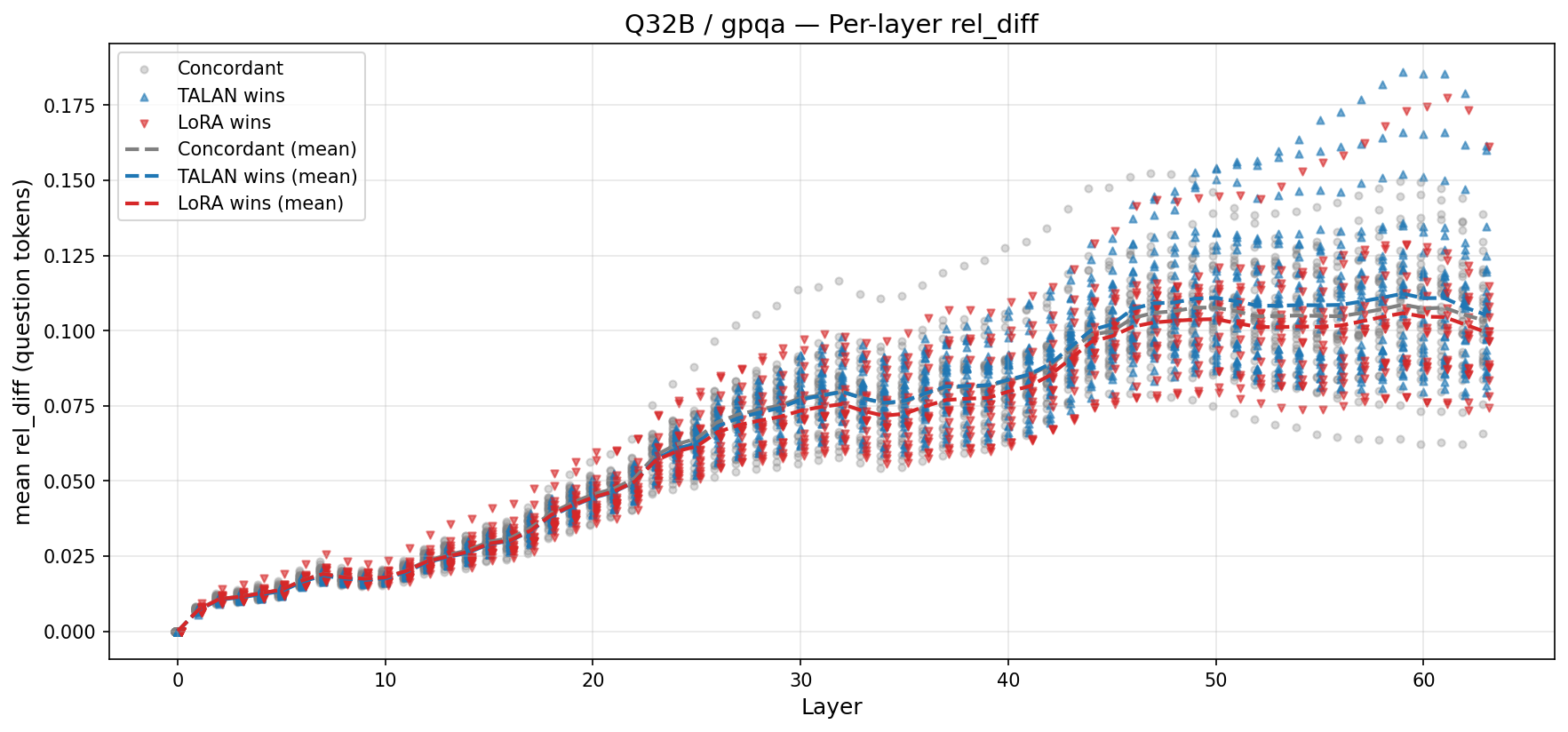} &
\includegraphics[width=0.46\textwidth]{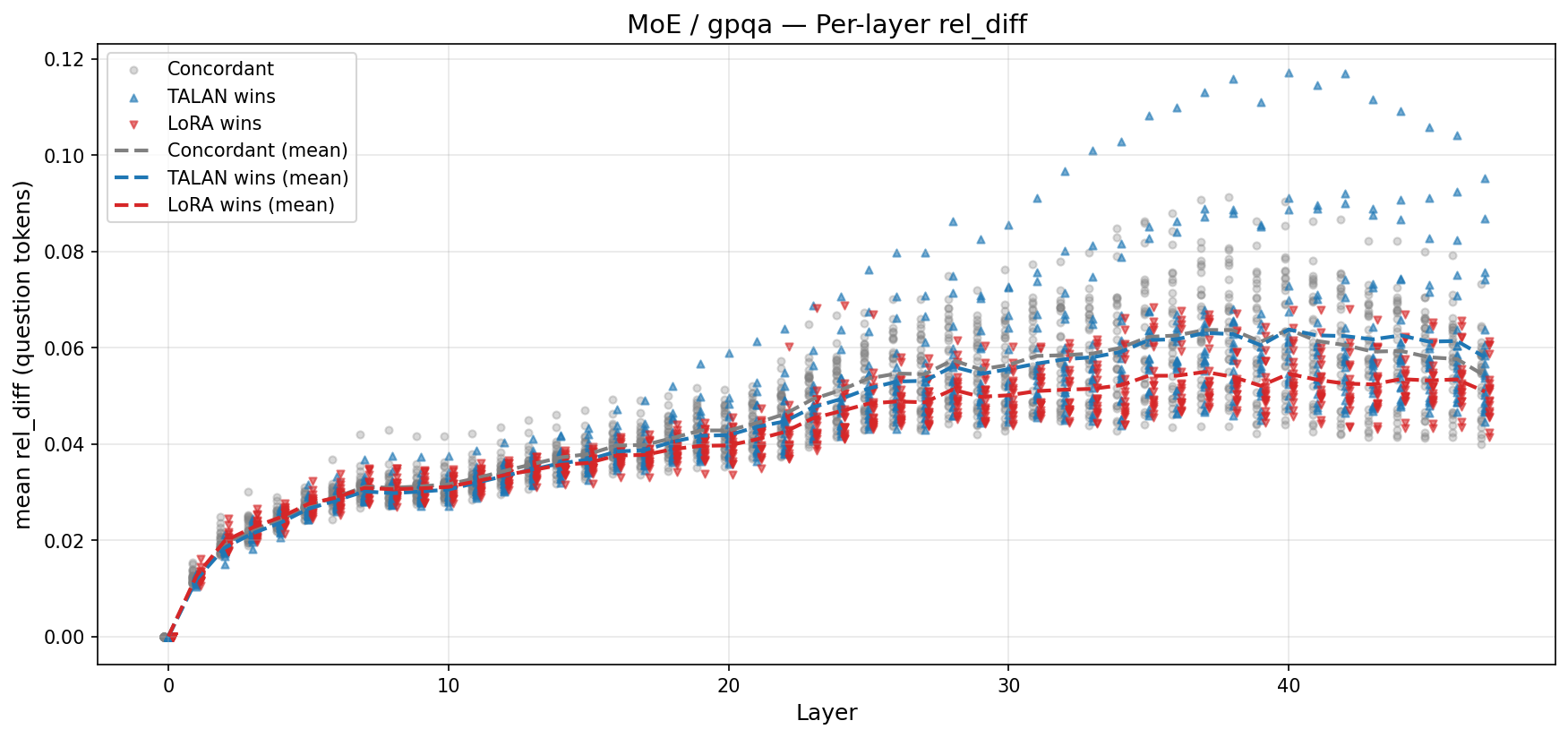} \\
\end{tabular}
\caption{\textbf{Per-layer activation divergence on \bench{GPQA Diamond}.}
Each point is one example; the vertical axis is relative activation
difference between \talan{}+LoRA and matched LoRA. The selected
insertion layer and later-layer amplification are visible across
backbones.}
\label{fig:per-layer-headline}
\end{figure}

\paragraph{Per-layer activation divergence.}
We compare activations at every transformer layer between matched LoRA,
\talan{}+LoRA, and a control with \talan{} hooks disabled. Across
backbones, the same qualitative pattern appears: a sharp divergence at
the selected insertion layer $\ell^\star$, amplification through later
layers, and backbone-specific shapes that are stable across prompt
distributions. Figure~\ref{fig:per-layer-headline} shows the GPQA
case. This is important because it links the smallness result in
Table~\ref{tab:lora-subspace} to task-level behavior: the intervention
does not need to be large at the writeback point if the host amplifies
it through later layers. In this sense, a small perturbation placed at
the right residual-stream location can be sufficient to steer
downstream activations enough to affect performance.

The insertion layers differ across hosts---Q8B at layer~4, Q32B at
layer~12, and DS32B and MoE at the embedding layer---but the lift-off
point in the activation curves tracks the configured insertion point in
each case. The full per-benchmark activation plots in
Appendix~\ref{app:activation} show the same qualitative behavior across
the four evaluated benchmarks. This host-specificity is consistent with
the configuration-space view of Section~\ref{sec:method-axes}: the
same \talan{} block is reused, but the effective insertion location and
writeback regime are selected per host.

\section{Conclusion}
\label{sec:conclusion}

We introduced \talan{}, a sequence-conditioned latent adaptation
architecture that inserts a summarize--remix--writeback side path into
the residual stream and co-trains it with a low-rank adapter. The
central design choice is to make the intervention small and
complementary: \talan{} contributes a bounded activation perturbation
that is architecturally separate from the LoRA/DoRA adapter update,
while the six-axis configuration space exposes where and how that
perturbation enters the host computation.

Across four Qwen3-family backbones and four STEM/code benchmarks,
selected \talan{} operating points improve matched adapter baselines
under both LoRA and DoRA. Under LoRA, the audited configurations reach
a cross-model mean gain of $+1.41$ percentage points and are
non-negative on all 16 model--benchmark cells. Under DoRA, \talan{}
reaches a cross-model mean gain of $+1.85$ percentage points and is
positive on all four backbones. A matched-initialization multi-seed
audit supports positive family-level effects while quantifying
seed-level variance (Appendix~\ref{app:matched}), and the added computational cost
remains small: less than
$1\%$ trainable parameters relative to the backbone and
$1.01$--$1.02\times$ inference overhead relative to matched LoRA.
A selected Llama-3.2-1B transfer check is positive under both LoRA and
rsLoRA over seven paired seeds, suggesting that the intervention can
transfer beyond the primary Qwen-family suite, though broader
non-Qwen coverage remains future work.

The analysis supports the small-complementary-perturbation view. The
adapter update is much larger in norm, yet its direction remains nearly
orthogonal to the \talan{} perturbation; meanwhile, per-layer activation
measurements show that a small local perturbation can propagate and
amplify through later layers after insertion. These findings suggest
that activation interventions do not need to dominate update magnitude
to affect model behavior. A useful intervention can instead occupy a
complementary direction and rely on the host's subsequent computation to
propagate its effect.

\paragraph{Limitations and scope.}
Our evaluation focuses on STEM and code benchmarks
(\bench{GPQA Diamond}, \bench{GSM8K}, \bench{MATH-500}, and
\bench{MBPP}); dialog, retrieval-grounded generation, multi-turn
instruction following, and non-English settings remain untested. The
primary suite uses Qwen-family hosts; the selected Llama-3.2-1B results
provide an initial non-Qwen transfer check, but transfer to broader
families such as Gemma or Mistral is not established here. We also test
one main LoRA recipe and one main DoRA recipe with matched rank, scale,
and target modules; broader adapter ranks, target modules, and larger
training mixtures may change the selected configurations. The MoE
operating point is a forward-active frozen-perturbation variant rather
than the same trained dense-backbone side path, so we treat it as a
related structural-prior finding. Finally, the
mechanistic analysis is descriptive: it shows that successful
configurations leave activation-level signatures consistent with the
proposed mechanism, but it does not yet provide a prescriptive rule for
selecting \talan{} configurations from backbone geometry alone.

Overall, \talan{} provides a compact platform for studying
sequence-conditioned activation modification inside standard
adapter-based SFT. Its results support a practical path for composing
small residual-stream interventions with low-rank adaptation and move
toward steerable activation-level adaptation, where perturbation
location, geometry, and magnitude are selected from the model's
internal structure rather than by undirected search.

\bibliographystyle{plainnat}
\bibliography{merge_ref}

\appendix
\section{Appendix Guide}

\paragraph{Organization.}
The appendix contains protocol details, configuration records,
statistical derivations, activation and weight analyses, qualitative
case studies, and matched-initialization audits. We group it into five
blocks: protocol and audited configurations
(Appendices~\ref{app:protocol}--\ref{app:operating-points});
cross-method and configuration-space result audits
(Appendices~\ref{app:cross-method}--\ref{app:search}); statistical and
mechanistic checks (Appendices~\ref{app:stat-derivation}--\ref{app:weights});
extended qualitative and matched-initialization analyses
(Appendices~\ref{app:cases}--\ref{app:matched}); workflow comparison
(Appendix~\ref{app:workflow}); and selected non-Qwen transfer evidence
(Appendix~\ref{app:nonqwen}).
Several analyses kept out of the main text for compactness---including
the sign-test derivation, matched-initialization audit, weight-level
audit, and extended case studies---are retained here for transparency.

\begin{itemize}[nosep,leftmargin=*]
  \item Appendix~\ref{app:protocol}: Training and evaluation protocol details (data, seed handling, eval extractor logic, hardware).
  \item Appendix~\ref{app:operating-points}: Audited per-backbone configuration fields.
  \item Appendix~\ref{app:cross-method}: Per-benchmark detail for cross-method comparison.
  \item Appendix~\ref{app:search}: Configuration-space audit and full-non-regression rarity (full per-axis breakdowns of the 2{,}590-configuration architecture audit).
  \item Appendix~\ref{app:stat-derivation}: Statistical-test derivations (sign, McNemar, bootstrap) and assumption discussion.
  \item Appendix~\ref{app:activation}: Full per-layer activation divergence plots for all four backbones $\times$ four benchmarks.
  \item Appendix~\ref{app:weights}: Complete weight-level interpretability audit (drift, recurring discriminators, signature--vs--$\Delta$ correlation).
  \item Appendix~\ref{app:cases}: Extended case-study transcripts with per-token divergence panels.
  \item Appendix~\ref{app:matched}: Compact cross-adapter \textsc{matched-init} multi-seed sensitivity analysis.
  \item Appendix~\ref{app:workflow}: Single-stage \talan{} vs.\ multi-stage AdapterFusion workflow comparison (Figure~\ref{fig:workflow}).
  \item Appendix~\ref{app:nonqwen}: Selected non-Qwen paired transfer check.
\end{itemize}

\section{Training and Evaluation Protocol}
\label{app:protocol}

\paragraph{Training data.}
The training mixture comprises 5{,}000 ChatML-formatted SFT examples.
Composition: math 1{,}729 examples
(MathInstruct/OpenMathInstruct subsets aligned to MATH-500 and GSM8K
problem styles---no test-set leakage; the items are drawn from the
public training splits and are disjoint from the test items used for
evaluation); code 1{,}592 examples (CodeAlpaca, MBPP-style prompts that
are \emph{not} the held-out 500 MBPP test items); STEM 974 examples
(GPQA-style multiple-choice and explanatory passages drawn from the
public training questions, disjoint from the 198 \bench{GPQA Diamond} items);
general 705 examples (open-ended assistant turns from filtered
ShareGPT). The mixture is shuffled with a fixed seed once and used
identically for every backbone and every method (\talan{}, LoRA-only,
DoRA, AdapterFusion, RepE, CogSteer, DELIFT).

\paragraph{Training schedule.}
For every backbone and every method: 78 maximum optimization steps,
AdamW optimizer with cosine schedule, 10\% warmup ratio, peak learning
rate $2 \times 10^{-4}$ (Q32B, DS32B, Q8B) or $2.5 \times 10^{-5}$
(MoE), weight decay $0$ (Q32B, DS32B, Q8B) or $0.02$ (MoE), per-device
batch size 4, gradient accumulation 8 (effective batch 64), gradient
checkpointing on, BF16 mixed precision, DeepSpeed ZeRO-2 on
$2 \times \text{H100 80GB}$, gradient clipping 1.0. The matched LoRA
baseline within each backbone uses \emph{the same LR and WD} as its
\talan{} counterpart, and all controllable training-side
hyperparameters are matched within each audited pair.

\paragraph{Seed handling (full disclosure).}
We report the audited configuration at the seed at which the
configuration was first trained: \texttt{seed=42} for Q32B, DS32B, and
Q8B; \texttt{seed=46} for MoE. The matched LoRA baseline within each
backbone uses the same seed value. HuggingFace Trainer's \texttt{seed}
and \texttt{data\_seed} arguments are set explicitly;
\texttt{torch.manual\_seed} and \texttt{torch.cuda.manual\_seed\_all}
are called before module construction;
\texttt{dataset.shuffle(seed=...)} fixes the shuffle. Operating-point
selection (hereafter \emph{champion}) within each backbone is by best per-cell full non-regression at the audited seed,
following the standard PEFT-paper protocol. The broader
2{,}590-configuration audit (Appendix~\ref{app:search}) is not a
search for the best single configuration but a systematic exploration
of which regions of the six-axis configuration space yield
non-regressing configurations.

\paragraph{Inference settings.}
Greedy decoding (temperature $0$), maximum 8{,}192 new tokens,
deterministic SGLang inference backend, tensor parallelism 2 on
$2 \times \text{H100 80GB}$. For DeepSeek-R1-Distill we set
\texttt{--enable-deterministic-inference} to ensure deterministic
inference. Prompts use the model's native chat template
(\verb!<|im_start|>!/\verb!<|im_end|>! for the Qwen3 family;
\verb!<|begin_of_sentence|>!/\verb!<|end_of_sentence|>! for DeepSeek).
Few-shot budgets: GPQA 0-shot, GSM8K 8-shot CoT, MATH-500 4-shot CoT,
MBPP 3-shot.

\paragraph{Scoring (full disclosure of extractor logic).}
Canonical lm-evaluation-harness 0.4.8 with \texttt{math\_verify}~0.9.0:
\begin{itemize}[nosep,leftmargin=*]
  \item \bench{GPQA Diamond}: \texttt{MultiChoiceRegexFilter}, a 3-step
    regex cascade. Step 1 matches the regex
    \texttt{\textbackslash([A-D]\textbackslash)} (a parenthesized letter
    in the response). If no match, Step 2 falls back to choice-text
    matching against the four answer choices. If still no match, Step 3
    matches \texttt{:[\textbackslash s]*([A-D])} (a letter following a
    colon and optional whitespace). The first match wins; no match $=$
    incorrect.
  \item \bench{GSM8K}: \texttt{gsm8k\_cot} flexible-extract regex. The
    regex matches the answer following ``\#\#\#\#'' or, falling back, the
    last number in the response.
  \item \bench{MATH-500}: \texttt{minerva\_math.math\_verify}. The
    extractor parses the boxed answer expression, sympifies both gold
    and candidate, and \texttt{math\_verify} performs symbolic
    equivalence checking (handles equivalent forms such as $1/2$
    vs.\ $0.5$ vs.\ $\frac{1}{2}$).
  \item \bench{MBPP}: execution-sandbox \texttt{pass@1}. The extracted
    code block is run against the held-out test cases in a sandbox; the
    item is correct iff all assertions pass and no exception is raised.
\end{itemize}
This canonical scorer is the basis for the headline numbers
($+1.41$ pp / 4-of-4 / 16-of-16); statistical sanity checks are
reported in Appendix~\ref{app:stat-derivation}.

\paragraph{Per-cell absolute scores.}
Table~\ref{tab:absolute-scores} reports the absolute cell scores (TALAN
and matched LoRA, not just deltas) for completeness.

\begin{table}[!ht]
\centering
\caption{\textbf{Per-cell absolute scores} of \talan{}+LoRA / matched LoRA
under lm-eval-harness with \texttt{math\_verify}~0.9.0.
$\Delta$ = TALAN $-$ LoRA on each cell.}
\label{tab:absolute-scores}
\small
\begin{tabular}{lcccc}
\toprule
Backbone & GPQA & GSM8K & MATH-500 & MBPP \\
\midrule
Q32B  & $27.27 / 26.26$ & $91.36 / 91.05$ & $47.40 / 47.40$ & $74.20 / 73.80$ \\
DS32B & $39.90 / 35.86$ & $88.25 / 87.95$ & $61.80 / 61.80$ & $72.00 / 70.40$ \\
Q8B   & $31.31 / 23.23$ & $85.90 / 85.90$ & $60.80 / 59.20$ & $65.20 / 64.20$ \\
MoE   & $42.42 / 41.41$ & $94.24 / 94.09$ & $60.60 / 57.60$ & $77.00 / 77.00$ \\
\bottomrule
\end{tabular}
\end{table}

\paragraph{Training-side apple-to-apple guarantee.}
For each (TALAN champion, matched LoRA baseline) pair we performed a
training-side hyperparameter audit: zero red-flag mismatches across
all four pairs. All controllable
training-side dimensions (LoRA $r/\alpha$/targets/dropout/bias/task,
LR, weight\_decay, batch\_size, gradient\_accumulation, max\_steps,
bf16, gradient\_checkpointing, optim, scheduler, warmup, num\_workers,
dataloader, max\_length, truncation, num\_proc) are identical within
each pair. The only difference is the \talan{} module $+$ post-PEFT
re-enable $+$ \talan{} callback (per the six configuration axes).

\section{Audited Configurations}
\label{app:operating-points}

The four audited configuration records are listed below.
\texttt{seed=42} for the first three; \texttt{seed=46} for MoE. The
field \texttt{n\_themes} corresponds to the slot count $T$ described in
Section~\ref{sec:method-block}, and saved-weight tensor names prefixed
\texttt{theme\_} refer to the slot queries $Q$.

\paragraph{Q32B (Qwen3-32B): the headline configuration.}
\begin{verbatim}
{
  "type": "talan", "clustering_method": "attention",
  "n_themes": 2, "network_type": "multihead_local",
  "integration_type": "learned_gate", "gate_type": "vector",
  "alpha": 0.008, "freeze_steps": 0,
  "connection_point": "layer_12", "hidden_size": 5120
}
\end{verbatim}
\textbf{Trainability scope $\tau$}: full re-enable post-PEFT
(\texttt{if 'talan' in n: requires\_grad=True}).
\textbf{Gradient scale $\gamma$}: 500
(callback boosts non-($\alpha$|gate) \talan{} params by $500\times$
before optimizer step).

\paragraph{DS32B (DeepSeek-R1-Distill-Qwen-32B): the headline configuration.}
\begin{verbatim}
{
  "type": "talan", "clustering_method": "attention",
  "n_themes": 2, "network_type": "cross_attn",
  "integration_type": "learned_gate", "gate_type": "vector",
  "alpha": 0.001, "freeze_steps": 0,
  "connection_point": "embedding", "hidden_size": 5120
}
\end{verbatim}
\textbf{Trainability scope $\tau$}: full re-enable post-PEFT.
\textbf{Gradient scale $\gamma$}: 500.

\paragraph{Q8B (Qwen3-8B): the headline configuration.}
\begin{verbatim}
{
  "type": "talan", "clustering_method": "attention",
  "n_themes": 2, "network_type": "per_token_mix",
  "integration_type": "learned_gate", "gate_type": "vector",
  "alpha": 0.0015, "freeze_steps": 0,
  "connection_point": "layer_4", "hidden_size": 4096
}
\end{verbatim}
\textbf{Trainability scope $\tau$}: full re-enable post-PEFT, then
\texttt{activate\_steps=20} callback freezes \talan{} params at step
20 (TALAN trainable for steps $0$--$19$, frozen for steps $20$--$77$).
\textbf{Gradient scale $\gamma$}: 1000 (highest of any backbone).

\paragraph{MoE (Qwen3-30B-A3B): the headline configuration
\textnormal{---} frozen-perturbation variant.}
\textbf{Note:} Unlike the three dense-backbone configurations above,
the MoE configuration uses a \emph{frozen-perturbation} variant in which
\talan{} is randomly initialized and receives no gradient updates.
The LoRA adapter trains around the fixed side-path perturbation rather
than co-training with a learned intervention. This constitutes a
mechanistically distinct setting from the trained \talan{} intervention
on Q32B, DS32B, and Q8B; we report it separately as evidence that
adapter-compatible residual-stream perturbations need not be learned to
be useful (see Section~\ref{sec:trainability-axis-mechanism} below for
analysis).
\begin{verbatim}
{
  "type": "talan", "clustering_method": "soft_mos",
  "n_themes": 6, "network_type": "multihead_local",
  "integration_type": "learned_gate", "gate_type": "scalar",
  "alpha": 0.008, "freeze_steps": 35,
  "connection_point": "embedding", "hidden_size": 2048
}
\end{verbatim}
\textbf{Trainability scope $\tau$}: forward-active (\talan{} module
participates in the forward pass at random initialization but receives
no gradient updates throughout training; intentional design choice).
\textbf{Gradient scale $\gamma$}: --- (not applicable; no \talan{}
gradients to scale).
\textbf{\texttt{freeze\_steps=35}}: \talan{}'s forward output is masked
to identity for the first $35$ of $78$ steps (LoRA-only training during
warm-up); \talan{}'s residual contribution becomes active from step $36$
onward.
\textbf{\texttt{clustering\_method=soft\_mos}}: the MoE backbone uses a
soft mixture-of-slots summarization variant instead of the
attention-based soft assignment described in
Section~\ref{sec:method-block}; slots are combined via a learned softmax
mixture rather than per-token dot-product attention.

\section{Cross-Method Per-Benchmark Detail}
\label{app:cross-method}

For each compared method (\talan{}, LoRA-only, DoRA, AdapterFusion,
RepE, CogSteer, DELIFT), we report the per-cell delta versus the
matched LoRA baseline of the corresponding backbone, under canonical
lm-eval scoring (Table~\ref{tab:cross-method-detail}). All methods use
the same training data, the same LoRA recipe (where applicable), the
same evaluation protocol, and the same seed within each backbone.

\begin{table}[!ht]
\centering
\caption{\textbf{Cross-method per-cell deltas} (pp) versus matched LoRA,
lm-eval-harness with \texttt{math\_verify}~0.9.0. Rows
are methods, columns are (model, benchmark) cells. Methods are ordered
by non-regression count (last column).}
\label{tab:cross-method-detail}
\scriptsize
\setlength{\tabcolsep}{2.4pt}
\resizebox{\textwidth}{!}{%
\begin{tabular}{lcccc|cccc|cccc|cccc|c}
\toprule
& \multicolumn{4}{c|}{Q32B} & \multicolumn{4}{c|}{DS32B} & \multicolumn{4}{c|}{Q8B} & \multicolumn{4}{c|}{MoE} & cells $\geq 0$ \\
Method & GPQA & GSM & MATH & MBPP & GPQA & GSM & MATH & MBPP & GPQA & GSM & MATH & MBPP & GPQA & GSM & MATH & MBPP & / 16 \\
\midrule
\textbf{\talan{}+LoRA} & $+1.0$ & $+0.3$ & $0.0$ & $+0.4$ & $+4.0$ & $+0.3$ & $0.0$ & $+1.6$ & $+8.1$ & $0.0$ & $+1.6$ & $+1.0$ & $+1.0$ & $+0.2$ & $+3.0$ & $0.0$ & \textbf{16/16} \\
DoRA-only           & $-0.5$ & $-0.0$ & $+0.4$ & $-0.0$ & $+1.0$ & $+0.0$ & $+1.6$ & $+1.4$ & $-3.5$ & $+0.0$ & $+0.6$ & $+0.4$ & $-3.5$ & $+0.0$ & $+0.6$ & $+1.0$ & 11/16 \\
AdapterFusion       & $-2.5$ & $-0.6$ & $-0.4$ & $-0.4$ & $+1.0$ & $+0.7$ & $+1.0$ & $+0.5$ & $+5.5$ & $+0.5$ & $+1.0$ & $+0.6$ & $-1.5$ & $-0.0$ & $+0.4$ & $-0.5$ & 10/16 \\
DELIFT              & $-0.5$ & $-0.1$ & $+0.0$ & $+0.0$ & $+0.5$ & $+0.4$ & $+0.6$ & $+1.5$ & $+2.5$ & $+0.4$ & $+0.6$ & $+1.0$ & $-0.5$ & $+0.4$ & $+0.6$ & $+1.0$ & 12/16 \\
CogSteer            & $-0.5$ & $-0.1$ & $-0.0$ & $+0.4$ & $+0.5$ & $+0.0$ & $-0.4$ & $+0.0$ & $+0.5$ & $+0.0$ & $+0.4$ & $+0.0$ & $-1.5$ & $-0.4$ & $+1.0$ & $-0.0$ & 9/16 \\
RepE                & $-0.0$ & $-0.0$ & $-0.0$ & $-0.1$ & $-0.5$ & $-0.5$ & $-0.5$ & $+0.7$ & $+5.0$ & $+0.0$ & $+0.5$ & $+1.5$ & $-50$ & $-25$ & $-0.5$ & $-0.7$ & 5/16 \\
\bottomrule
\end{tabular}%
}
\end{table}

The non-regression rate column tells the story most cleanly. \talan{}
is the only method achieving 16-of-16; the next-closest (DELIFT) loses
on 4 cells; AdapterFusion loses on 6 cells (and requires
$\approx 2\times$ training cost).

\section{Configuration-Space Audit and Full-Non-Regression Rarity}
\label{app:search}

This appendix records the methodology, the per-backbone breakdown,
and the design-space structure of the $2{,}590$-configuration
\talan{} architecture audit referenced from
Section~\ref{sec:experiments} (one-paragraph pointer) and the
analysis discussion of Section~\ref{sec:mechanism}. The purpose of
this audit is not to search for a single best configuration but to
characterize which regions of the six-axis configuration space
produce non-regressing configurations and why. The audit's
headline (185/$2{,}590$ $=$ $7.1\%$ full-non-regression rate) and
design-space visualization (Figure~\ref{fig:ablationlandscape})
show that viable configurations occupy a structured,
host-conditional region rather than appearing at random.

\subsection{Audit scope and scoring}

The audit covers \emph{all} \talan{} architecture configurations
trained during our internal architecture audit on the four
audited backbones, evaluated under the same 4-benchmark suite
(\bench{GPQA Diamond}, \bench{GSM8K}, \bench{MATH-500}, \bench{MBPP}).
Training data, LoRA recipe, training schedule, and seed handling
match the headline protocol (Appendix~\ref{app:protocol}).
\emph{Full non-regression} is defined as $\Delta_{\mathrm{cell}}
\geq 0$ on all four cells of the corresponding backbone; in this
appendix we abbreviate it as ``G1''.

\paragraph{Important note on scoring methodology.}
The audit numbers in Table~\ref{tab:app-search-scale} use the
\emph{audit-time cached} scoring (the runtime-extraction ``ORIG''
scorer from the multi-scorer audit);
the headline four audited configurations are independently
re-scored under the lm-eval-harness scorer with
\texttt{math\_verify}~0.9.0. We do not re-score the full
2{,}590-configuration audit under lm-eval scoring because doing so
would require re-running $2{,}590$ inference passes (each at the
same generation budget as the headline). The two scorers are not
identical: per-cell-G1 ratios differ between the cached scorer
($13/16$) and the lm-eval scorer ($16/16$) on the four audited
configurations, while per-model full non-regression holds ($4/4$) under both.
The 7.1\% rarity therefore characterizes the \talan{} family under
its native cached scoring; the headline 4-of-4 / 16-of-16 result is
established under lm-eval re-scoring of the four configurations.
We flag this distinction here rather than burying it: it bounds the
audit's claim to ``rarity inside the family under cached scoring,''
which is the relevant quantity for characterizing the configuration
space, and it does not affect the lm-eval-scored headline.

\subsection{Per-backbone breakdown}

\begin{table}[!ht]
\centering
\caption{\textbf{Per-backbone configuration-space summary.} G1 rate
(non-negative deltas on all four aligned benchmarks within the
backbone) ranges from 2.1\% on Q30B-A3B (MoE) to 11.4\% on DS32B,
indicating that the structural difficulty of full non-regression
varies meaningfully across backbone architectures.}
\label{tab:app-search-scale}
\small
\begin{tabular}{lccc}
\toprule
Backbone & Configs tested & G1 achieved & G1 rate \\
\midrule
Q32B  & 309 & 9  & 2.9\% \\
DS32B & 790 & 90 & 11.4\% \\
Q8B   & 727 & 70 & 9.6\% \\
Q30B-A3B (MoE) & 764 & 16 & 2.1\% \\
\midrule
\textbf{Total} & \textbf{2{,}590} & \textbf{185} & \textbf{7.1\%} \\
\bottomrule
\end{tabular}
\end{table}

The per-backbone variation is itself informative. Among the three
dense backbones with trained \talan{} interventions, DS32B admits
the largest fraction of G1-achieving configurations (11.4\% of
790), consistent with its champion being a relatively flexible
cross-attention mixer at the embedding locus, while Q32B admits
the smallest (2.9\% of 309). Q30B-A3B (MoE), which uses the
mechanistically distinct frozen-perturbation variant
(Appendix~\ref{app:operating-points}), admits the smallest fraction
overall (2.1\% of 764 audited configurations), consistent with the
additional constraint that LoRA must compose around a fixed
random perturbation rather than co-training with a learned one. The same
geometry shows up in
Figure~\ref{fig:ablationlandscape}~(D)---winners occupy
backbone-specific (connection depth $\times$ slot count) positions.

\subsection{Per-axis structure and reading}

\begin{figure*}[!ht]
\centering
\includegraphics[width=0.96\textwidth]{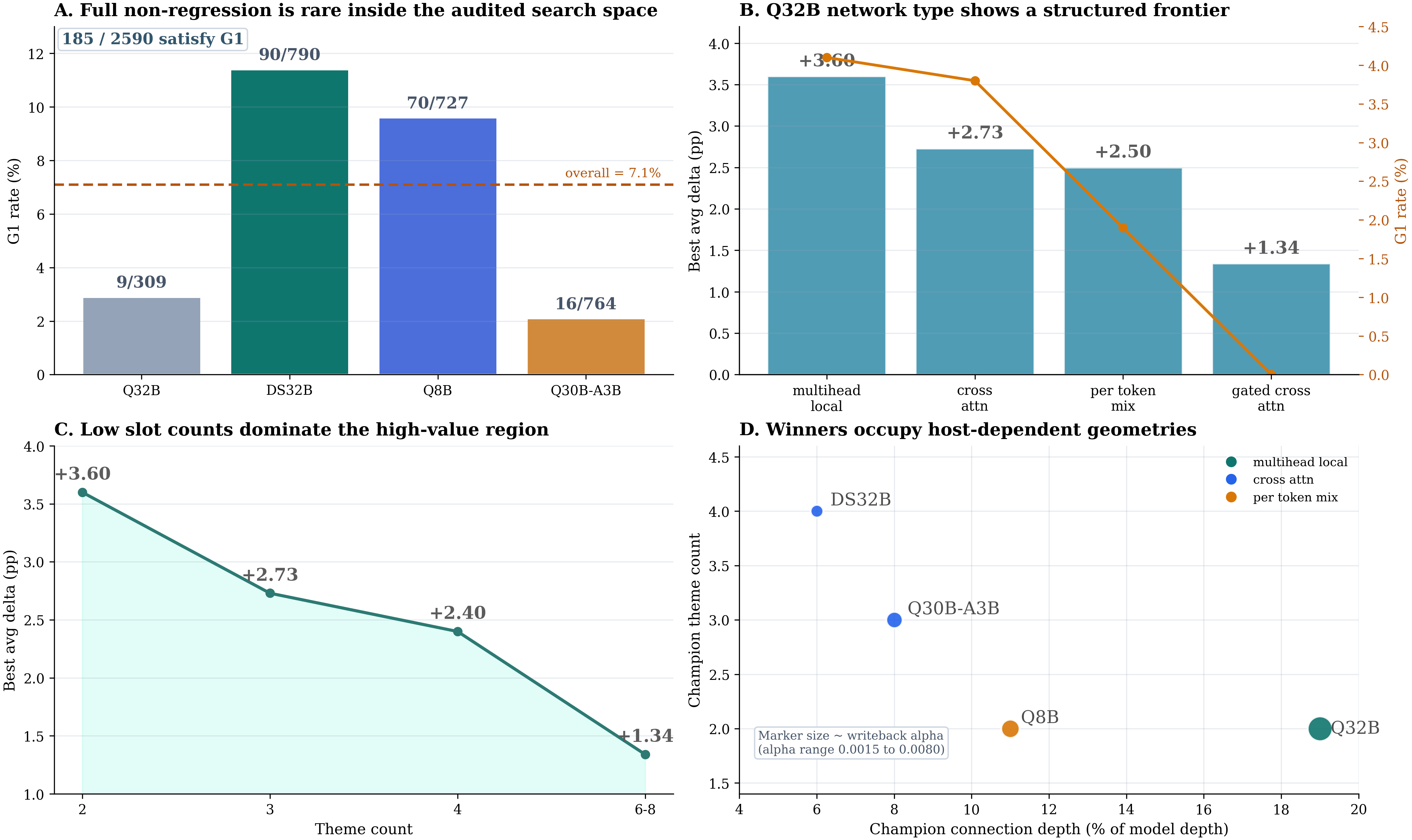}
\caption{\textbf{Architecture-audit view of the \talan{} family.}
(A) Per-backbone G1 rate; the dashed line marks the overall $7.1\%$
rate. (B) On Q32B, the four mixers form a structured frontier in
best-average-delta vs.\ G1 rate. (C) Low slot counts dominate the
high-value region ($T = 2$ reaches the largest best-avg delta, with
monotone decay through $T = 6$--$8$). (D) Champions occupy
host-dependent geometries in (connection depth $\times$ slot count)
space; marker size encodes the writeback coefficient $\alpha_\pi$.}
\label{fig:ablationlandscape}
\end{figure*}

The four panels of Figure~\ref{fig:ablationlandscape} together show
that the high-value region of the \talan{} family is structured, not
diffuse: low slot counts ($T = 2$--$3$), shallow-to-mid intervention
depth, and small writeback strengths dominate. The audit therefore
supports a \emph{configuration-space} reading of the results: viable
configurations sit in a narrow, host-conditional region predicted
by the same six axes, rather than appearing diffusely across the
space.
The per-family weight signatures of
Appendix~\ref{app:weights} pick out the same high-value
region from saved weights alone, providing an independent
verification of the structure shown here.

\section{Statistical-Test Derivations}
\label{app:stat-derivation}

This appendix expands on the three statistical lenses used in the
audit, including assumption discussion,
robustness checks, and the per-cell McNemar and bootstrap tables
referenced from the main text.

\subsection{Sign test on cells: derivation and assumptions}

Let $\Delta_i = \mathrm{TALAN}_i - \mathrm{LoRA}_i$ for
$i = 1, \dots, 16$ be the per-cell deltas. Define
$X_i = \mathbf{1}[\Delta_i \geq 0]$ and $X = \sum_i X_i$. Under the
null $H_0\!:\! P(\Delta_i \geq 0) = 0.5$ \emph{independently for
each $i$}, $X \sim \mathrm{Binom}(16, 0.5)$. The observed value is
$X = 16$. The exact one-sided $p$-value is
\[
P(X \geq 16 \mid H_0)
= \binom{16}{16} (0.5)^{16} (0.5)^{0}
= 2^{-16}
= 1.5258789 \times 10^{-5},
\]
which we report as $p = 1.53 \times 10^{-5}$ to three significant
figures. The two-sided $p$-value is
$2 \cdot 2^{-16} = 3.05 \times 10^{-5}$ (the null is symmetric and
the ``extreme outcome'' is symmetric).

\paragraph{The independence assumption.}
The i.i.d.\ assumption is the strongest aspect of this null. Cells
within a backbone are correlated: if a particular configuration
lands favorably for one backbone, several of its benchmarks may
shift together. We address this in three ways: (i) by reporting
the more conservative \emph{backbone-level} sign test
($P((\bar{\Delta}_b \geq 0)_{b=1}^{4}) = 2^{-4} = 0.0625$, treating
each of the four backbones as one independent trial---marginally
significant); (ii) by reporting the per-cell McNemar test on
discordant items, which makes only the (very mild) assumption of
within-cell exchangeability of the two methods on the same items
(a property guaranteed by our matched-protocol setup); and (iii)
by reporting per-cell bootstrap CIs that resample item-level
correctness within each cell.

\paragraph{The $\Delta\geq 0$ vs.\ $\Delta>0$ choice.}
The null $P(\Delta\geq 0)=0.5$ is the standard one-sided sign-test
null when ties are broken in favor of the alternative. Under the
strict variant that excludes the $4$ ties at exactly $0.00$ and
counts only $\Delta > 0$, we have $X' = 12$ out of $n' = 12$
non-tie cells (equivalently $X' = 12$ out of $16$ if we keep the
denominator), giving
$P(X' \geq 12 \mid \mathrm{Binom}(16, 0.5)) = 0.038$.
The $\Delta\geq 0$ formulation is the headline because the alternative
hypothesis under test is non-regression (a tied cell does not
violate non-regression).

\subsection{Per-cell McNemar test}

For cell $i$, define the $2 \times 2$ contingency table
$(a_i, b_i; c_i, d_i)$: $a_i$ = items both correct, $b_i$ = TALAN
correct \& LoRA wrong, $c_i$ = TALAN wrong \& LoRA correct,
$d_i$ = both wrong. The McNemar statistic is
$\chi^2_i = (b_i - c_i)^2 / (b_i + c_i)$, distributed as $\chi^2_1$
under $H_0\!:\! b_i = c_i$. The four \bench{GPQA Diamond} cells
yield:

\begin{table}[!ht]
\centering
\small
\caption{\textbf{Per-cell McNemar test on \bench{GPQA Diamond}
cells.}}
\label{tab:mcnemar-gpqa}
\setlength{\tabcolsep}{4pt}
\resizebox{\textwidth}{!}{%
\begin{tabular}{lcccc}
\toprule
Cell & TALAN-unique wins ($b$) & LoRA-unique wins ($c$) & McNemar $\chi^2$ & one-sided $p$ \\
\midrule
Q8B  / GPQA & $28$ & $11$ & $7.41$ & $0.011$ \\
DS32B / GPQA & $26$ & $12$ & $5.16$ & $0.034$ \\
Q32B / GPQA & $26$ & $12$ & $5.16$ & $0.034$ \\
MoE  / GPQA & $17$ & $15$ & $0.13$ & $0.36$ (positive; not individually significant) \\
\bottomrule
\end{tabular}%
}
\end{table}

The summed-across-$16$-cells discordant net advantage is $+162$
items (TALAN-unique $591$; LoRA-unique $429$; total discordants
$1{,}020$).

\subsection{Bootstrap 95\% CIs on per-cell deltas}

We bootstrap the per-cell $\Delta$ ($B = 2{,}000$ item resamples
per cell, sampled with replacement at the item level). Two cells
have 95\% CIs that entirely exclude zero; a third
(Q32B / \bench{MATH-500}) is reported for completeness but its CI
includes zero:

\begin{table}[!ht]
\centering
\small
\caption{\textbf{Bootstrap 95\% CIs} for the three cells with the
largest net deltas. The first two CIs exclude zero; the third
includes zero.}
\label{tab:bootstrap-ci}
\begin{tabular}{lccc}
\toprule
Cell & Net $\Delta$ (items) & 95\% CI (items) & $\Pr[\Delta > 0]$ \\
\midrule
Q8B  / GPQA      & $+13$ & $[+1, +27]$ & $0.974$ \\
Q8B  / MBPP      & $+9$  & $[+1, +18]$ & $0.973$ \\
Q32B / MATH-500  & $+16$ & $[-4, +37]$ & $0.940$ \\
\bottomrule
\end{tabular}
\end{table}

\section{Per-Layer Activation Divergence (Full)}
\label{app:activation}

Section~\ref{sec:insight-design-language} reports the headline activation analysis
on \bench{GPQA Diamond}. This appendix provides the full per-layer
activation divergence plots for all four backbones across all four
benchmarks. Perturbation amplification through depth is generically
expected in deep residual networks; the analysis therefore targets
two more specific questions: whether the \talan{} perturbation
\emph{persists} rather than being absorbed, and whether the resulting
divergence pattern is a stable property of the model variant rather
than varying with the prompt distribution.

\subsection{Methodology}

We collect activations on a single forward pass over each prompt (no
generation), capturing the residual-stream input at each decoder layer.
For each example we compute the relative L2 difference
(\texttt{rel\_diff}) between (a) the matched LoRA baseline and (b) the
\talan{}+LoRA champion; and between (c) the \talan{}+LoRA champion
with \talan{}-module hooks attached and (d) the same model with the
\talan{}-module hooks disabled. The (c) vs.\ (d) comparison isolates the
direct contribution of the \talan{} module from the indirect
contribution of the co-trained LoRA adapter.

\subsection{Per-layer relative-L2 divergence (all four backbones)}

\begin{figure}[!ht]
\centering
\begin{tabular}{cc}
\includegraphics[width=0.46\textwidth]{figures/figures_v2/Q8B_gpqa_rel_diff_per_layer_scatter.png} &
\includegraphics[width=0.46\textwidth]{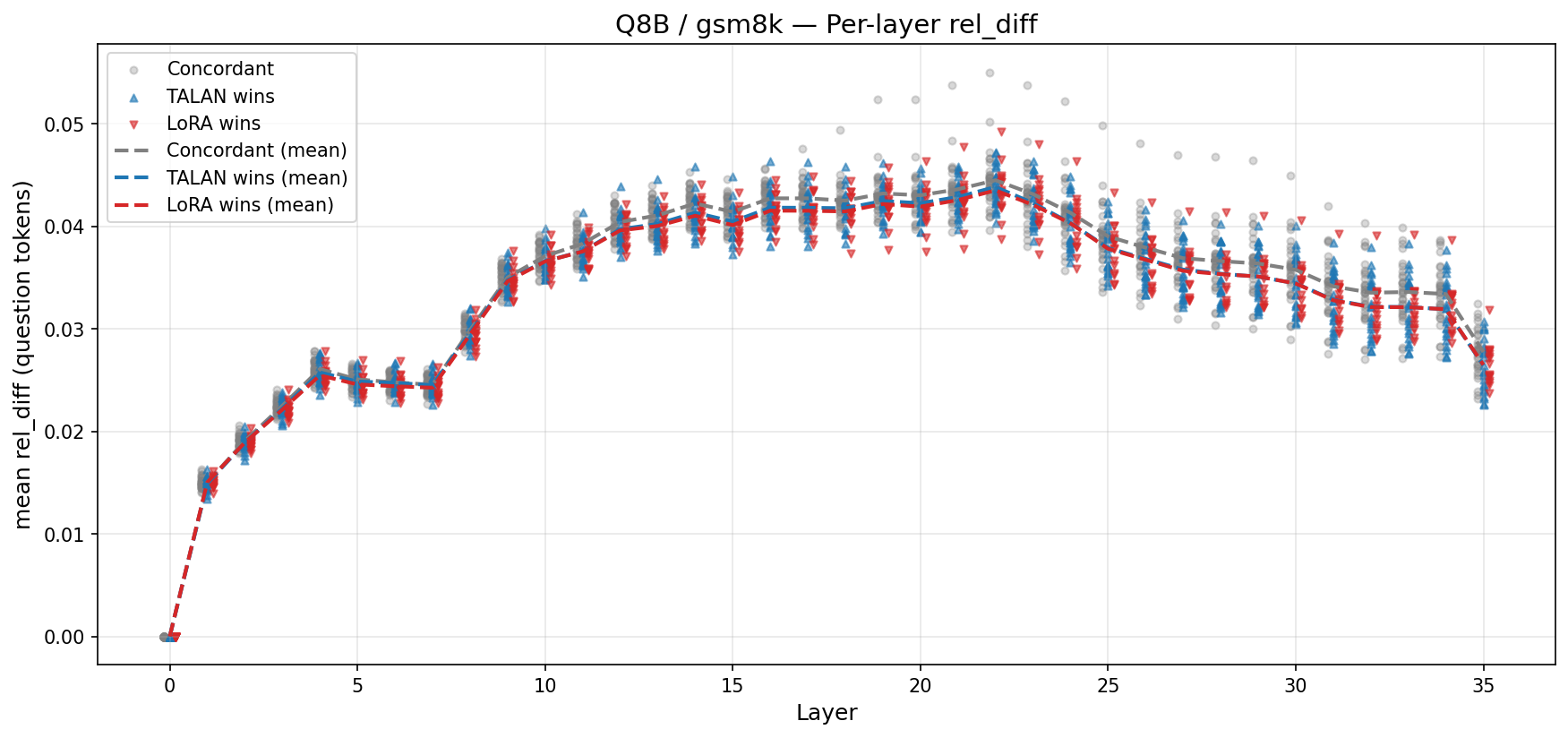} \\
\includegraphics[width=0.46\textwidth]{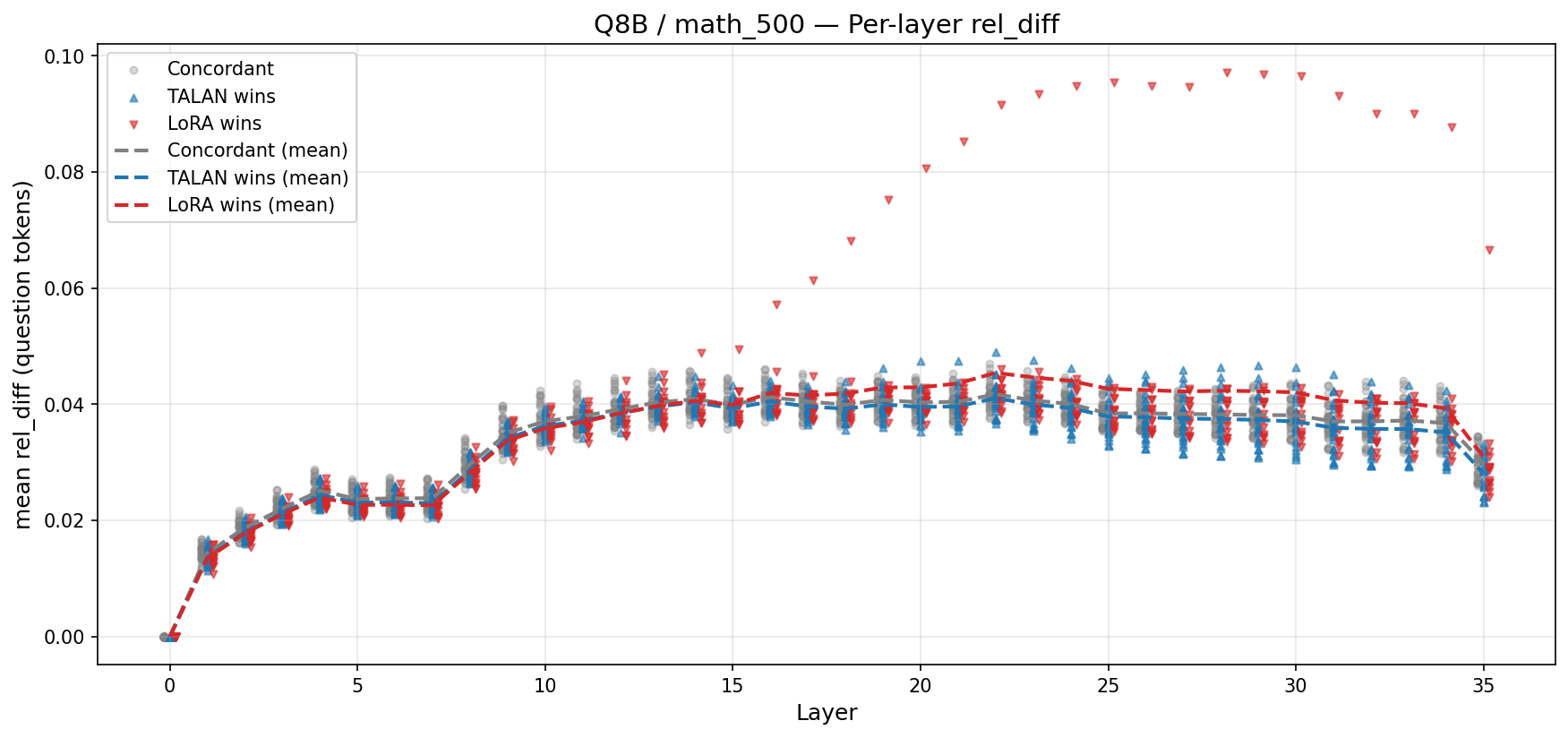} &
\includegraphics[width=0.46\textwidth]{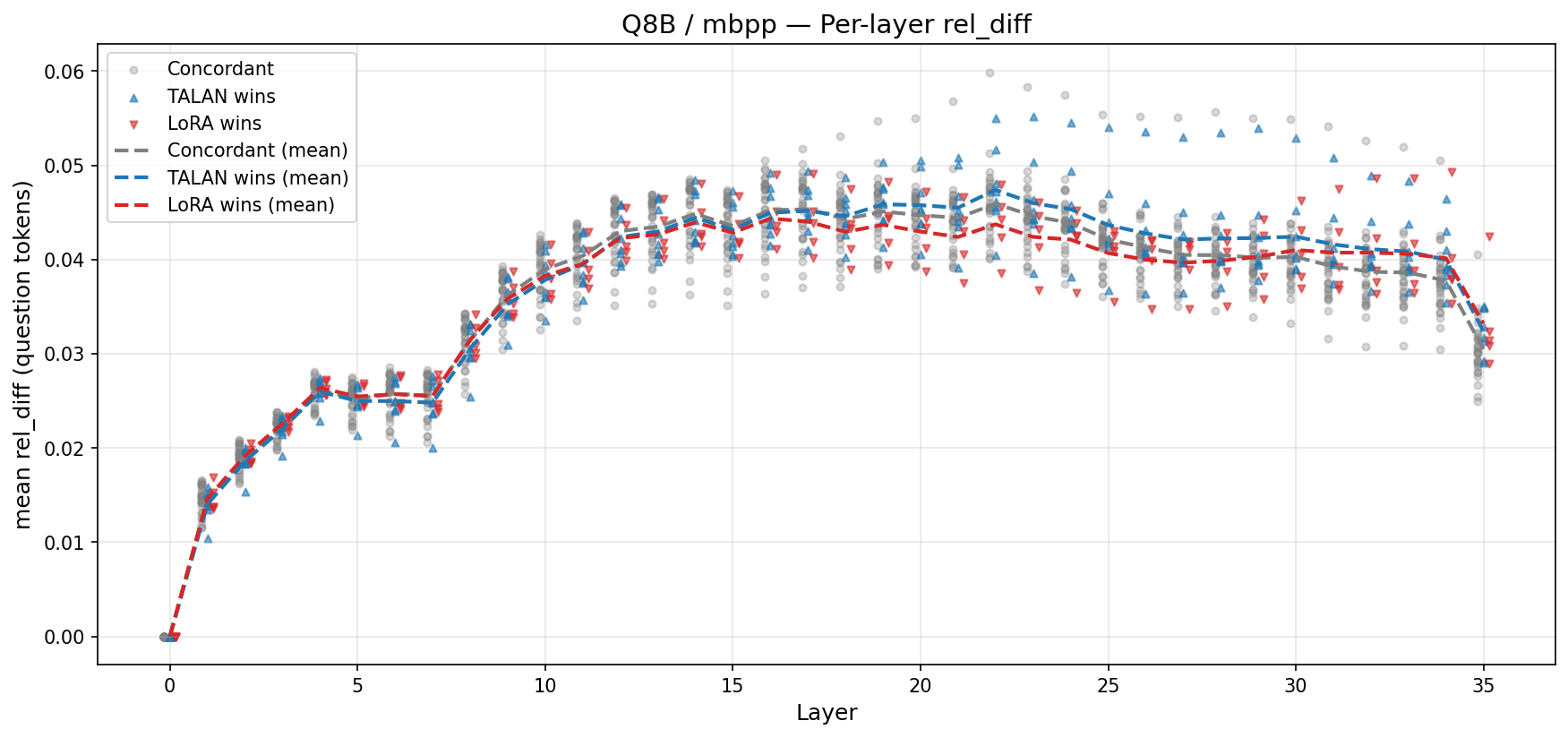} \\
\end{tabular}
\caption{\textbf{Qwen3-8B per-layer relative-L2 divergence},
four benchmarks (\bench{GPQA Diamond}, \bench{GSM8K}, \bench{MATH-500}, \bench{MBPP}). Each point is
one example; vertical axis is the rel\_diff between \talan{}+LoRA
and the matched LoRA baseline at that layer. Insertion locus: layer
$4$.}
\label{fig:per-layer-q8b}
\end{figure}

\begin{figure}[!ht]
\centering
\begin{tabular}{cc}
\includegraphics[width=0.46\textwidth]{figures/figures_v2/DS32B_gpqa_rel_diff_per_layer_scatter.png} &
\includegraphics[width=0.46\textwidth]{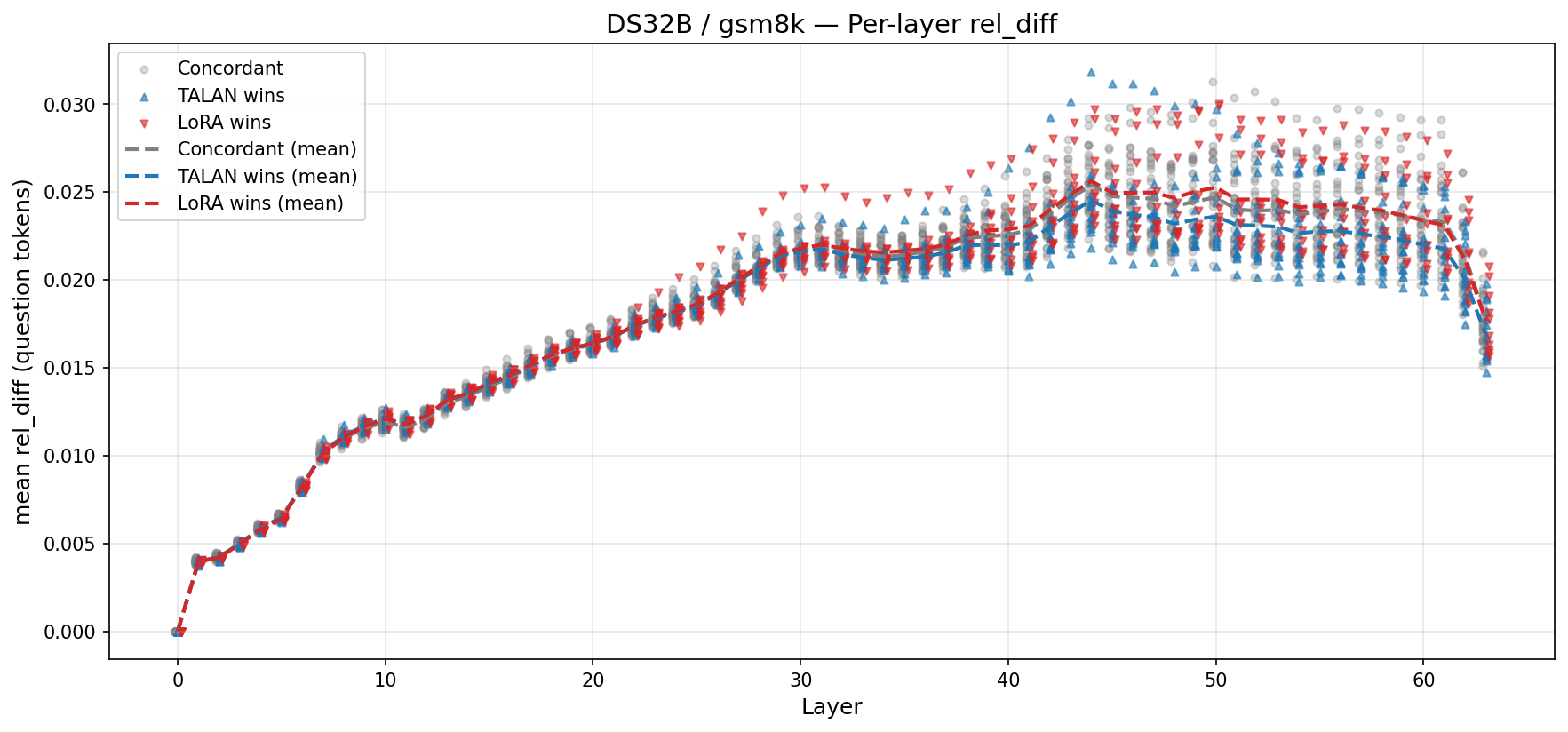} \\
\includegraphics[width=0.46\textwidth]{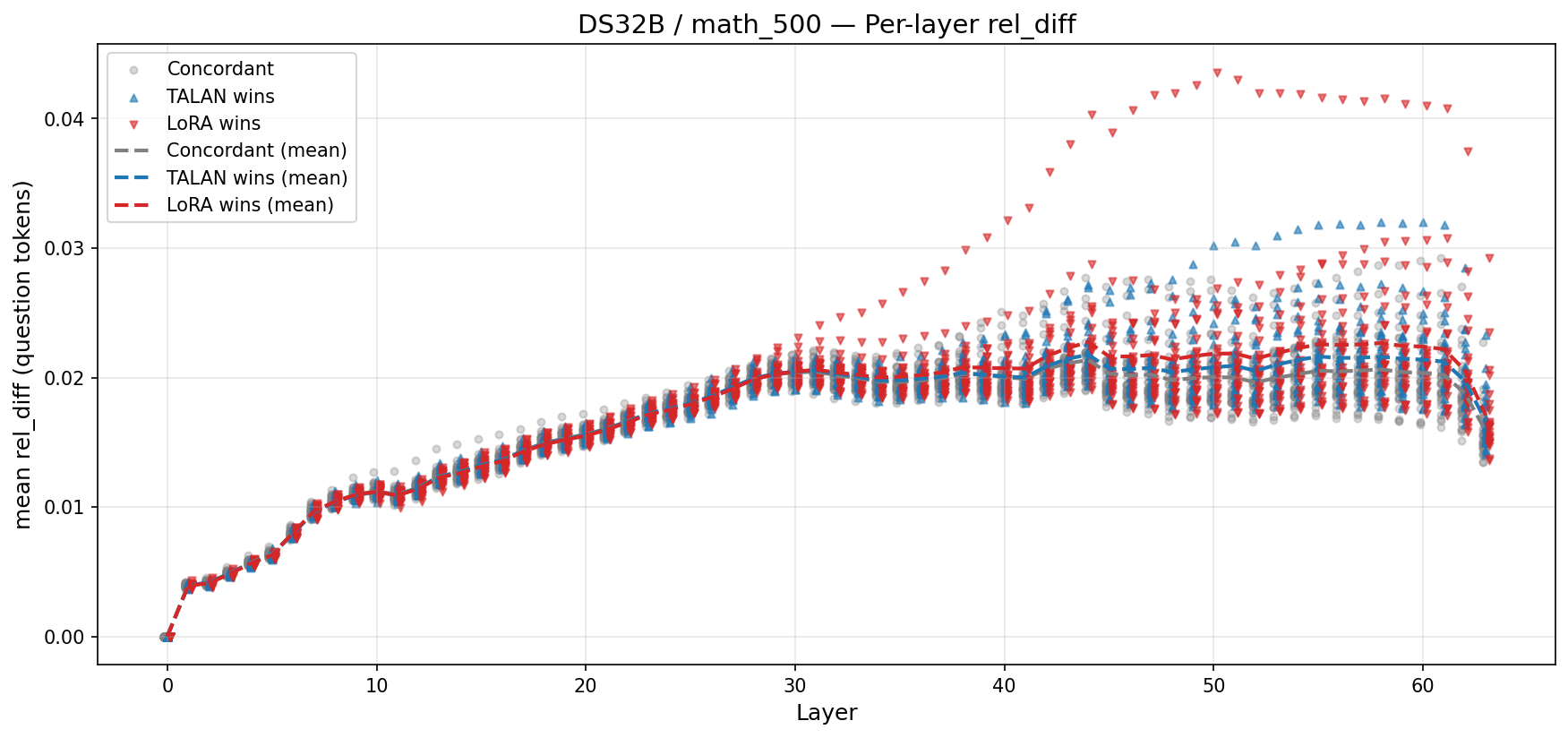} &
\includegraphics[width=0.46\textwidth]{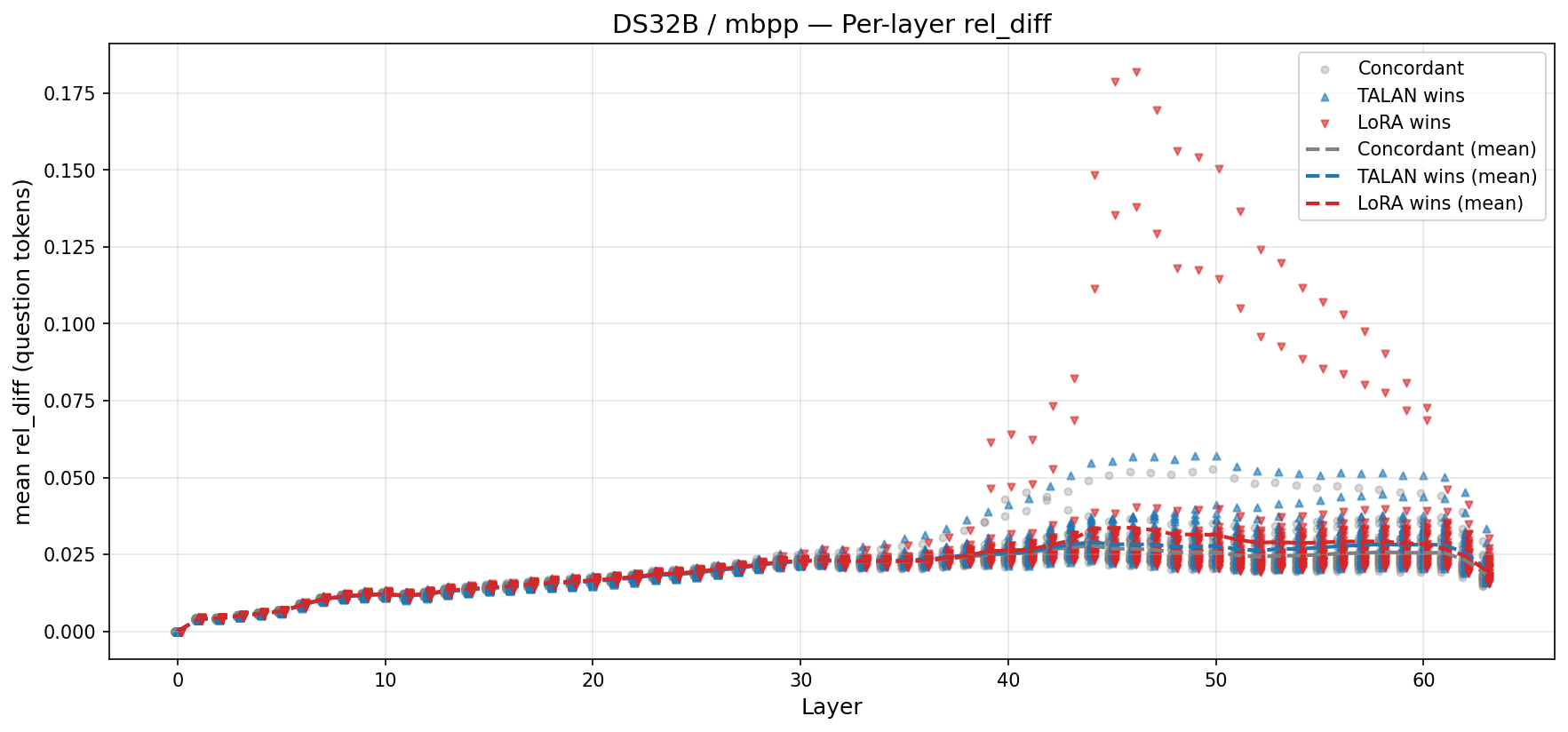} \\
\end{tabular}
\caption{\textbf{DeepSeek-R1-Distill-Qwen-32B per-layer relative-L2
divergence}, four benchmarks. Insertion locus: embedding (layer $0$).}
\label{fig:per-layer-ds32b}
\end{figure}

\begin{figure}[!ht]
\centering
\begin{tabular}{cc}
\includegraphics[width=0.46\textwidth]{figures/figures_v2/Q32B_gpqa_rel_diff_per_layer_scatter.png} &
\includegraphics[width=0.46\textwidth]{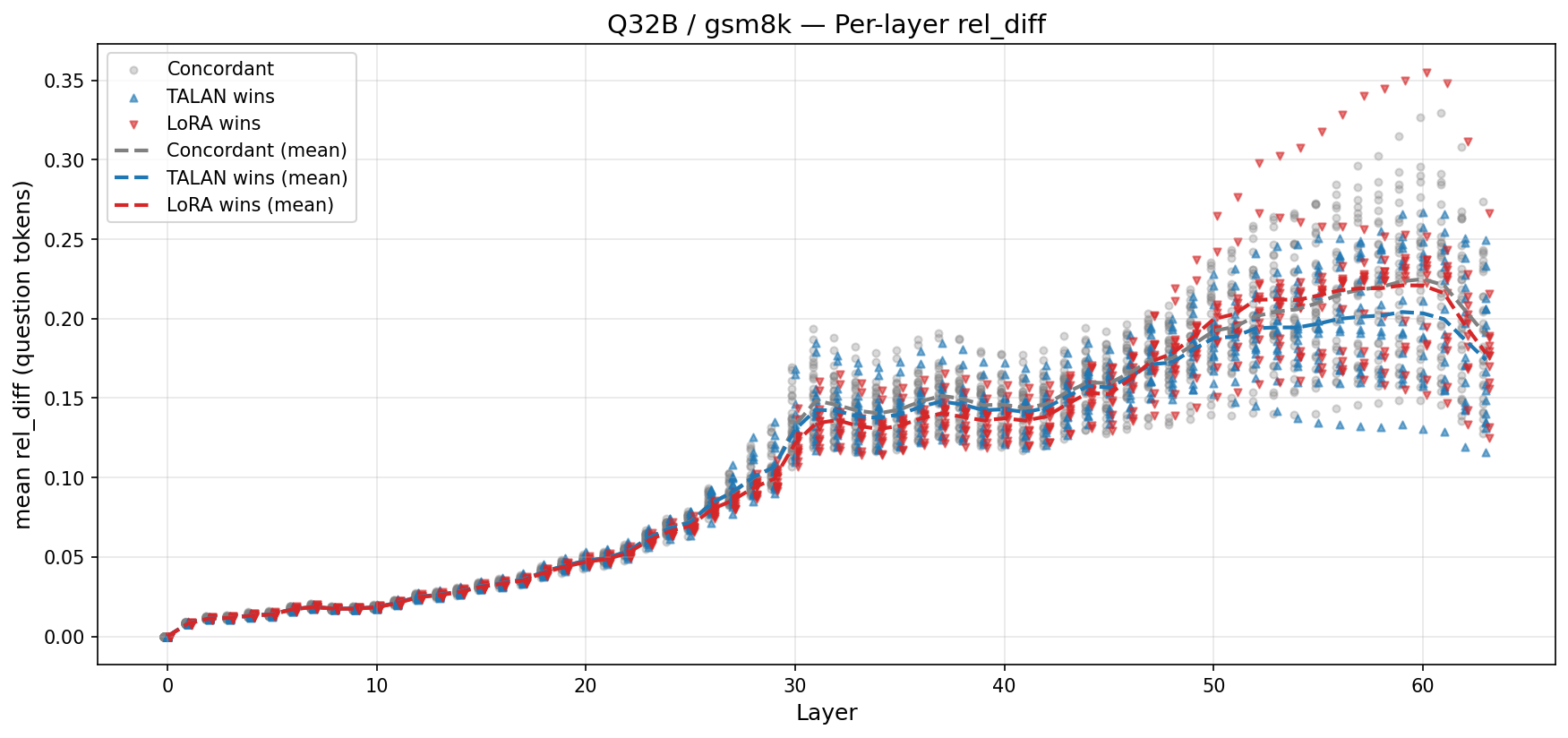} \\
\includegraphics[width=0.46\textwidth]{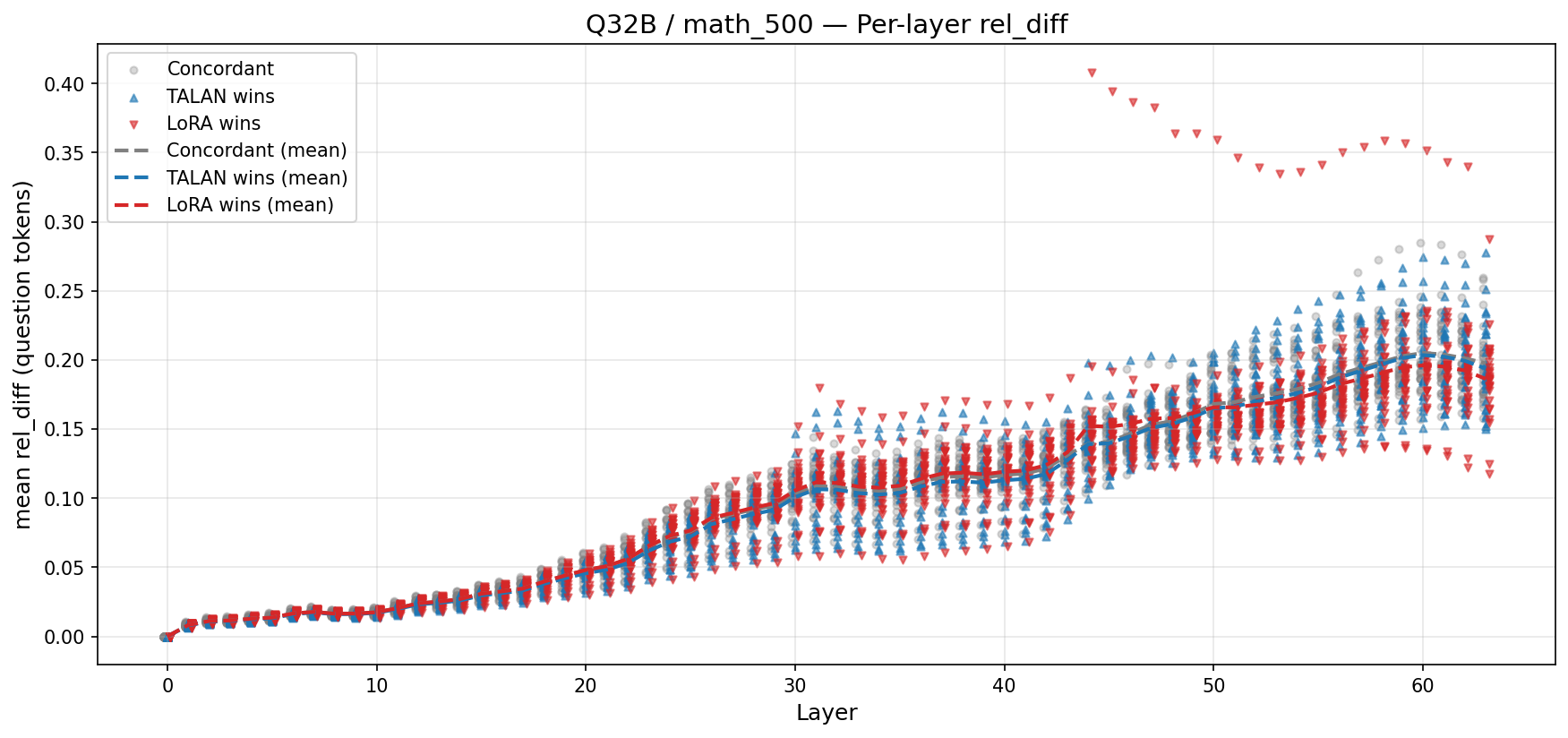} &
\includegraphics[width=0.46\textwidth]{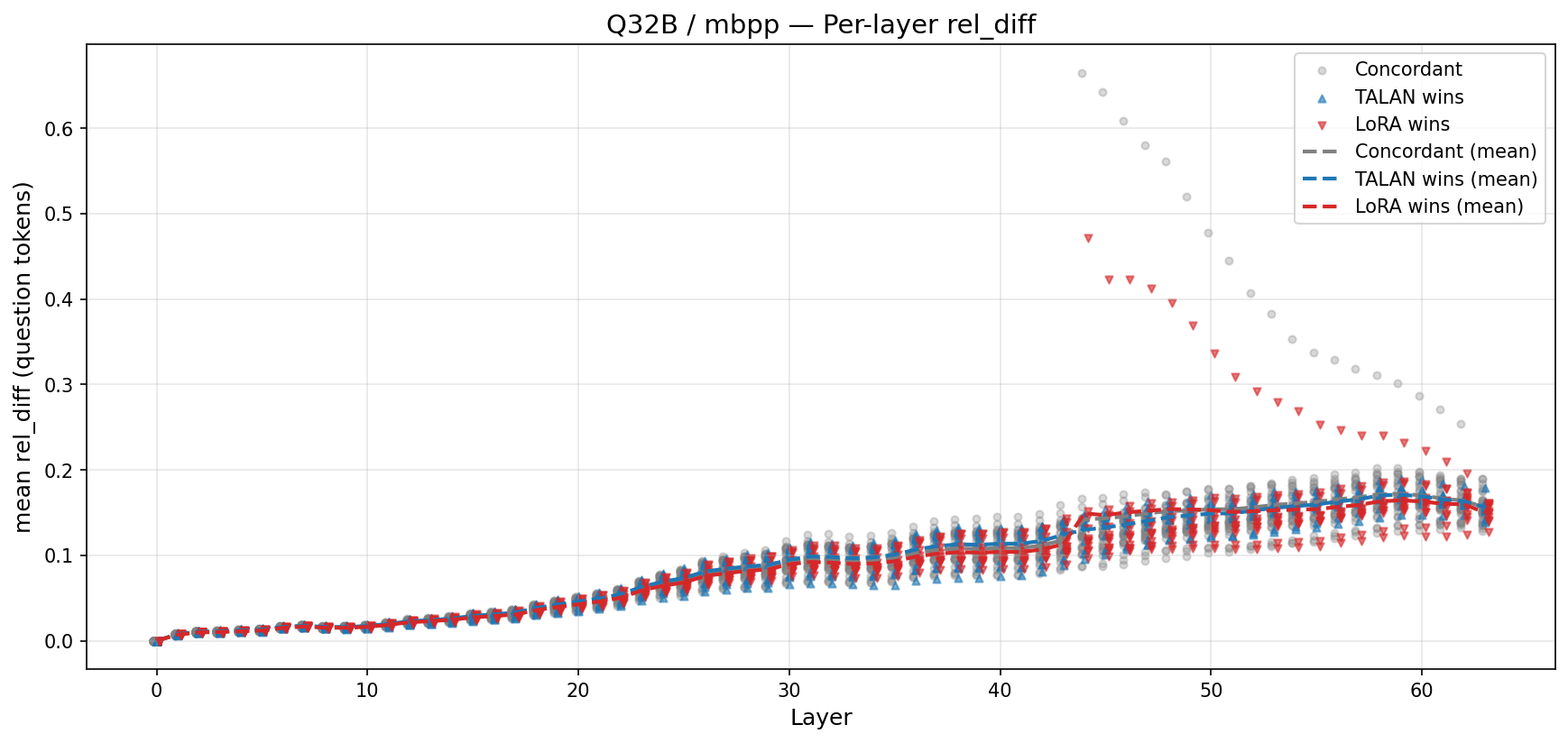} \\
\end{tabular}
\caption{\textbf{Qwen3-32B per-layer relative-L2 divergence}, four
benchmarks. Insertion locus: layer $12$.}
\label{fig:per-layer-q32b}
\end{figure}

\begin{figure}[!ht]
\centering
\begin{tabular}{cc}
\includegraphics[width=0.46\textwidth]{figures/figures_v2/MoE_gpqa_rel_diff_per_layer_scatter.png} &
\includegraphics[width=0.46\textwidth]{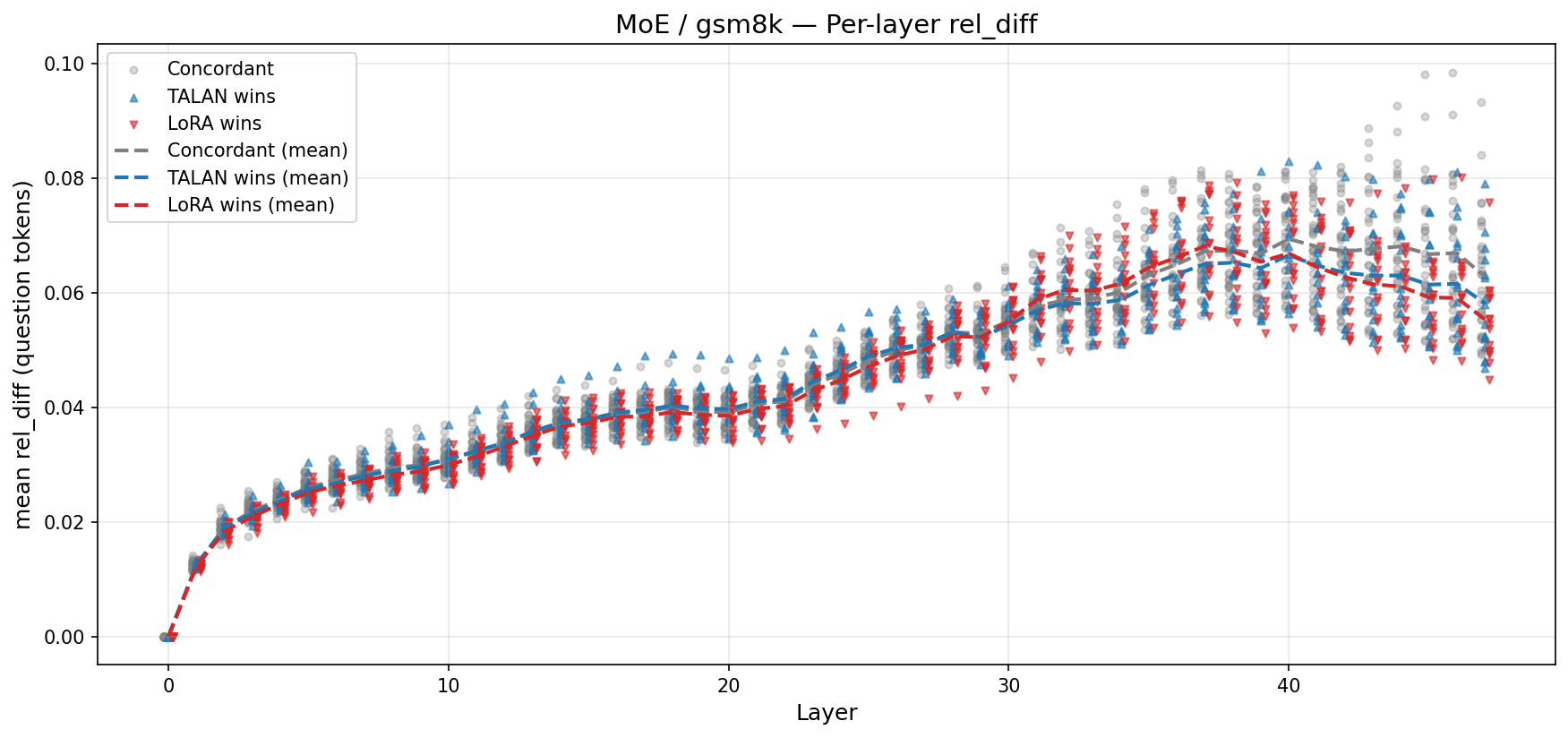} \\
\includegraphics[width=0.46\textwidth]{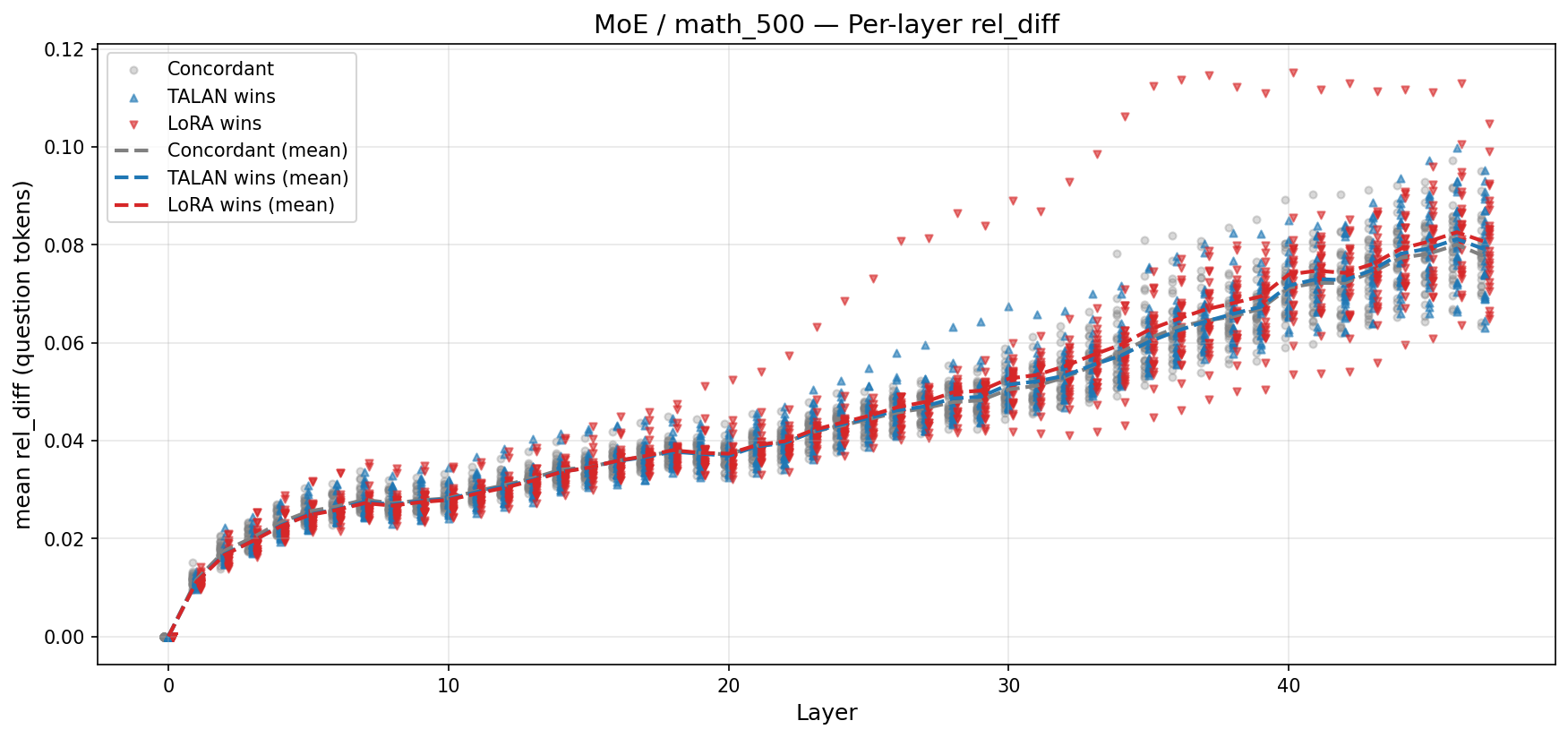} &
\includegraphics[width=0.46\textwidth]{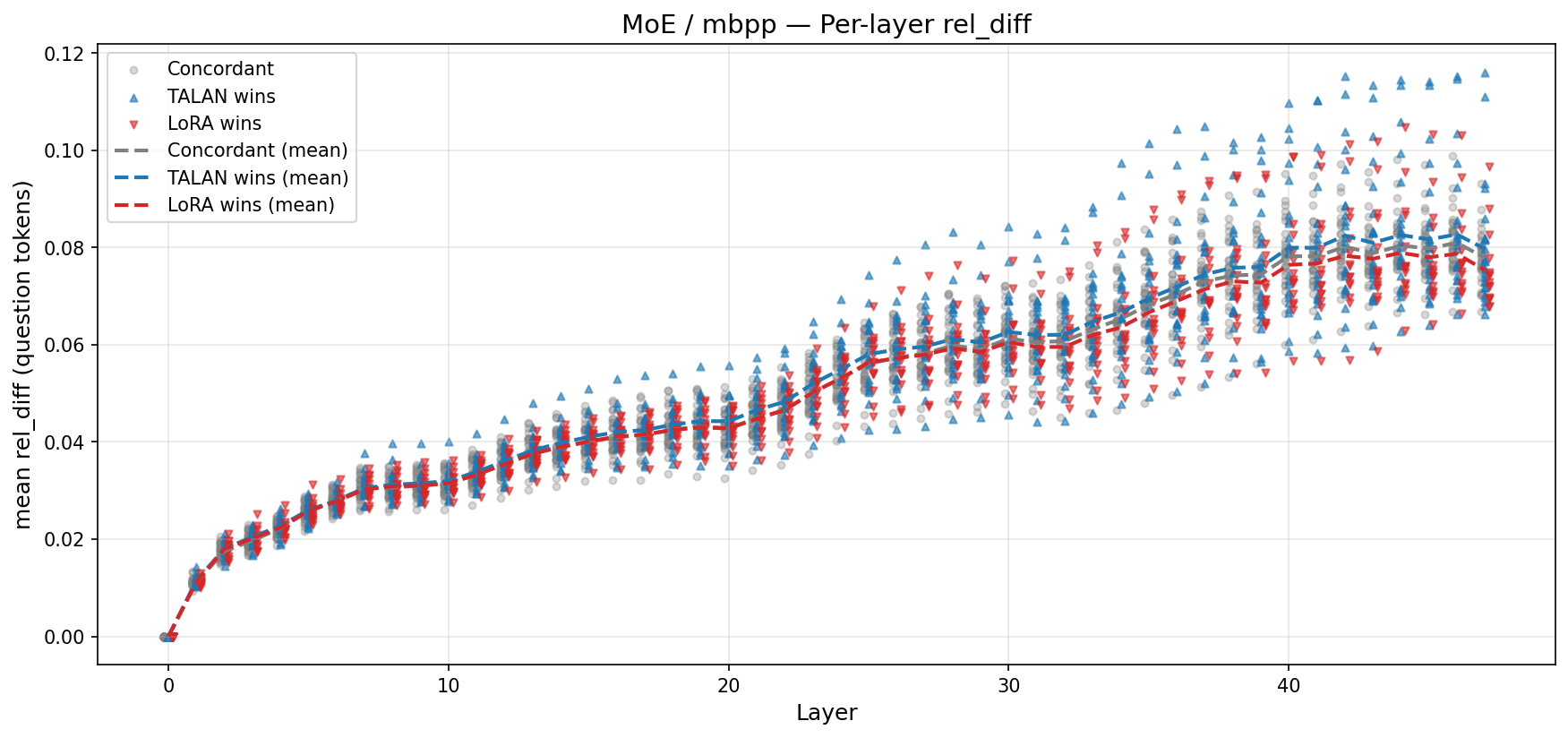} \\
\end{tabular}
\caption{\textbf{Qwen3-30B-A3B (MoE) per-layer relative-L2 divergence},
four benchmarks. Insertion locus: embedding (layer $0$).}
\label{fig:per-layer-moe}
\end{figure}

Figures~\ref{fig:per-layer-q8b}--\ref{fig:per-layer-moe} show the
per-layer relative-L2 difference between \talan{}+LoRA and the matched
LoRA baseline for each of the four backbones, with one panel per
benchmark. Insertion loci differ across backbones (Q8B at
layer~$4$; Q32B at layer~$12$; DS32B and MoE at the embedding); the
per-layer-rel\_diff curves' lift-off matches the configured
insertion locus on every backbone, and the curves grow monotonically
through subsequent layers.

\subsection{Per-layer cosine similarity (all four backbones)}

\begin{figure}[!ht]
\centering
\begin{tabular}{cc}
\includegraphics[width=0.46\textwidth]{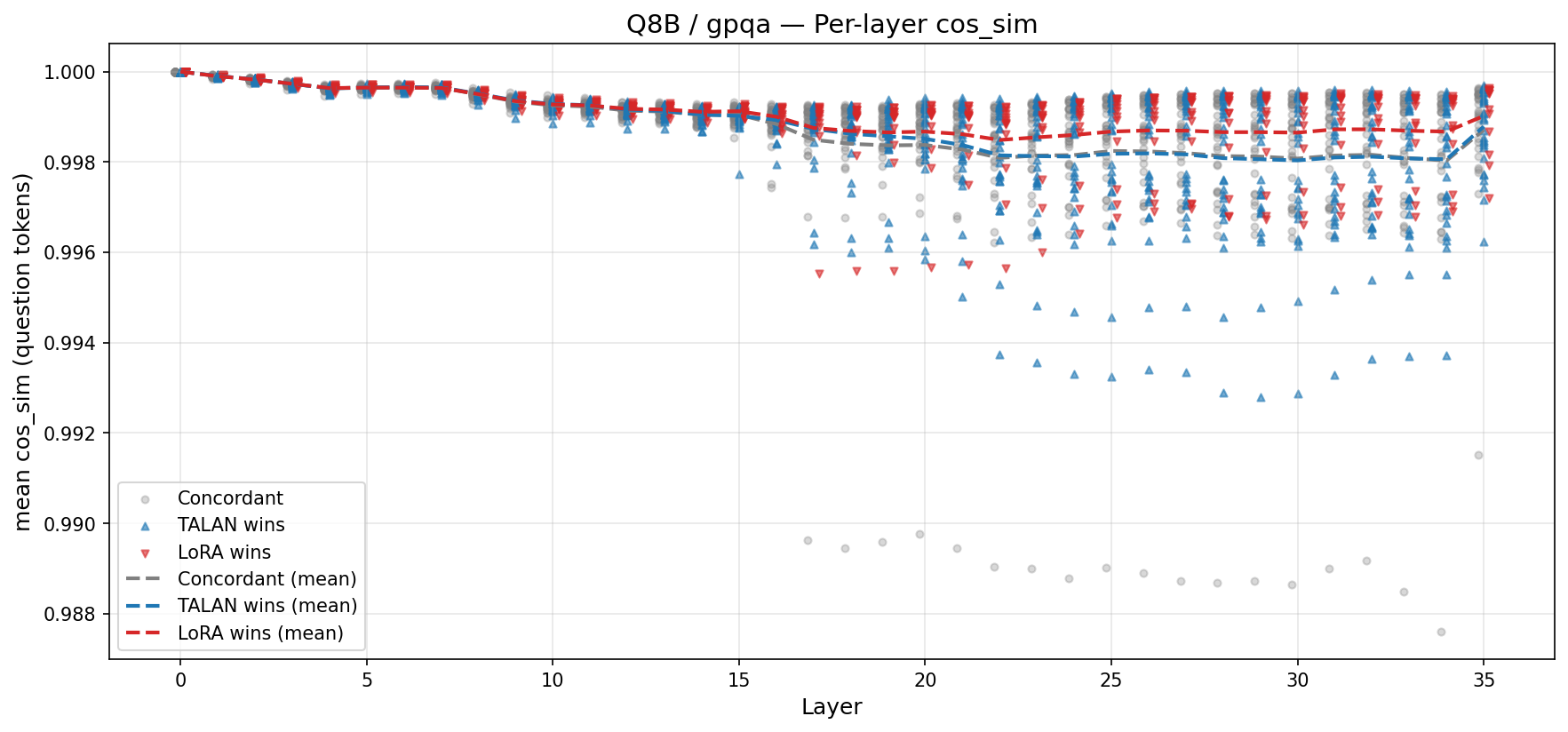} &
\includegraphics[width=0.46\textwidth]{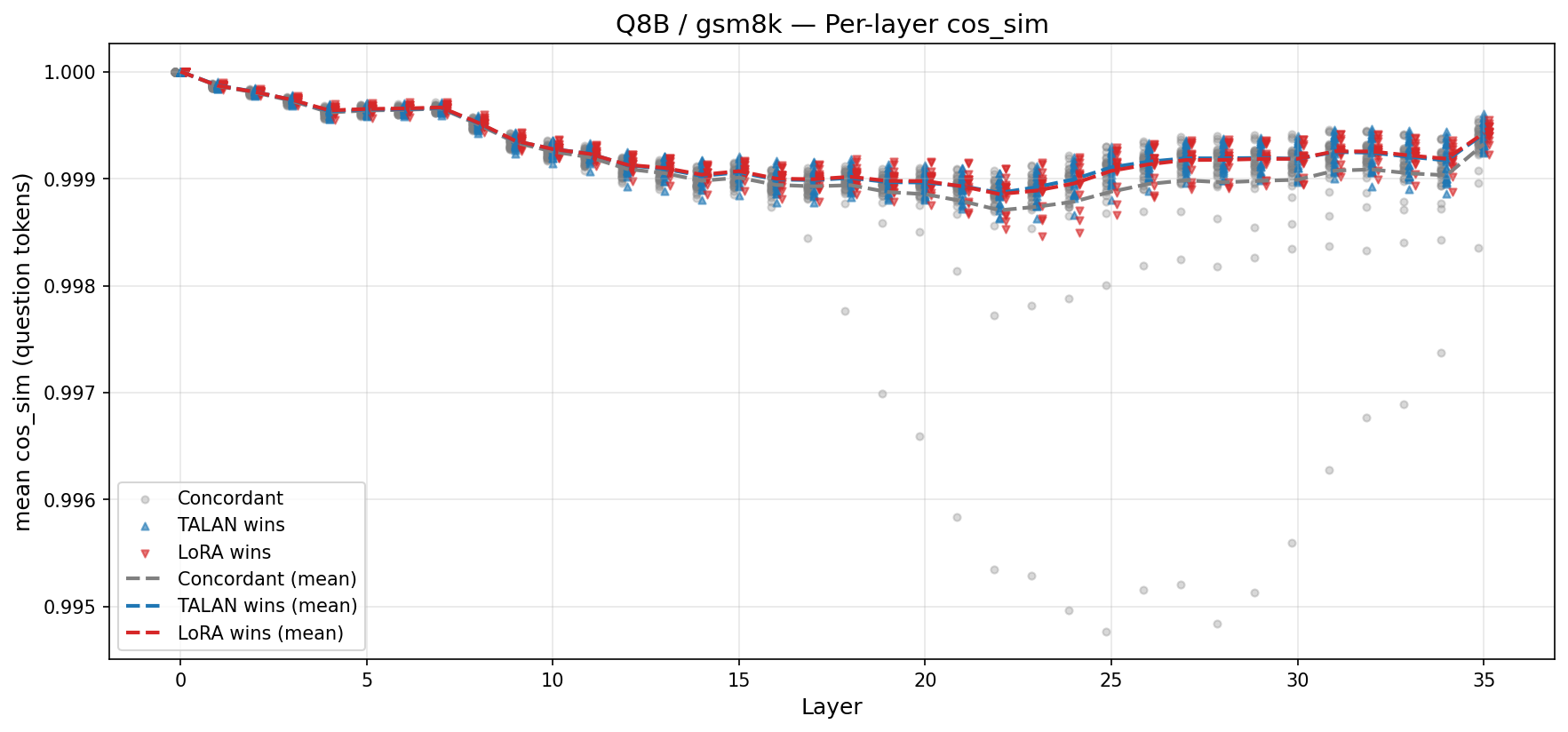} \\
\includegraphics[width=0.46\textwidth]{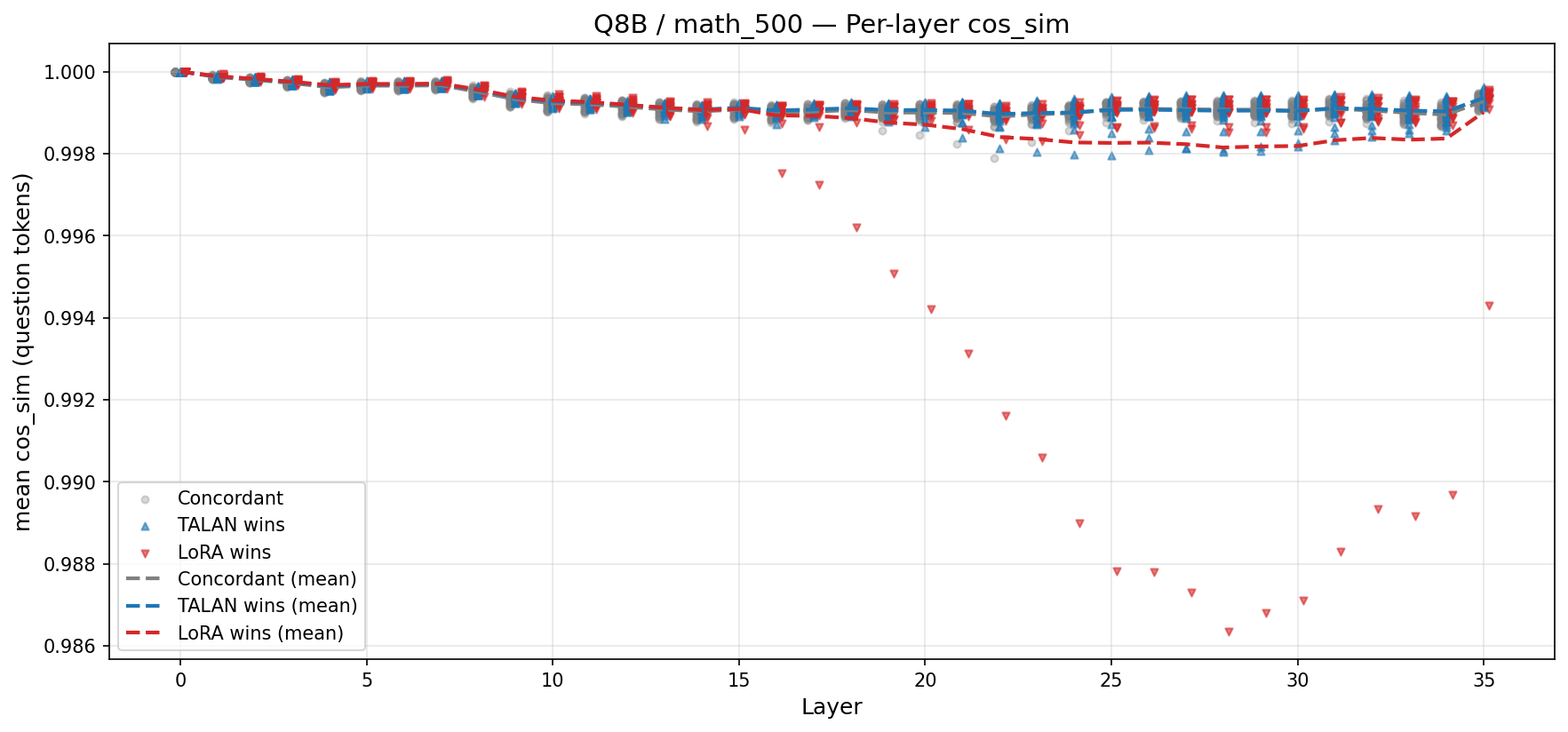} &
\includegraphics[width=0.46\textwidth]{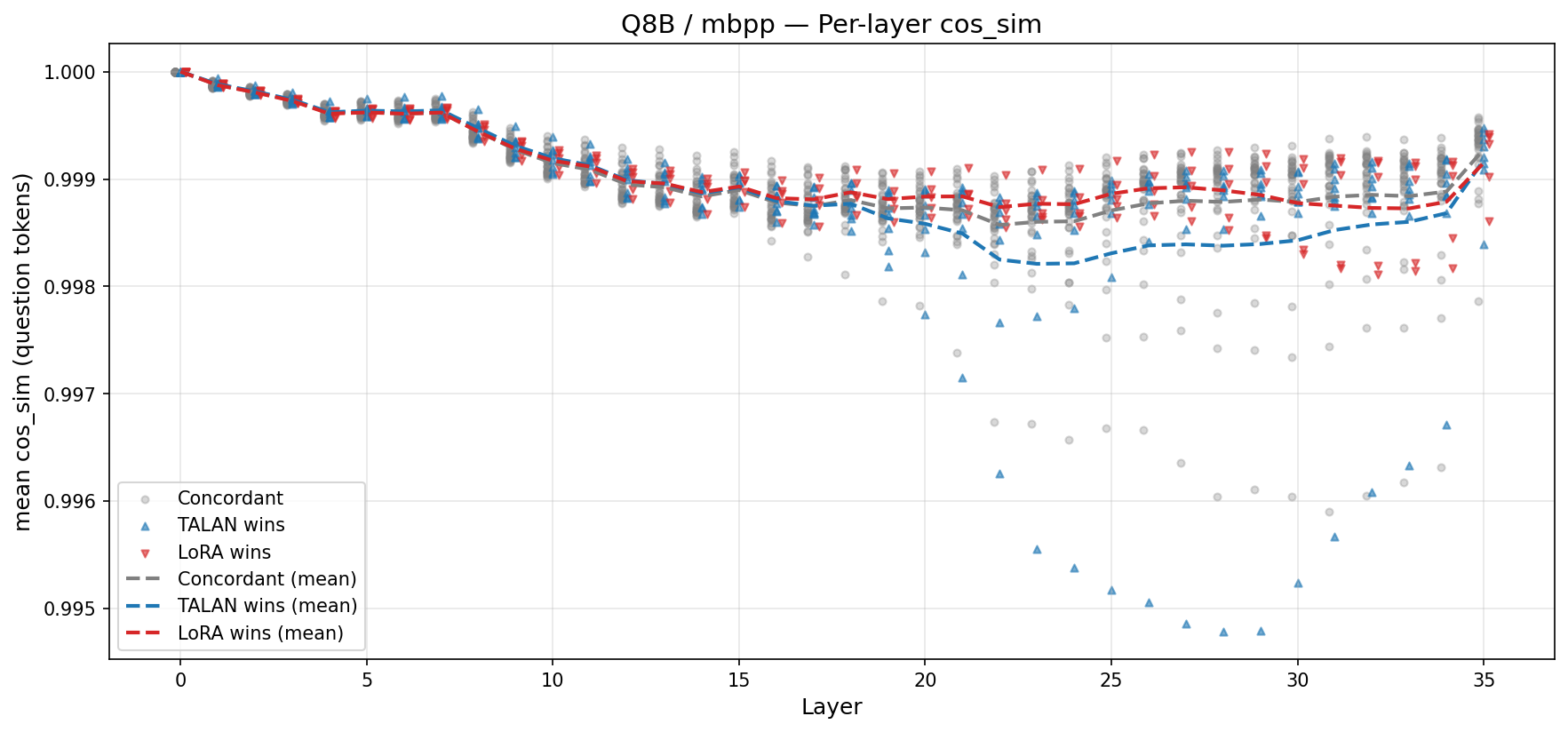} \\
\end{tabular}
\caption{\textbf{Qwen3-8B per-layer cosine similarity} between
\talan{}+LoRA and matched LoRA activations, four benchmarks.}
\label{fig:cos-q8b}
\end{figure}

\begin{figure}[!ht]
\centering
\begin{tabular}{cc}
\includegraphics[width=0.46\textwidth]{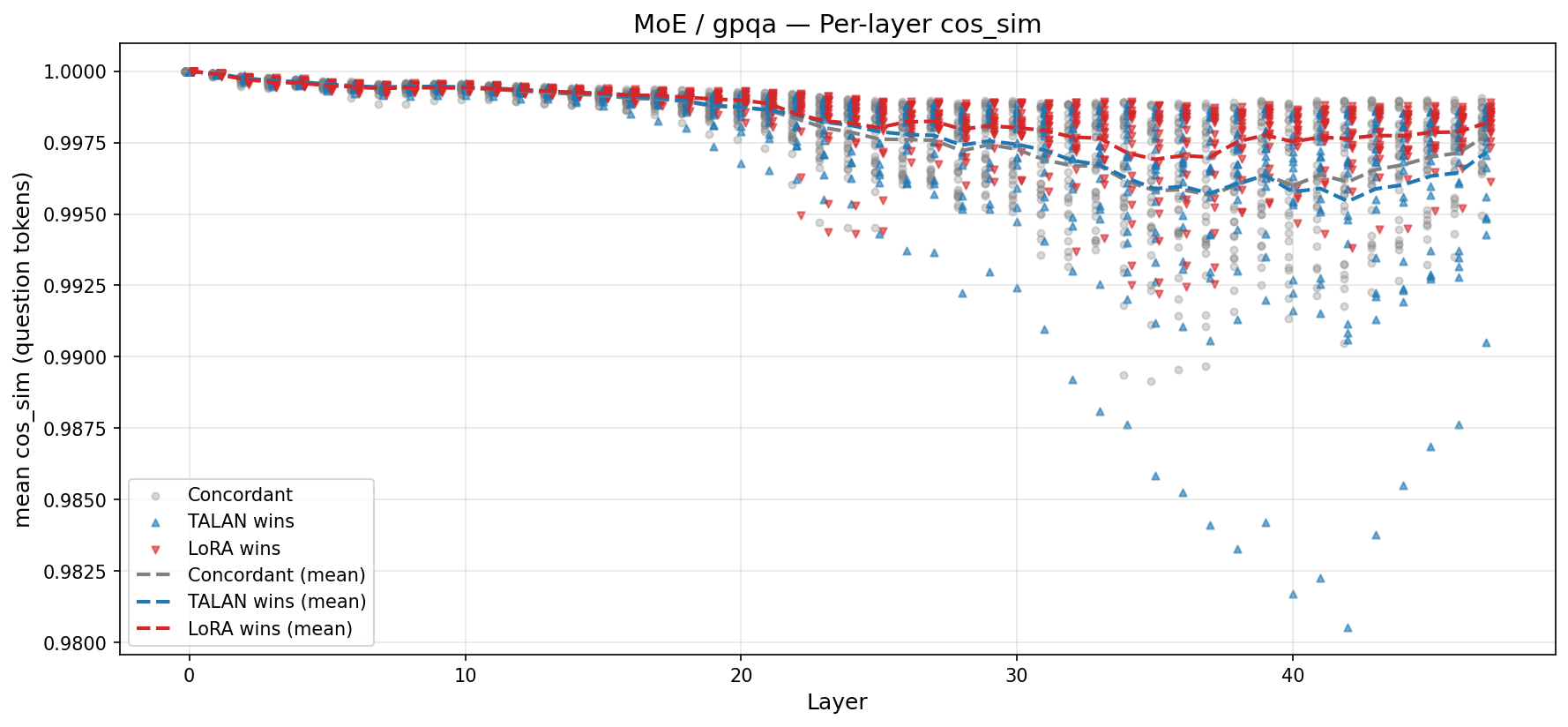} &
\includegraphics[width=0.46\textwidth]{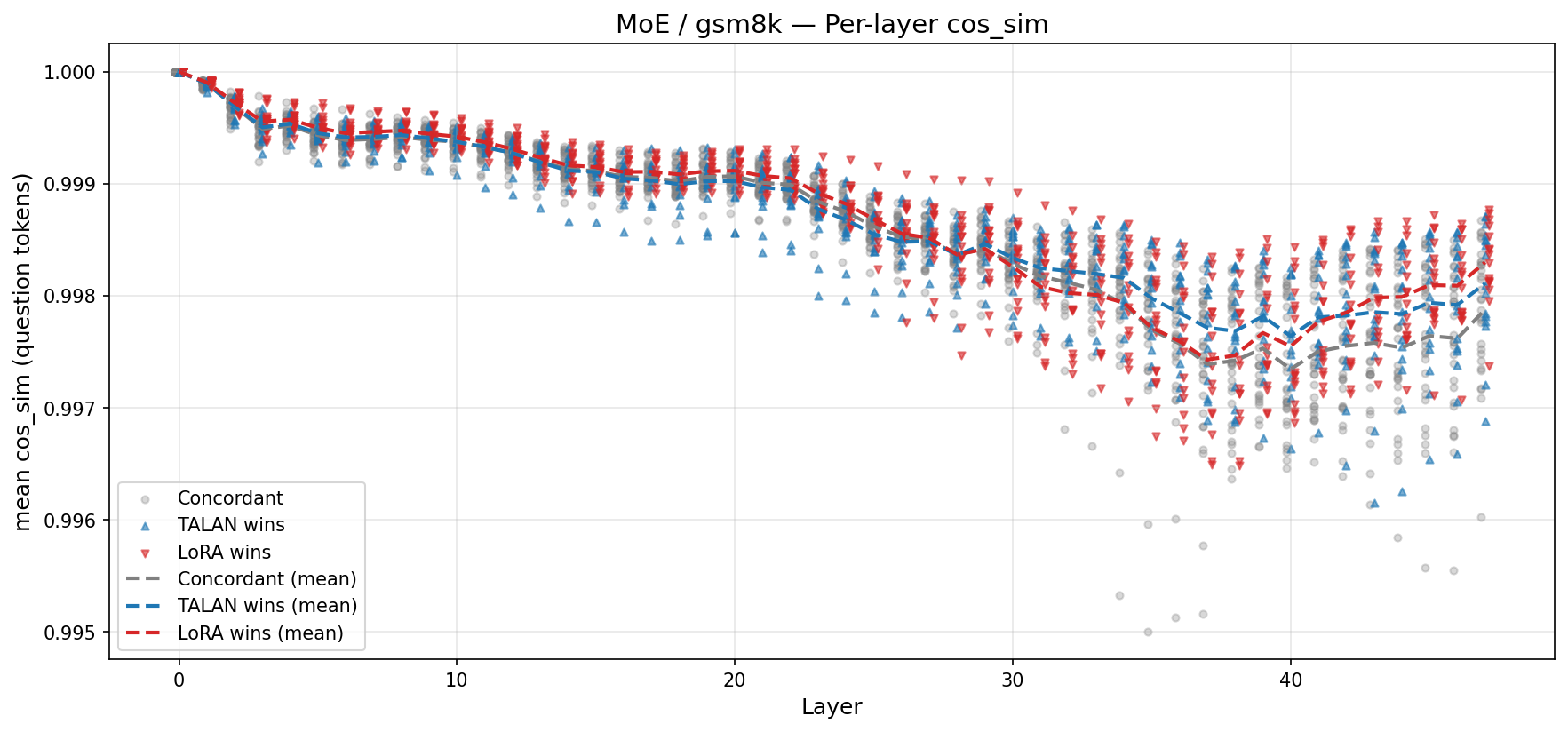} \\
\includegraphics[width=0.46\textwidth]{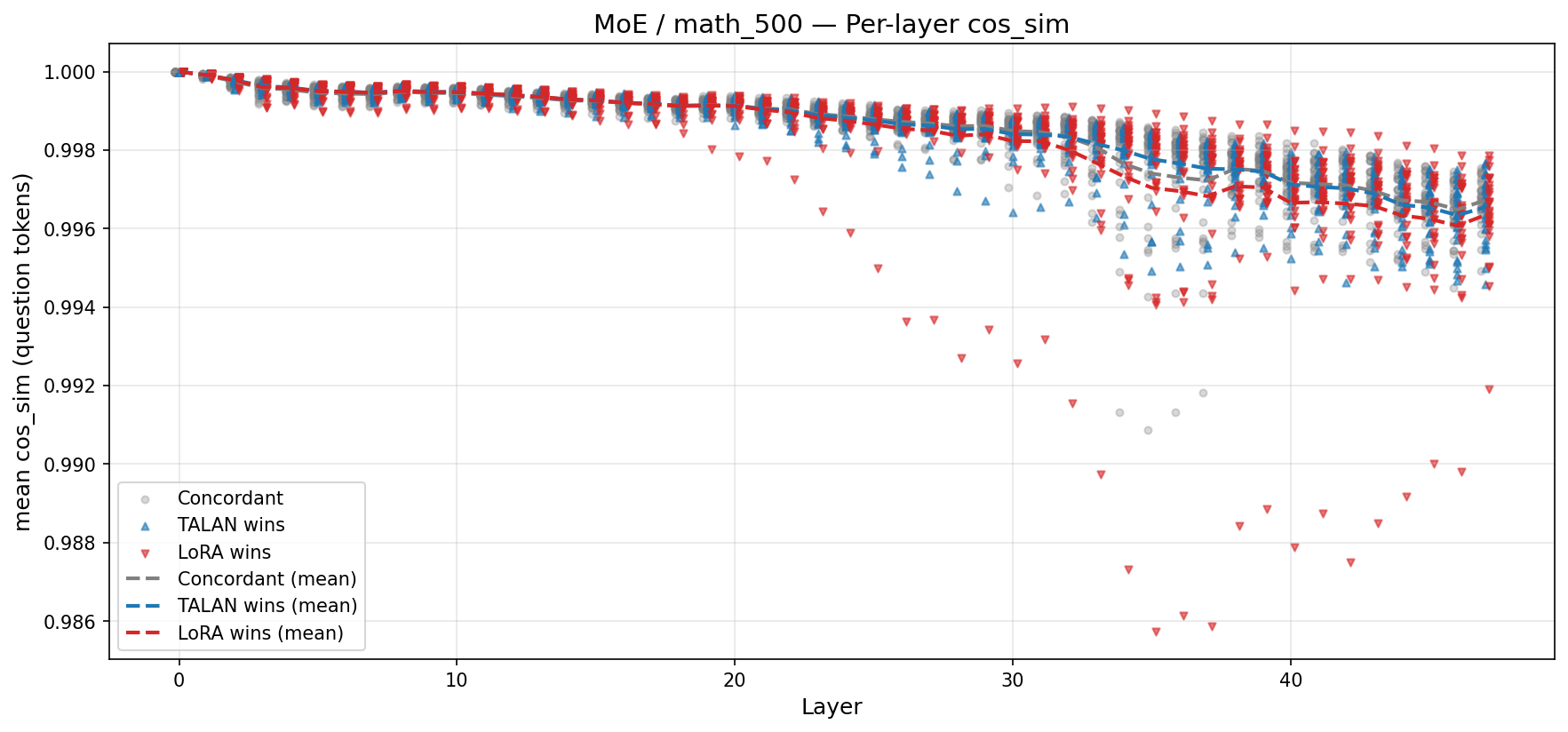} &
\includegraphics[width=0.46\textwidth]{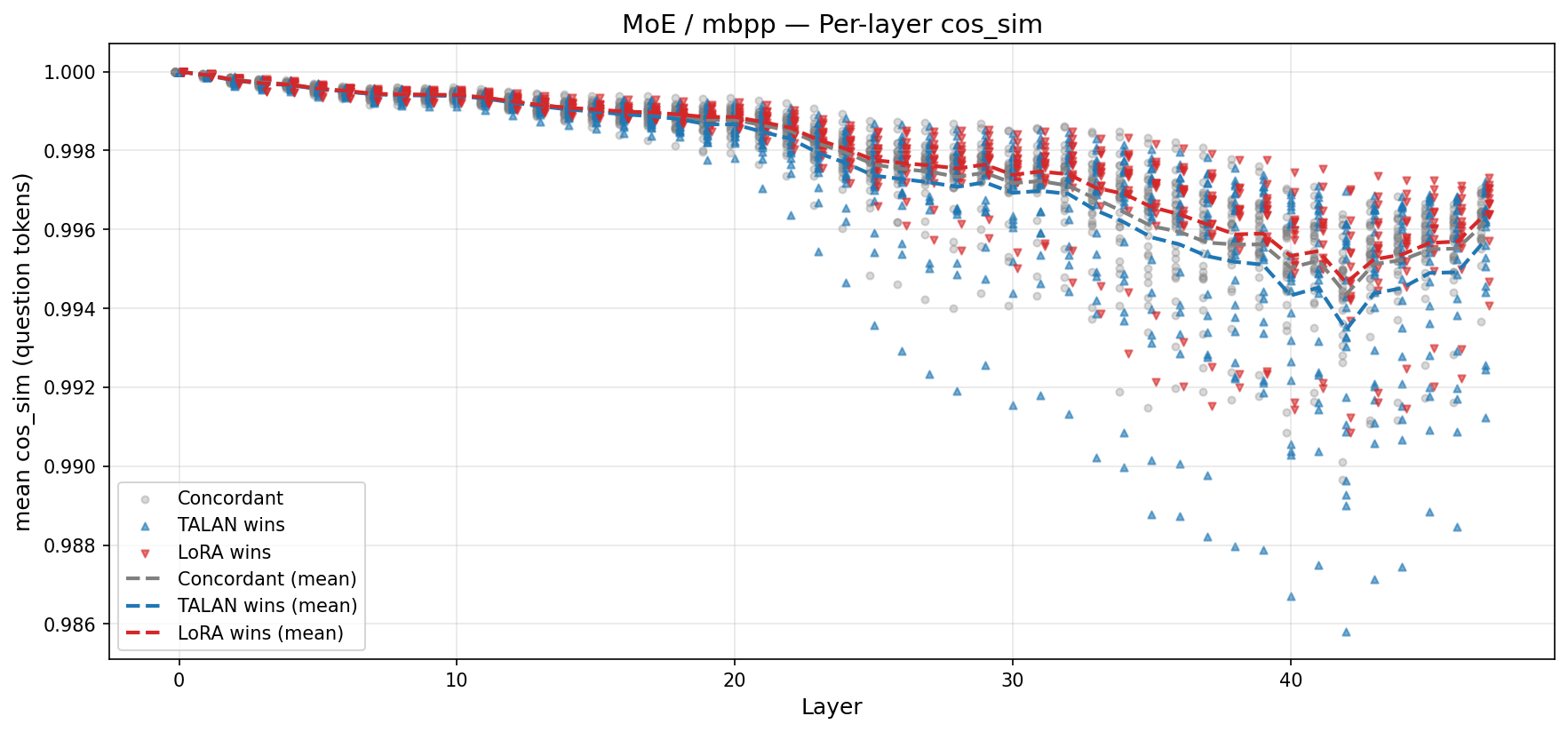} \\
\end{tabular}
\caption{\textbf{Qwen3-30B-A3B (MoE) per-layer cosine similarity},
four benchmarks.}
\label{fig:cos-moe}
\end{figure}

The cosine-similarity counterpart (Figures~\ref{fig:cos-q8b}--%
\ref{fig:cos-moe}) shows the per-layer cosine between the
residual-stream activations of \talan{}+LoRA and the matched LoRA
baseline. Cosine values stay close to $1.0$ at the insertion layer
(small perturbation) and decay smoothly with depth, consistent with
amplification of a small but coherent direction.

\begin{figure}[!ht]
\centering
\begin{tabular}{cc}
\includegraphics[width=0.46\textwidth]{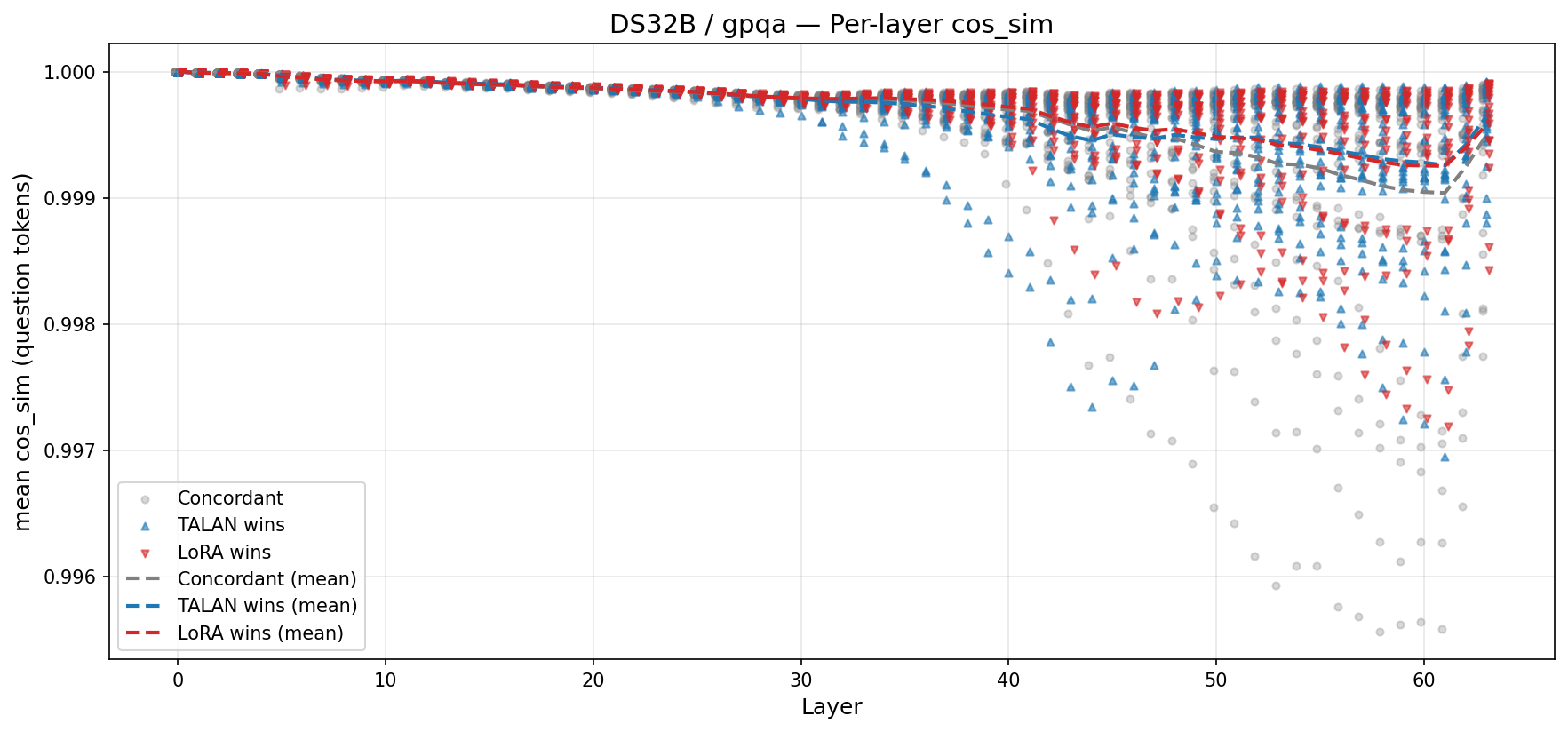} &
\includegraphics[width=0.46\textwidth]{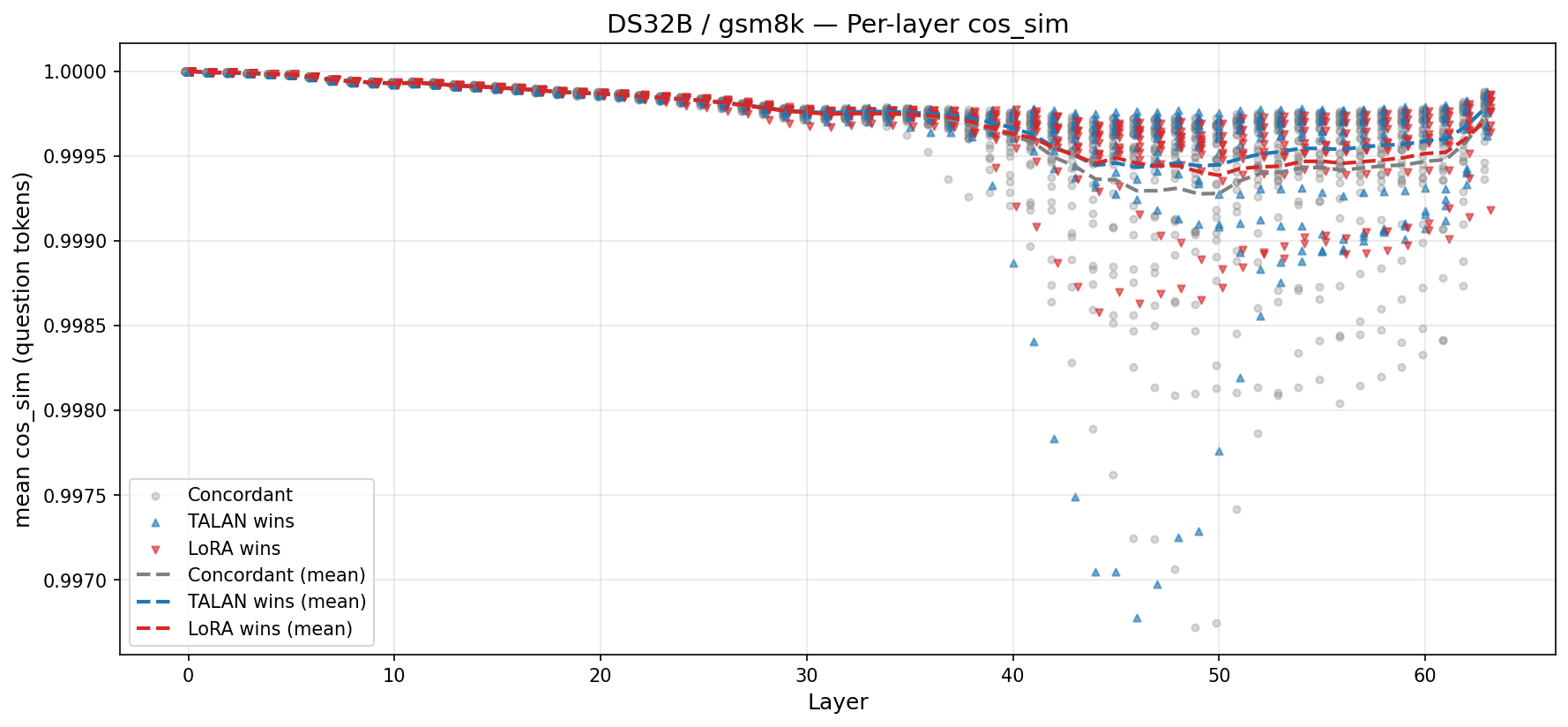} \\
\includegraphics[width=0.46\textwidth]{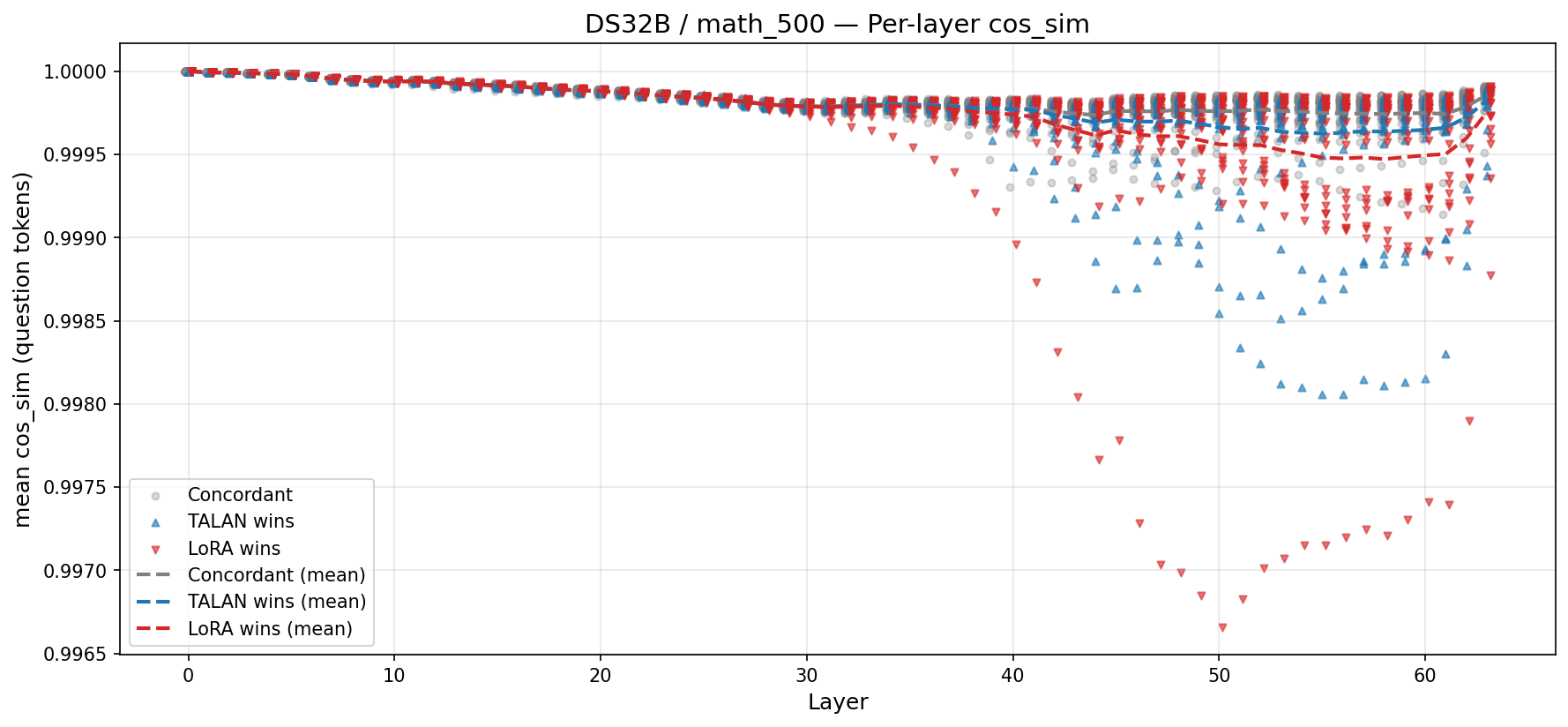} &
\includegraphics[width=0.46\textwidth]{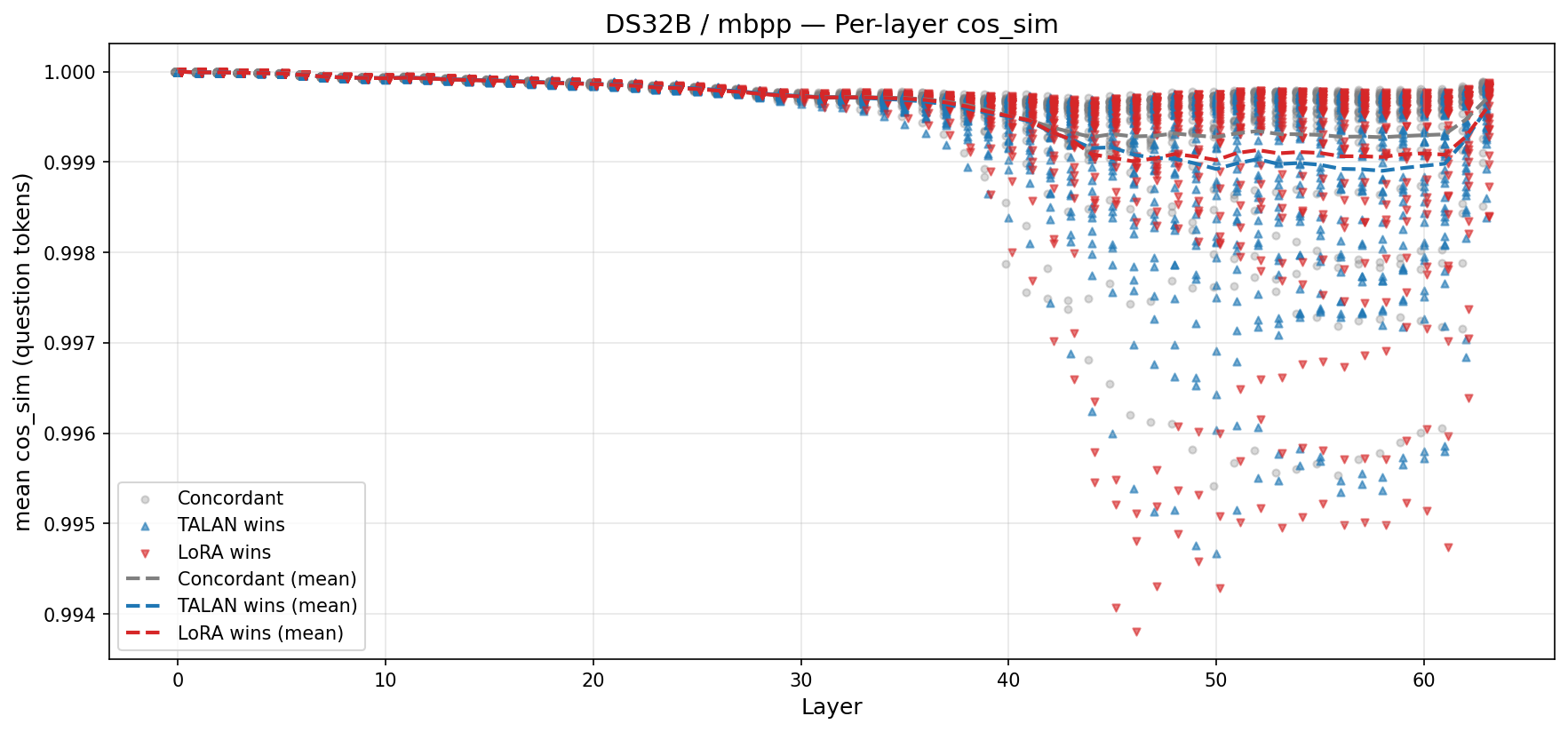} \\
\end{tabular}
\caption{\textbf{DeepSeek-R1-Distill-Qwen-32B per-layer cosine
similarity}, four benchmarks.}
\label{fig:cos-ds32b}
\end{figure}

\begin{figure}[!ht]
\centering
\begin{tabular}{cc}
\includegraphics[width=0.46\textwidth]{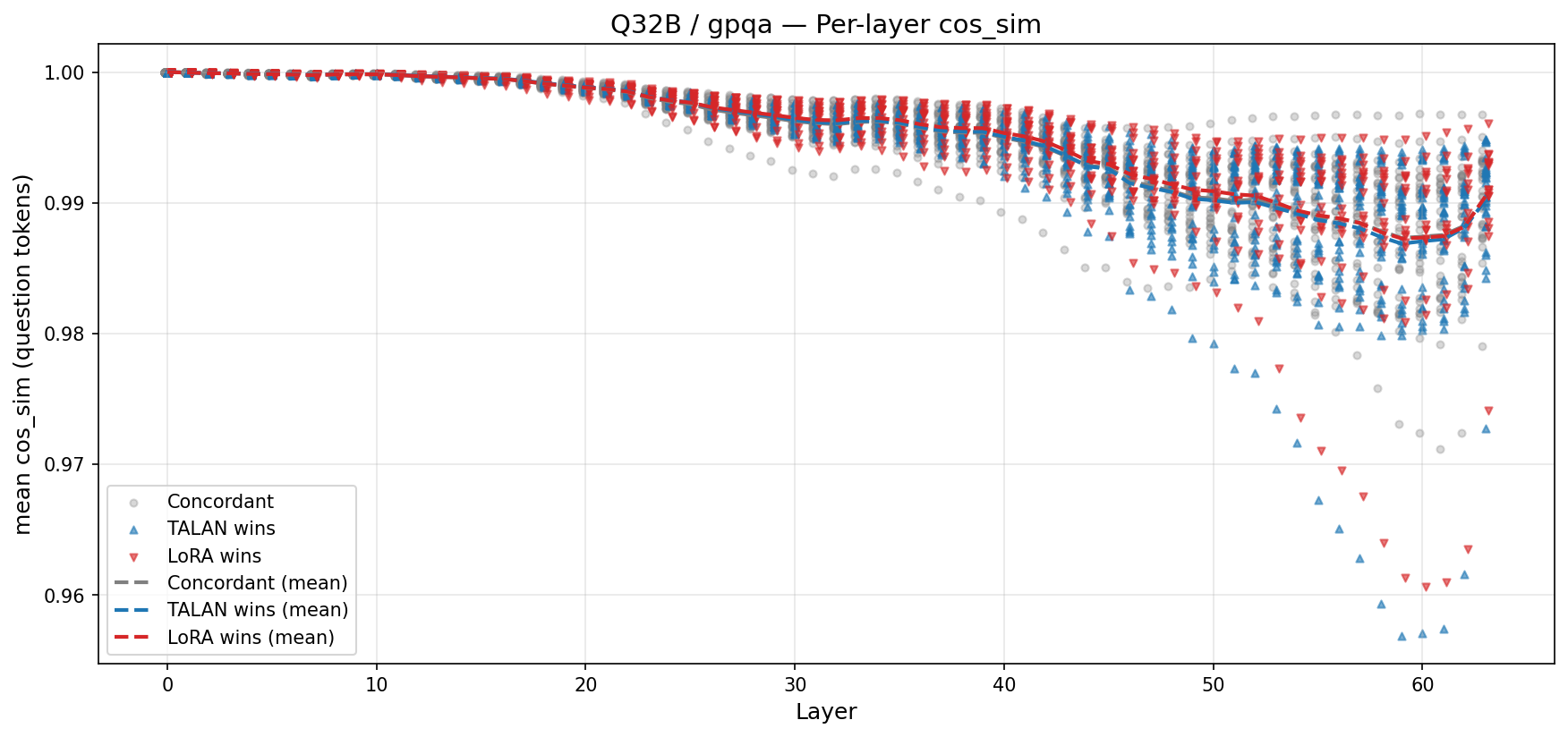} &
\includegraphics[width=0.46\textwidth]{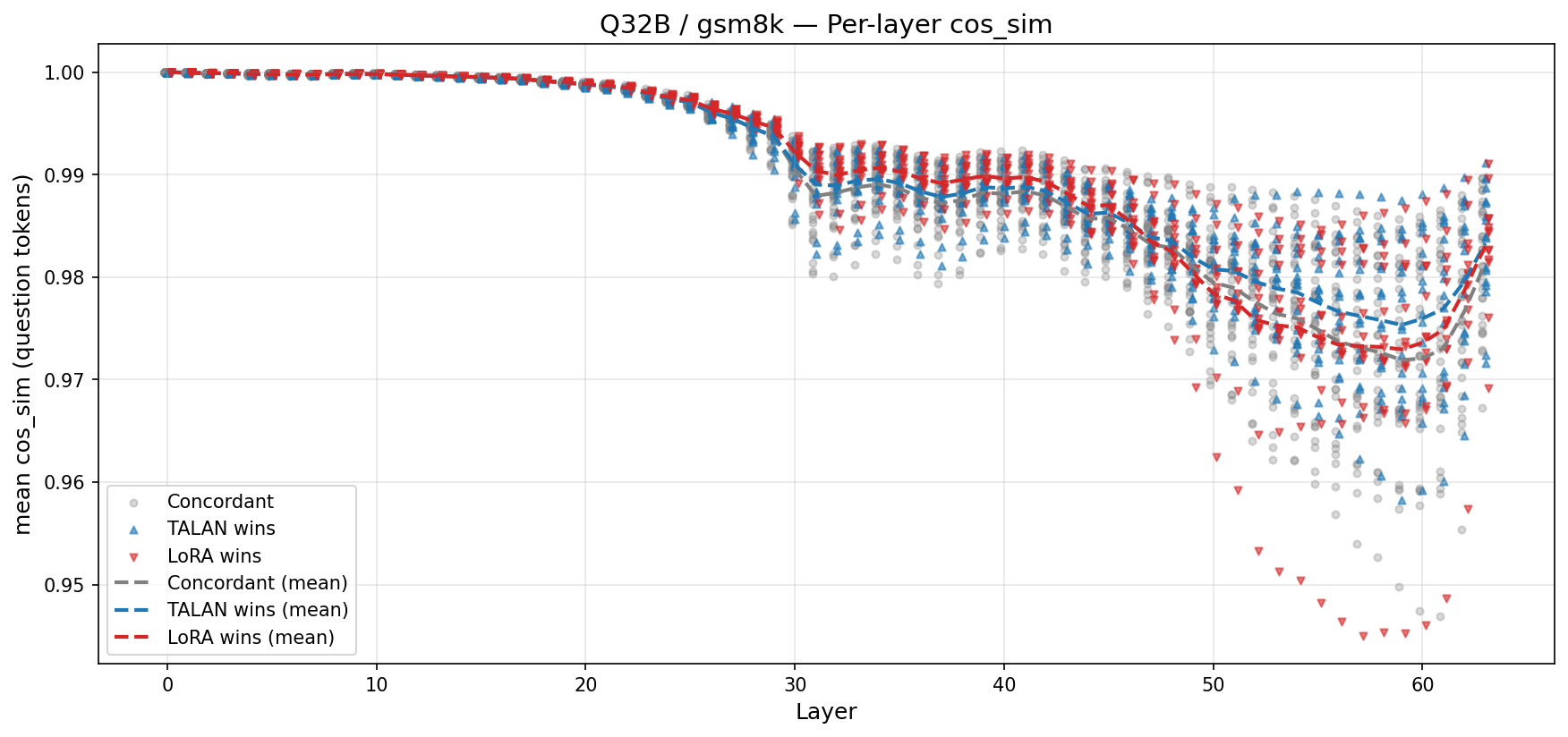} \\
\includegraphics[width=0.46\textwidth]{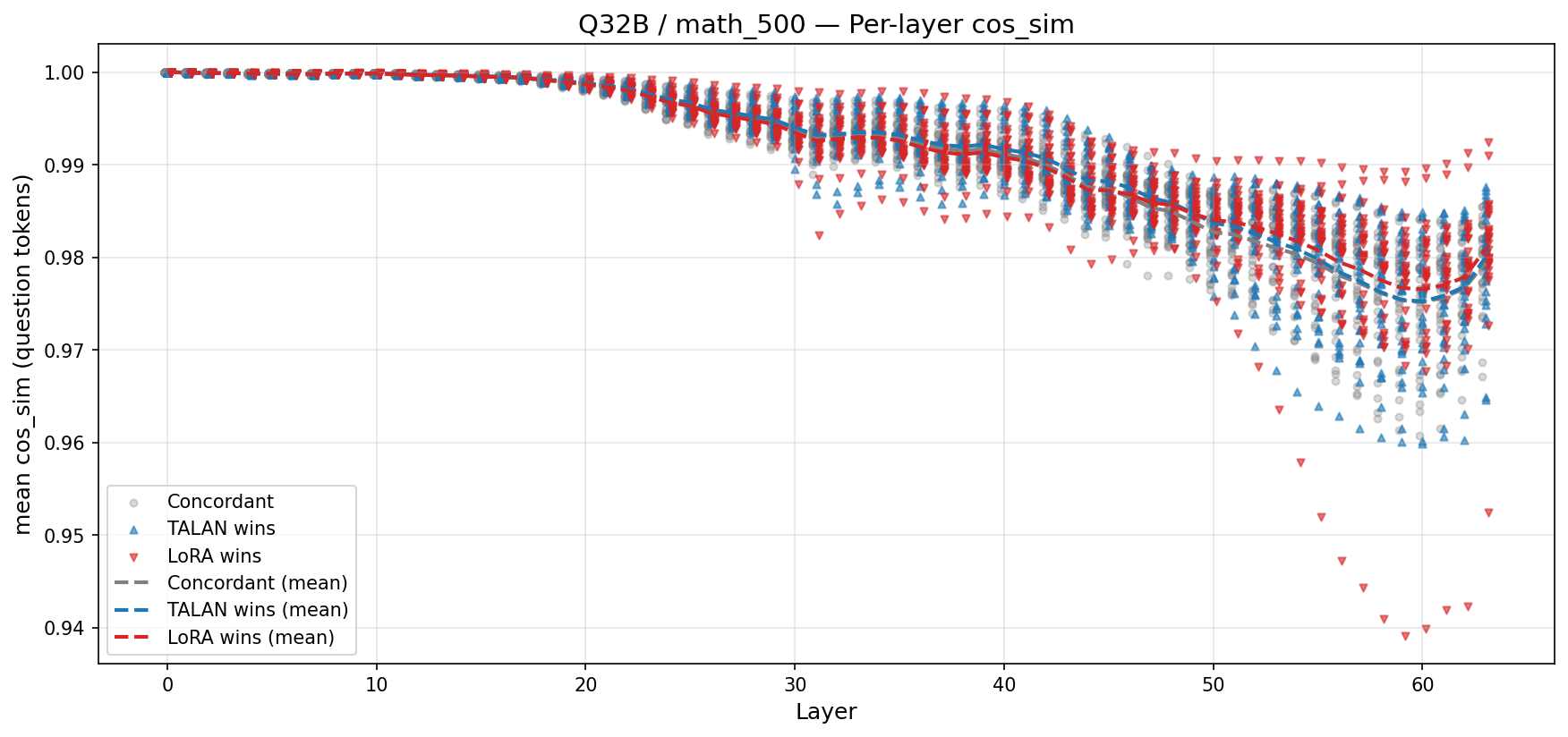} &
\includegraphics[width=0.46\textwidth]{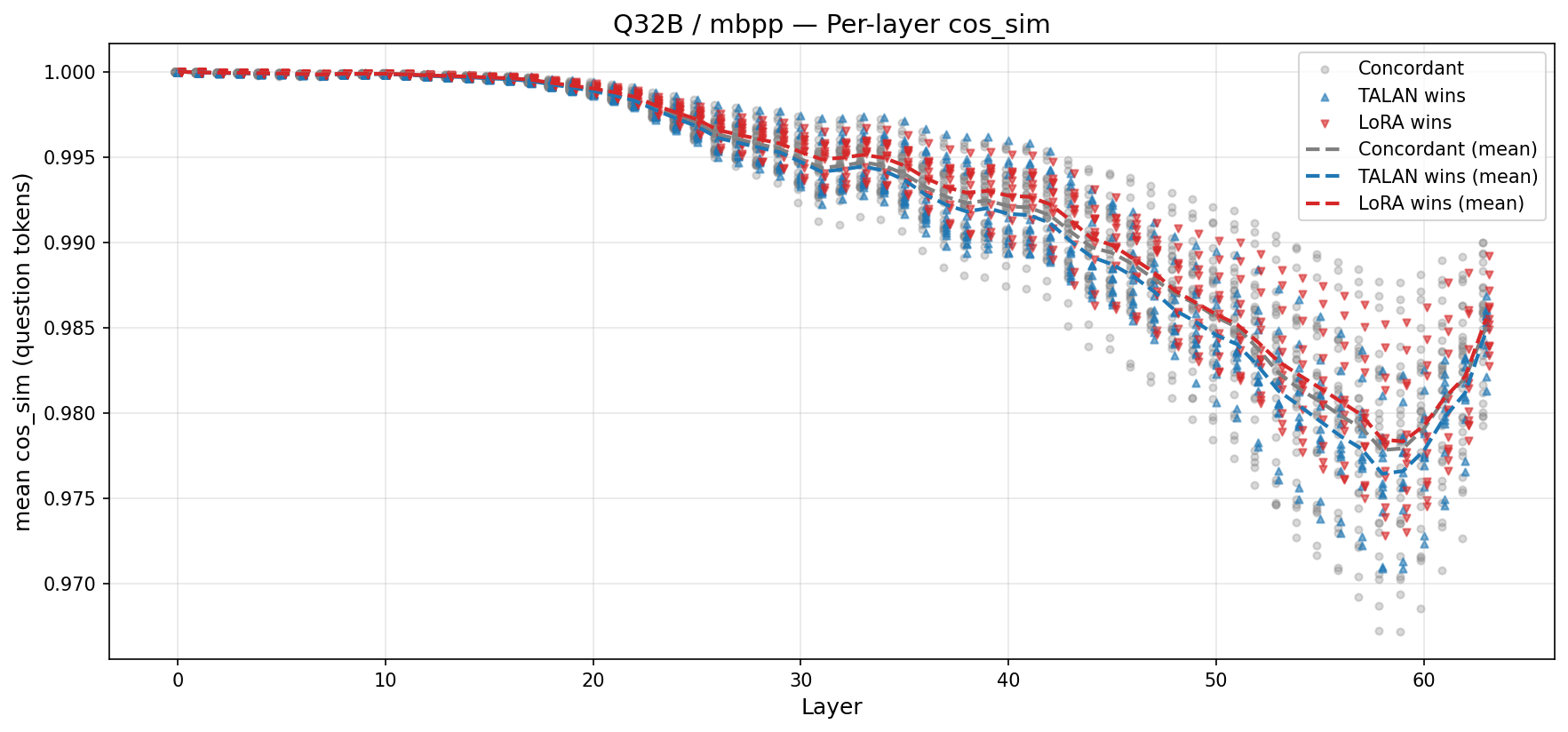} \\
\end{tabular}
\caption{\textbf{Qwen3-32B per-layer cosine similarity}, four
benchmarks.}
\label{fig:cos-q32b}
\end{figure}

\subsection{Per-layer observations}

Three observations are consistent across all 16 (backbone $\times$
benchmark) plots:
\begin{enumerate}[nosep,leftmargin=*]
  \item \textbf{Insertion-layer step.} The \talan{} module's insertion
    locus is visible as the layer at which the rel\_diff curve first
    lifts off zero. For Q32B the lift is at layer 12, for Q8B at layer
    4, for DS32B and MoE at the embedding (layer 0). This confirms
    that the configured insertion point is the actual source of the
    divergence.
  \item \textbf{Persistence (not absorption).} The rel\_diff grows
    monotonically through the network rather than being absorbed or
    decaying. This is a necessary condition for a small insertion-layer
    perturbation to affect downstream outputs; it is expected in deep
    residual networks but not guaranteed---a perturbation orthogonal to
    the directions that subsequent layers amplify could be suppressed.
    The monotonic growth confirms that the \talan{} perturbation lies
    in directions that the host network propagates.
  \item \textbf{Cross-benchmark stability (the main finding).} The
    shape of the per-layer divergence curve is essentially identical
    across the four benchmarks for each backbone. The representation
    shift \talan{} induces is a property of the model variant, not of
    the prompt distribution. This distinguishes the \talan{}
    intervention from prompt-dependent noise: a random or
    data-dependent perturbation would be expected to produce
    prompt-distribution-dependent divergence shapes, whereas the
    observed stability suggests a structural intervention whose
    propagation is governed by the host network's layer-wise geometry
    rather than by the input.
\end{enumerate}

\section{Weight-Level Interpretability (Full)}
\label{app:weights}

This appendix provides the full per-tensor weight-level analysis
supporting the activation-level findings of
Section~\ref{sec:mechanism}.

\subsection{Drift audit}

For each champion, we construct a fresh \talan{} module from the same
config with \texttt{torch.manual\_seed(0)}, then per-tensor compute
$\mathrm{rel\_dist} = \|\mathrm{saved} - \mathrm{fresh}\|_2 /
\max(\|\mathrm{saved}\|, \|\mathrm{fresh}\|)$. For random-init tensors
(theme queries, in\_proj weights, out\_proj weights, theme projections),
$\mathrm{rel\_dist} \approx \sqrt{2} \approx 1.41$ would mean ``saved is
indistinguishable from a fresh draw.'' For deterministic-init tensors
(\texttt{integration.gate\_vector} init $= \alpha \cdot \mathbf{1}$,
biases init $= 0$), drift away from the init formula indicates training
updates.

We find that the random structural tensors (theme queries, in\_proj
weights, out\_proj weights, theme projections) stay close to
fresh-sample distance on every champion. For the three trained dense
champions, the trainable signal lives in the deterministic
\texttt{integration.gate\_vector} and the MHA biases. The MoE operating
point is forward-active but frozen by design, so it is reported
separately rather than counted as a trained \talan{} side path.

\begin{table}[!ht]
\centering
\caption{\textbf{Drift audit verdicts.} ``Movable det.\ drifted'' counts
the deterministic tensors (gate\_vector + biases) whose drift exceeds
the noise floor. The three dense champions show measurable
deterministic-parameter training; the MoE operating point is
forward-active but frozen by design.}
\label{tab:drift-summary}
\small
\begin{tabular}{lccc}
\toprule
Champion & Random tensors fresh-looking & Movable det.\ drifted & Verdict \\
\midrule
Q32B headline configuration          & 3 / 3 (100\%) & 3 / 3 & TRAINED \\
DS32B headline configuration         & 3 / 3 (100\%) & 3 / 3 & TRAINED \\
Q8B headline configuration           & 5 / 5 (100\%) & 1 / 1 & TRAINED \\
MoE headline configuration           & 3 / 4 (75\%)  & 0 / 2 & FROZEN \\
\bottomrule
\end{tabular}
\end{table}

\subsection{Per-bucket within-(backbone, mixer-family) discriminators}

For each (backbone, mixer-family) bucket with at least one winning
and one losing configuration matched on
(layer, $T$, $\alpha$, integration, gate), we compute the
within-bucket Spearman rank correlation between candidate
weight-statistics and the measured cross-cell $\Delta$. The
within-bucket comparison controls for coarse hyperparameter
confounds (the same backbone, mixer family, layer, slot count,
writeback coefficient, integration, and gate type are held fixed
inside each bucket).

\begin{table}[!ht]
\centering
\caption{\textbf{Within-(backbone, mixer-family) bucketed
discriminator signal.} Top recurring within-bucket signals
($n_{\text{buckets}}\!\geq\!3$, mean $|r|\!\geq\!0.4$,
sign consistency $\!\geq\!70\%$). Pooled across families, the same
metrics dilute to small Cohen $|d|\!\leq\!0.1$ effect sizes
(Simpson's paradox); we therefore report them only inside their
family of validity.}
\label{tab:bucketed-discriminators}
\small
\begin{tabular}{llcccc}
\toprule
Backbone & Family & Metric & $n$-bkts & mean $|r|$ & sign-cons. \\
\midrule
DS32B & gated\_cross\_attn & \texttt{theme\_pairwise\_cos\_std} & 4 & 0.499 & 100\% \\
Q32B  & multihead\_local   & \texttt{theme\_pairwise\_cos\_std} & 4 & 0.530 & 75\% \\
DS32B & multihead\_local   & \texttt{v\_to\_qk\_ratio}          & 8 & 0.502 & 75\% \\
MoE   & gated\_cross\_attn & \texttt{theme\_pairwise\_cos\_std} & 4 & 0.477 & 75\% \\
Q8B   & multihead\_local   & \texttt{v\_stable\_rank}            & 9 & 0.391 & 78\% \\
MoE   & multihead\_local   & \texttt{v\_to\_qk\_ratio}          & 18 & 0.422 & 72\% \\
MoE   & cross\_attn        & \texttt{theme\_norm\_std}           & 18 & 0.415 & 72\% \\
\bottomrule
\end{tabular}
\end{table}

\paragraph{Reading.}
\texttt{theme\_pairwise\_cos\_std} (lower in winners) is the most
cross-bucket-stable within-bucket signal we find: it recurs in
$4$ of $4$ multihead/gated families with sign-consistency
$\geq 75\%$. \texttt{v\_stable\_rank} polarity is family-specific
(lower in winners on Q8B/multihead and MoE/multihead, higher in
winners on DS32B/cross\_attn and Q32B/multihead) and we do not pool
it across families. The activation-probe failure-mode filter
parameters (per-backbone magnitude bands rejecting mixer collapse
and over-perturbation) together form a weight-only screening
protocol complementing the activation-level analysis of
Section~\ref{sec:mechanism}.

\paragraph{Note on prior pooled signatures.}
Earlier internal versions of this audit reported
\texttt{theme\_q\_norm\_std} and \texttt{gate\_vec\_cv} as cross-family
discriminators. With the negatives extended to $5$ per backbone in a
follow-up audit, the pooled effect dilutes to $|d|\!\leq\!0.1$
(matched-$\alpha$ pos and neg checkpoints overlap on $6$--$8$ of every
$10$). We therefore restrict the present claim to the within-bucket
discriminators of
Table~\ref{tab:bucketed-discriminators}.

\subsection{\texorpdfstring{Signature--vs--$\Delta$ correlation}{Signature-vs-Delta correlation}}

\begin{figure*}[!ht]
\centering
\includegraphics[width=0.98\textwidth]{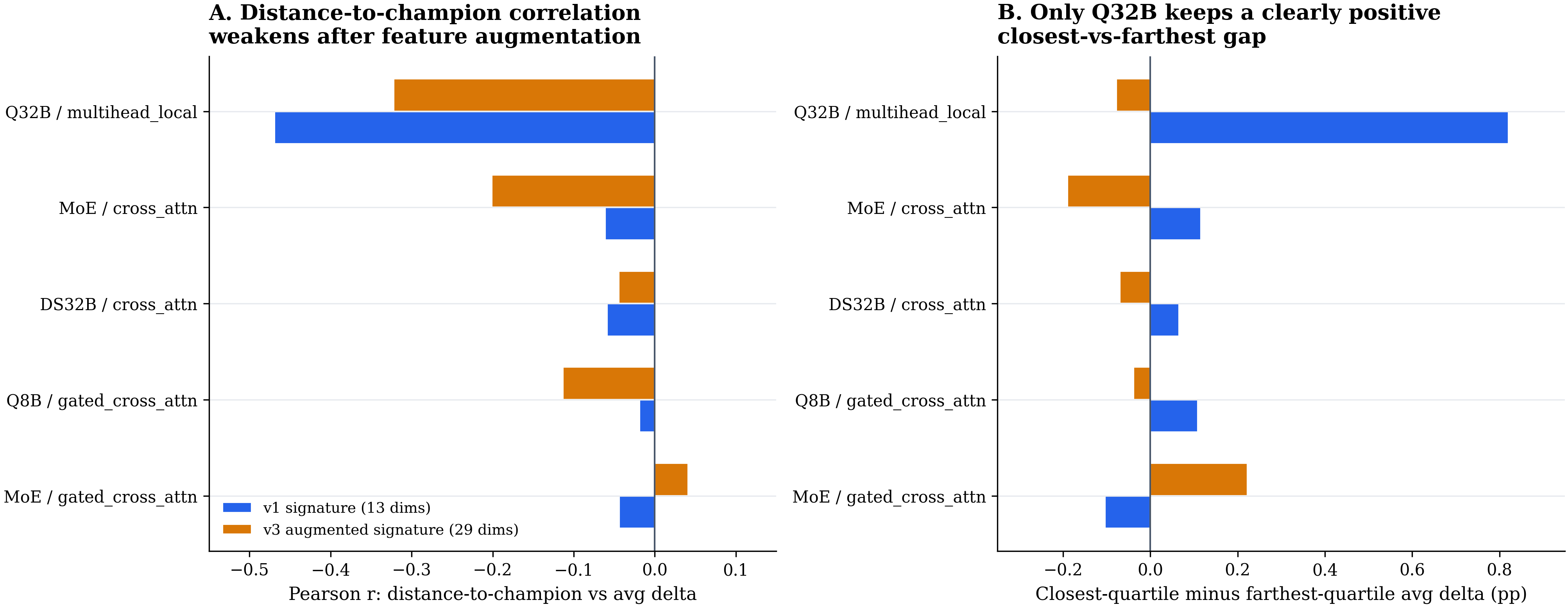}
\caption{\textbf{Distance-to-champion correlation audit.} The original
13-dimensional champion signature and the later 29-dimensional
augmented signature tell the same negative story: only
Q32B/\texttt{multihead\_local} exhibits a moderate negative
distance-vs-$\Delta$ correlation, and the augmented feature set
weakens rather than improves that signal. The main paper therefore
does not rely on a universal champion-distance score.}
\label{fig:app-signature-corr}
\end{figure*}

The older ``distance to a champion signature'' audit and the later
29-dimensional augmented-feature revision agree on the same qualitative
conclusion: a global champion-distance score is \emph{not} a universal
screening signal. Figure~\ref{fig:app-signature-corr} compares the two
audits bucket-by-bucket. Only Q32B/\texttt{multihead\_local} shows a
moderate negative distance-vs-$\Delta$ correlation in the original
13-dimensional audit ($r=-0.47$, closest-vs-farthest quartile gap
$+0.82$~pp), and the augmented signature weakens rather than
strengthens that signal ($r=-0.32$, quartile gap $-0.08$~pp). The
other audited buckets remain weak or inverted. This is why the present
paper's main claim is framed around recurring within-bucket
discriminators rather than a universal distance-to-champion metric.

\subsection{Ensemble component decomposition}

\begin{figure*}[!ht]
\centering
\includegraphics[width=0.98\textwidth]{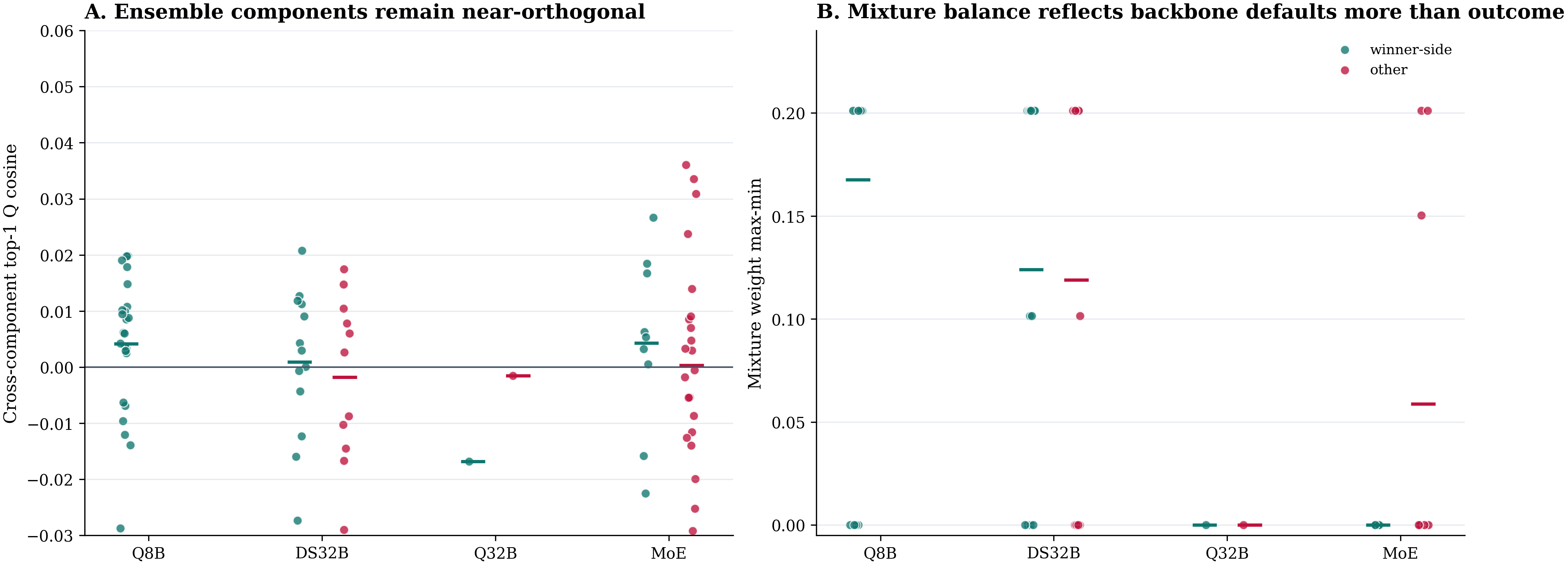}
\caption{\textbf{Ensemble-component audit over 82 checkpoints.} (A)
Cross-component top-1 Q singular-direction cosine remains centered near
zero for both winner-side and other checkpoints, indicating that
ensemble components behave like ordinary near-orthogonal random draws.
(B) Mixture imbalance is driven more by backbone defaults than by
outcome class. This supports the reading that any ensemble benefit
lives in activation-level composition, not in a special static
per-component signature.}
\label{fig:app-ensemble-decomp}
\end{figure*}

A plausible alternative story is that ensemble \talan{} winners derive
from unusually aligned component pairs or from learned nontrivial
mixture weights. We therefore decomposed $82$ ensemble checkpoints into
per-component Q/K/V signatures, cross-component top-1 singular-vector
cosine, and mixture balance. Figure~\ref{fig:app-ensemble-decomp}
shows the negative result: component pairs remain near-orthogonal on
both winner-side and other checkpoints, while mixture max--min is
largely backbone-default (Q8B and DS32B often preserve an imbalanced
mixture, MoE remains close to $0.5/0.5$). Static per-component weights
therefore do not explain the ensemble winners; if ensembles add value,
that value likely lives in the activation-level composition of two
random projections rather than in a special static alignment visible in
saved weights alone.

\subsection{Trainability axis distinguishes two mechanisms}
\label{sec:trainability-axis-mechanism}

The trainability axis $\tau$ splits the four audited configurations
into two mechanistically distinct groups.

\paragraph{Trained intervention (Q32B, DS32B, Q8B).}
For the three dense-backbone configurations trained with full
re-enable and a non-trivial gradient scale $\gamma$, the
\texttt{integration.gate\_vector} drifts substantially during
training: \texttt{gate\_vec\_std} ranges $0.00045$--$0.00204$ and
\texttt{gate\_vec\_cv} ranges $0.24$--$1.83$ across the three.
These configurations instantiate the core \talan{} proposal:
a learned, sequence-conditioned side-path intervention co-trained
with the adapter.

\paragraph{Frozen-perturbation variant (MoE).}
For the MoE configuration (forward-active / frozen-by-design
trainability), the gate vector stays exactly at initialization
($\text{std} = 0$). No \talan{} parameters are updated; the gain
comes entirely from a random-init forward perturbation that the
LoRA adapter learns to compose around. This is a separate finding
from the trained \talan{} intervention: it shows that a fixed,
unlearned residual-stream perturbation at the right location can
provide a useful structural prior for adapter training, even without
sequence conditioning. The two mechanisms are complementary---the
trained variant shows that learning the perturbation improves
results on dense backbones, while the frozen variant shows that
even random structure at the insertion point can benefit adapter
composition on MoE architectures.

\subsection{LoRA-subspace alignment (full)}

We measure $\Delta_{\mathrm{TALAN}} = \mathrm{TALAN}(X) - X$ and
$\Delta_{\mathrm{LoRA}} = X \cdot \sum_{\ell \in q\_proj} (B_\ell A_\ell)^T \cdot \mathrm{scale}$
(summed over all $q$-projection LoRA pairs across layers, with
$\mathrm{scale} = \alpha / r = 64/32 = 2$), at fixed synthetic input
$X \sim \mathcal{N}(0, \sigma^2 I_d)$ with $\sigma = \sqrt{d}/20$.
Per-token cosines and magnitude ratios are reported in
Table~\ref{tab:lora-subspace} of the main text. The orthogonality
(cosine $\approx 0$) holds for both winners and losers; the magnitude
ratio (LoRA 80--1{,}700$\times$ larger than \talan{}) is an
architectural property of the LoRA scale, not of the training outcome.

\section{Extended Case Studies}
\label{app:cases}

This appendix provides case studies
covering both improvements and degradations, together with their
per-token representation-divergence patterns.

\subsection{Structural organization (Q8B / MATH-500)}

\textbf{Problem.} ``Find the number of ordered pairs $(a,b)$ of integers
such that $|a + bi| \le 5$.''

\textbf{Gold answer.} 81.

\textbf{Matched LoRA baseline} (wrong, answers 84): enumerates all 11
values of $a$ from $-5$ to $+5$, correctly counts valid $b$ values for
each ($1 + 7 + 9 + 9 + 9 + 11 + 9 + 9 + 9 + 7 + 1$), but mis-sums to
84 in a single arithmetic step.

\textbf{\talan{}+LoRA} (correct, answers 81): groups symmetric cases
($a$ and $-a$ give the same $b$ count), reducing the sum to 6 terms:
$11 + 18 + 18 + 18 + 14 + 2 = 81$.

\textbf{Observation.} Both models compute the same per-case counts. The
difference is in aggregation strategy. \talan{}+LoRA exploits the
symmetry of the problem to produce a shorter, less error-prone
summation.

\subsection{Multi-step algebraic transformation (DS32B / MATH-500)}

\textbf{Problem.} ``Write $\sqrt{2} + 1/\sqrt{2} + \sqrt{3} + 1/\sqrt{3}$
in the form $(a\sqrt{2} + b\sqrt{3})/c$ with $a, b, c$ positive
integers and $c$ minimal; find $a+b+c$.''

\textbf{Gold answer.} 23.

\textbf{Matched LoRA baseline} (wrong, answers 13): correctly combines
$\sqrt{2} + 1/\sqrt{2} = 3/\sqrt{2}$ and
$\sqrt{3} + 1/\sqrt{3} = 4/\sqrt{3}$, then skips the rationalization
step, jumping directly to
$3/\sqrt{2} + 4/\sqrt{3} = (3\sqrt{2} + 4\sqrt{3})/6$. Reports
$a=3, b=4, c=6 \Rightarrow a+b+c = 13$.

\textbf{\talan{}+LoRA} (correct, answers 23): same initial combination,
then rationalizes each term: $3/\sqrt{2} = 3\sqrt{2}/2$,
$4/\sqrt{3} = 4\sqrt{3}/3$. Finds the LCD~$=6$, adjusts numerators to
$9\sqrt{2}/6$ and $8\sqrt{3}/6$, combines to
$(9\sqrt{2} + 8\sqrt{3})/6$. Reports
$a=9, b=8, c=6 \Rightarrow a+b+c = 23$.

\textbf{Observation.} Both models know the correct method. The
difference is that the LoRA baseline attempts to combine irrational
denominators in one step (skipping rationalization), while
\talan{}+LoRA executes the transformation through explicit intermediate
steps: rationalize $\to$ find LCD $\to$ adjust numerators $\to$
combine.

\subsection{When \talan{} hurts: geometric reasoning (Q8B / MATH-500)}

To present a balanced view we include cases where \talan{} hurts.

\textbf{Problem.} ``$\sin D = 0.7$ in the diagram below. What is $DE$?''
(Right triangle $DEF$ with right angle at $E$; $EF = 7$; gold $DE =
\sqrt{51}$.)

\textbf{Matched LoRA baseline} (correct, answers $\sqrt{51}$):
correctly identifies $\sin D = \mathrm{opposite}/\mathrm{hypotenuse} =
EF/DF = 7/DF$, gets $DF = 10$, then $DE = \sqrt{100-49} = \sqrt{51}$.

\textbf{\talan{}+LoRA} (wrong, answers 10): writes
$\sin D = EF/DE$ (incorrectly treating $DE$ as the hypotenuse), gets
$DE = 7/0.7 = 10$. Even states ``$DF$ is the hypotenuse'' in the middle
of the solution but contradicts this in the formula.

\textbf{Observation.} The error is in mapping the geometric
relationship---confusing which side is the hypotenuse vs.\ adjacent.
The \talan{} model's text is internally inconsistent, suggesting the
perturbation may have disrupted the spatial reasoning component on this
particular problem.

\subsection{When \talan{} hurts: excessive verification (DS32B / MATH-500)}

\textbf{Problem.} ``Let $z$ be a complex number such that
$z + 1/z = (1 + \sqrt{5})/2$. Find $z^{85} + 1/z^{85}$.''

\textbf{Gold answer.} $-2$.

\textbf{Matched LoRA baseline} (correct): recognizes $z + 1/z = \phi$
(golden ratio), writes $z = e^{i\theta}$ with $\cos\theta = \phi/2$,
identifies $\theta = \pi/5$, computes
$z^{85} + 1/z^{85} = 2\cos(85\pi/5) = 2\cos(17\pi) = -2$.

\textbf{\talan{}+LoRA} (wrong, extracted as $-2/3$): also discovers the
period-5 pattern ($w_5 = -2$), but second-guesses whether $z$ is on the
unit circle, re-derives the polar form analysis from scratch, gets
confused, re-derives again, and \emph{runs out of tokens} before
concluding. The answer extractor picks up $-2/3$ from the few-shot
example solutions rather than the actual answer.

\textbf{Observation.} \talan{}+LoRA discovers the correct approach but
enters a verification loop, exhausting the token budget before the
final answer. \talan{}'s tendency toward careful verification can
become counterproductive when it triggers excessive self-doubt on
correct intermediate results.

\subsection{Pattern summary across case studies}

Across the case-study set the dominant qualitative pattern is that
\talan{}+LoRA improves \emph{deliberation discipline}: it retains side
constraints introduced early in the prompt, exploits structural
symmetries when applicable, and executes transformations through
explicit intermediate steps rather than attempting one-step shortcuts
that drop information. On problems where the perturbation disrupts
spatial reasoning or triggers excessive verification, \talan{} can
hurt; the headline 16-of-16 non-regression at the cell level is the
\emph{net} of these competing effects across hundreds of items per
benchmark, not a per-item guarantee.

\begin{table}[!ht]
\centering
\caption{\textbf{Summary of observed patterns in case studies.}}
\label{tab:case-study-summary}
\small
\begin{tabular}{lll}
\toprule
\textbf{Pattern} & \textbf{\talan{} helps} & \textbf{\talan{} hurts} \\
\midrule
Constraint tracking         & Retains constraints LoRA drops & --- \\
Structural organization     & Groups by symmetry             & --- \\
Step-by-step transformation & Explicit intermediate steps    & --- \\
Geometric/spatial reasoning & ---                            & Confuses relationships \\
Verification behavior       & ---                            & Excessive self-doubt, token exhaustion \\
\bottomrule
\end{tabular}
\end{table}

\section{Cross-Adapter \textsc{matched-init} Multi-Seed Analysis}
\label{app:matched}

This appendix summarizes the cross-adapter \textsc{matched-init}
multi-seed protocol referenced in Section~\ref{sec:setup}. The purpose
is to quantify seed sensitivity without shifting the paper away from
its main contribution: a sequence-conditioned residual side path that
composes with standard PEFT adapters. We therefore report the
family-level paired statistic and use it as a sensitivity check rather
than as an additional headline claim about every individual cell.

\paragraph{Protocol (recap).}
The standard PEFT-paper training convention sets
\texttt{torch.manual\_seed(seed)} once at process start and lets
adapter initialization (LoRA-A's Kaiming, LoRA-B's zeros, plus the
random side-path init when applicable) consume the RNG state
non-deterministically across the treatment vs.\ baseline runs (because
the side-path allocation in TALAN runs offsets the RNG by an unequal
amount). The \textsc{matched-init} protocol re-calls
\texttt{torch.manual\_seed(seed)} \emph{immediately before}
\texttt{get\_peft\_model()} for both treatment and baseline, pinning
the same adapter init at each numerical seed. We sweep four canonical
seeds $\{42,43,44,48\}$ and pair each treatment seed against the
matched-init baseline at the \emph{same} seed value; we additionally
swap the LoRA adapter for a DoRA adapter under the same protocol. Each
(configuration, adapter) cell is therefore $n=4$ paired samples;
the paired $t$-statistic is
$t = \mathrm{mean}/(\mathrm{std}/\sqrt{n})$, and G1 at a seed means
all four per-benchmark deltas at that seed are non-negative.

\paragraph{Pooled cross-backbone matched-init paired statistic.}
Table~\ref{tab:cross-adapter} reports the pooled cross-backbone
result, computed by aggregating every available paired matched-init
delta on cells in the \talan{} family across all backbones and all
available seeds (canonical
$\{42,43,44,48\}$ plus extension seeds where collected). This pooled
view is the sensitivity statistic we use because per-cell sample size
is small ($n=4$ canonical), the 5{,}000-example training set introduces
visible training variance, and the intended claim is adapter
portability at the \talan{}-family level rather than seed invariance of
every individual operating point.

\begin{table}[!ht]
\centering
\caption{\textbf{Adapter portability under \textsc{matched-init}:
pooled paired statistic.} $n$ = total paired deltas across all
\talan{} configurations and seeds per adapter. Both adapters yield
$t > 3.8$, while the standard deviation captures seed-level spread.}
\label{tab:cross-adapter}
\small
\begin{tabular*}{\columnwidth}{@{\extracolsep{\fill}}lcccc@{}}
\toprule
Adapter (matched-init) & $n$ paired deltas & Mean $\Delta$ & Std $\Delta$ & Paired $t$ \\
\midrule
LoRA-matched & 55 & $+0.50$ & $0.97$ & $\mathbf{3.84}$ \\
DoRA-matched & 41 & $+0.47$ & $0.80$ & $\mathbf{3.81}$ \\
\bottomrule
\end{tabular*}
\\[0.4em]
\footnotesize
$\Delta$ in pp. This table is intended as a compact variance and
adapter-portability check, not as a claim that every single
configuration is seed-invariant.
\end{table}

\paragraph{Interpretation.}
The matched-init audit is intentionally not the centerpiece of the
paper. It shows that the positive effect is not solely a single
original-initialization artifact, but it also shows meaningful
seed-level spread. We therefore use the audit to calibrate the claim:
the main contribution is the \talan{} architecture and its
adapter-compatible operating points, while matched-init provides a
variance check supporting positive family-level trends.

\section{Workflow Comparison: Single-Stage \talan{} vs.\ Multi-Stage AdapterFusion}
\label{app:workflow}

\begin{figure}[!ht]
\centering
\includegraphics[width=0.96\columnwidth]{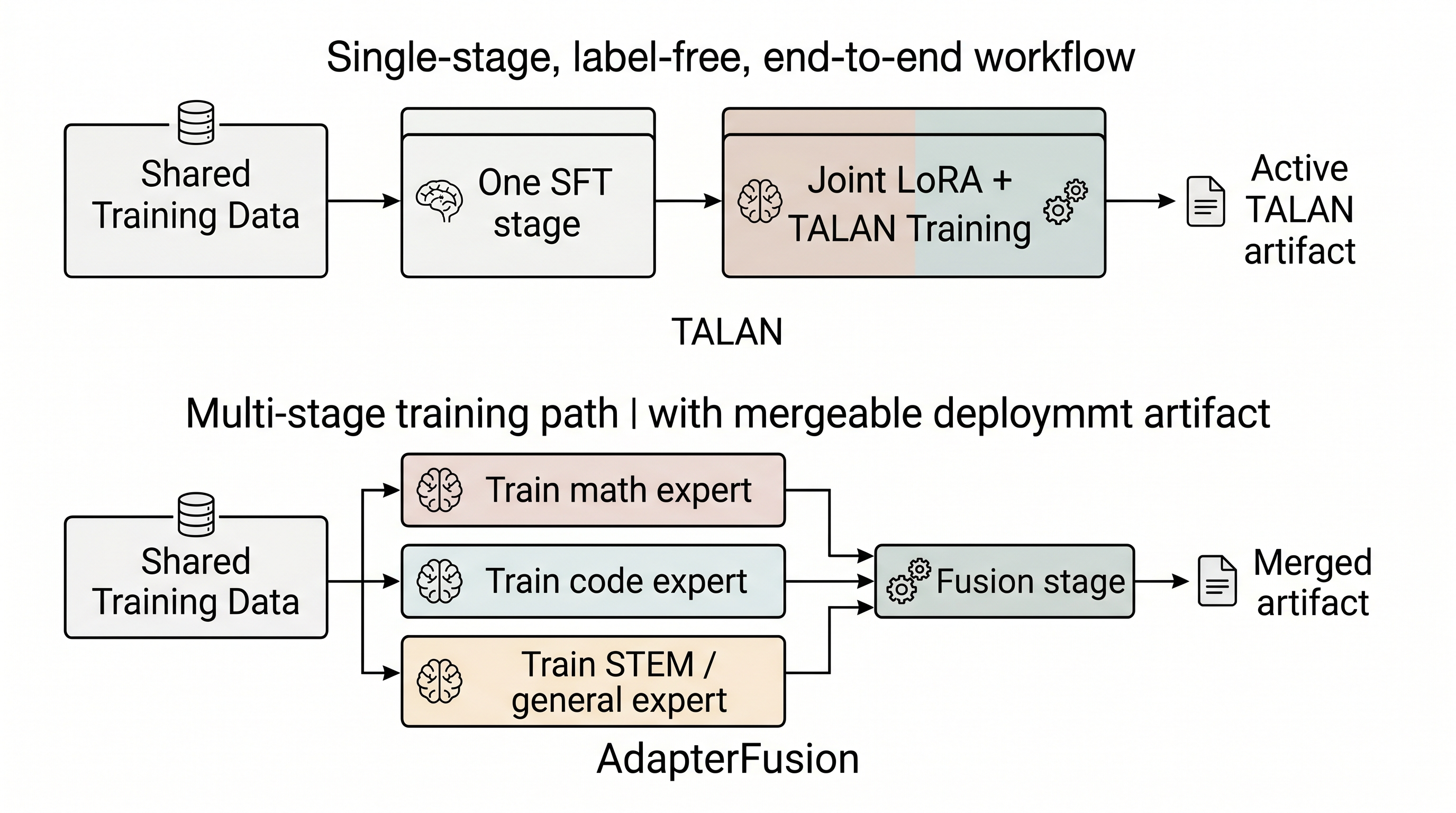}
\caption{\textbf{Single-stage \talan{} vs.\ multi-stage AdapterFusion
workflow.} \talan{}: shared training data $\to$ one SFT stage with
joint LoRA $+$ \talan{} training $\to$ active \talan{} path.
AdapterFusion: shared training data $\to$ per-domain adapter
trainings $\to$ separate fusion stage $\to$ fused adapter path.
\talan{}'s single-stage path keeps one active adaptation path and one
inference pass.}
\label{fig:workflow}
\end{figure}

A natural alternative to \talan{} is to train per-domain adapters
and learn a fusion stage on top, as in
AdapterFusion~\citep{pfeiffer2021adapterfusion}.
Figure~\ref{fig:workflow} contrasts the two workflows. \talan{}
requires no per-domain training, no domain labels at training time,
no fusion stage, and keeps a single active adaptation path (LoRA
adapter $+$ \talan{} state, totaling $\leq 1\%$ trainable
parameters). AdapterFusion requires per-domain adapter trainings
followed by a fusion stage; the resulting inference path has
$\approx 2\times$ training cost and larger overhead due to multiple
adapter passes. Empirically
(Table~\ref{tab:main} in the main text), AdapterFusion reaches a
comparable single-cell peak on Q8B but regresses on Q32B and MoE;
the single-stage \talan{} workflow is both cheaper and more
consistent across backbone architectures.

\section{Selected Non-Qwen Paired Transfer Check}
\label{app:nonqwen}

The primary experiments deliberately keep the backbone family fixed to
make the four-backbone configuration audit interpretable. After that
audit, we consolidated additional non-Qwen paired \talan{}+baseline
results from the same four-benchmark evaluation suite. We use the
authoritative paired summary to add only a narrow transfer check to the
paper: Llama-3.2-1B settings with seven paired seeds and positive
average deltas. We do not use raw-only runs without matched baselines or
exploratory families outside this criterion as claims.

\begin{table}[!ht]
\centering
\caption{\textbf{Authoritative paired Llama-3.2-1B transfer summary.}
Rows are selected by requiring seven paired seeds and positive mean
delta against the matched PEFT baseline.}
\label{tab:nonqwen-app}
\small
\begin{tabular}{lcccccc}
\toprule
Adapter & Seeds & Positive seeds & Mean $\Delta$ & Std. $\Delta$ & GPQA & GSM8K / MATH / MBPP \\
\midrule
LoRA   & 7 & 5/7 & $+$1.34 & 3.04 & $+$1.37 & $+$0.08 / $+$1.26 / $+$2.66 \\
rsLoRA & 7 & 5/7 & $+$0.60 & 0.58 & $+$1.95 & $+$0.27 / $+$0.20 / $+$0.00 \\
\bottomrule
\end{tabular}
\end{table}

The LoRA row has a larger seed-level standard deviation, while the
rsLoRA row is smaller but more stable. We therefore use this result
only to support limited transfer beyond the primary Qwen-family suite,
not to claim broad non-Qwen coverage.

\end{document}